\documentclass{article} 
\usepackage{nips13submit_e,times}
\pdfoutput=1
\usepackage{hyperref}
\usepackage{url}
\usepackage{epsfig}
\usepackage{graphicx}
\usepackage{amsmath}
\usepackage{amssymb}
\usepackage{caption}
\usepackage{subcaption}
\usepackage{xspace}

\newcommand{\source}{{source}\xspace}
\newcommand{\target}{{target}\xspace}

\newcommand{\argmin}{\operatornamewithlimits{argmin}}

\newcommand{\noprior}{\textbf{SVM}\xspace}
\newcommand{\SVcorr}{\textbf{SVM-SV}\xspace}
\newcommand{\MVcorr}{\textbf{SVM-MV}\xspace}

\newcommand{\MVcovmat}{\textbf{SVM-$\Sigma$}\xspace}

\newcommand{\MVcovmatTDtoND}{\textbf{SVM-$\Sigma$-TD2ND}\xspace}
\newcommand{\MVcovmatTDtoALL}{\textbf{SVM-$\Sigma$-TD2ALL}\xspace}
\newcommand{\MVcovmatNBtoALL}{\textbf{SVM-$\Sigma$-NB2ALL}\xspace}

\newcommand{\apvpd}{{AP+VP-D}\xspace}
\newcommand{\apvpc}{{AP+VP-C}\xspace}

\newcommand{\sourcew}{\textbf{w}^s}
\newcommand{\targetw}{\textbf{w}^t} \newcommand{\w}{\textbf{w}}
\newcommand{\x}{\textbf{x}} \newcommand{\tw}{\tilde{\textbf{w}}}
\newcommand{\tx}{\tilde{\textbf{x}}} 

\newcommand\myparagraph[1]{\vspace{1pt}\noindent\textbf{#1}\quad}

\title{Multi-View Priors for Learning Detectors From Sparse Viewpoint Data}

\author{Bojan Pepik$^1$ \And Michael Stark$^{1,2}$ \And Peter Gehler$^3$ \And Bernt Schiele$^1$\\
\and
\begin{tabular}{c}
\small
$^1$Max Planck Institute for Informatics, $^2$Stanford University,\\ \small $^3$Max Planck Institute for Intelligent Systems\\
\end{tabular}
}

\nipsfinalcopy 

\begin{document}

\maketitle

\begin{abstract}
While the majority of today's object class models provide only 2D
bounding boxes, far richer output hypotheses are desirable including
viewpoint, fine-grained category, and 3D geometry estimate. However,
models trained to provide richer output require larger amounts of
training data, preferably well covering the relevant aspects such as
viewpoint and fine-grained categories. In this paper, we address this
issue from the perspective of transfer learning, and design an object
class model that explicitly leverages correlations between visual
features. Specifically, our model represents prior distributions over
permissible multi-view detectors in a parametric way -- the priors are
learned once from training data of a source object class, and can
later be used to facilitate the learning of a detector for a target
class. As we show in our experiments, this transfer is not only
beneficial for detectors based on basic-level category
representations, but also enables the robust learning of detectors
that represent classes at finer levels of granularity, where training
data is typically even scarcer and more unbalanced. As a result, we
report largely improved performance in simultaneous 2D object
localization and viewpoint estimation on a recent dataset of
challenging street scenes.

\end{abstract}

\section{Introduction}
\label{sec:intro}
%
%
Motivated by higher-level tasks such as scene understanding and object tracking
it has been argued that object class models should not only provide 
flat, 2D bounding box detections but rather provide
%
more expressive output, such as a viewpoint
estimate~\cite{savarese07iccv,su09iccv,liebelt10cvpr,stark10bmvc,payet11iccv,villamizar11bmvc,glasner11iccv}
or an estimate of the 3D geometry of the object of
interest~\cite{pepik12cvpr,xiang12cvpr,pepik12eccv,fidler12nips,hejrati12nips,zia13cvpr}.
Similarly, there has been in increased interest in object 
representations that allow more fine-grained distinctions than
basic-level categories~\cite{lan13iccv,hoai13cvpr,stark12bmvc}, for two reasons. First, these
representations potentially perform better in recognition, as
they explicitly address the modes of intra-class variation. And
second, they, again, can provide further inputs to higher-level
reasoning (e.g., in the form of fine-grained category labels).

However, today's methods for 3D and fine-grained object representations suffer
from a major weakness: for robust parameter estimation, they
tend to require an abundance of annotated training data that covers
all relevant aspects (viewpoints, sub-categories) with sufficient
density.
%
Unfortunately, this abundance of training data cannot be expected in general.
Even in the case of dedicated multi-view
datasets~\cite{savarese07iccv,ozuysal09cvpr,arie-nachimson09iccv,stark12bmvc}
or when resorting to artificially rendered CAD
data~\cite{liebelt10cvpr,stark10bmvc,zia113drr},
the distribution of the number of available training images over
object categories is known to be highly unbalanced and
heavy-tailed~\cite{wang10cvpr,salakhutdinov11cvpr,lim11nips}. This is
particularly pronounced for categories at finer levels of granularity,
such as individual types or brands of
cars\footnote{Fig.~\ref{fig:cartypesstat} and~\ref{fig:carmodelstat}
  in Sect.~\ref{sec:suppl} show the viewpoint data statistics for {\em
    car-types} and {\em car-models} on KITTI~\cite{geiger12cvpr}.}.

Transfer learning has been acknowledged as a promising way to
leverage scarce training data, by reusing once acquired knowledge as a
regularizer in novel learning tasks~\cite{feifei06pami,stark09iccv,gao12eccv}.
While it has been shown that transfer learning can be beneficial for
performance, its use in computer vision has, so far, mostly been
limited to either classification
tasks~\cite{feifei06pami,zweig07iccv,rohrbach10cvpr,berg10eccv} or
flat, 2D bounding box detection~\cite{aytar11iccv,gao12eccv},
neglecting both the 3D nature of the recognition problem and more
fine-grained object class representations.

The starting point and major contribution of this paper is therefore
to design a transfer learning technique that is particularly tailored
towards multi-view recognition (encompassing simultaneous bounding box
localization and viewpoint estimation). It boosts detector performance
for scarce and unbalanced training data, lending itself to
fine-grained object representations.

To that end, we represent transferable knowledge as prior
distributions over permissible
models~\cite{feifei06pami,gao12eccv}, in two different flavors.
The first flavor (Sect.~\ref{sec:approach_model_one}) captures sparse
correlations between HOG
cells in a multi-view deformable part model
(DPM~\cite{felzenszwalb10pami}), across viewpoints. While this is
similar in spirit to~\cite{gao12eccv} in terms of statistical
modeling, we explicitly leverage 3D object
geometry~\cite{pepik12cvpr,pepik12eccv} in order to establish
meaningful correspondences between HOG cells, in a fully automatic
way. As we show in our experiments (Sect.~\ref{sec:exps}), this
already leads to some improvements in performance in comparison
to~\cite{gao12eccv}.
The second flavor (Sect.~\ref{sec:approach_model_two}) extends the
sparse correlations to a full, dense covariance matrix that
potentially relates each HOG cell to every other HOG cell, again
across viewpoints -- this can be seen as directly learning
transformations between different views, where the particular choice
of source and target cells can function as a regularizer on the
learned transformation, and leads to substantial improvements in
simultaneous bounding box localization and viewpoint
estimation.
Both flavors are simple to implement (covariance computation
for prior learning and feature transformation for prior application)
and hence widely
applicable, yet lead to substantial performance improvements for
realistic training data with unbalanced viewpoint distributions.

Our paper makes the following specific contributions:
{\em First}, to our knowledge, our work is the first attempt to
explicitly design a transfer learning technique for multi-view
recognition and fine-grained object representations.
{\em Second}, we propose two flavors of learning prior
distributions over permissible multi-view detectors, one based on
sparse correlations between cells and one based on the full
covariance. Both are conveniently expressed as instantiations of a
class of structured priors that are easy to implement and can be
readily applied to current state-of-the-art multi-view
detectors~\cite{felzenszwalb10pami,pepik12cvpr,pepik12eccv}.
And {\em third}, we provide an in-depth experimental study of our
models, first investigating multi-view transfer learning under the
controlled conditions of a standard multi-view
benchmark~\cite{savarese07iccv} (Sect.~\ref{sec:exp_comparison}), and
finally demonstrating improved performance for simultaneous object
localization and viewpoint estimation on realistic street
scenes~\cite{geiger12cvpr} (Sect.~\ref{sec:exp_multiview_detection}).
\section{Related work}
\label{sec:rw}
The problem of scarce training data has mainly been addressed in two
different ways in the literature.

\myparagraph{Generating training data.}  The first way is to
explicitly generate more training data for the task at hand, e.g., by
rendering CAD models of the object class of interest. Rendered data
has successfully been applied in the context of multi-view
recognition~\cite{liebelt08cvpr,liebelt10cvpr,stark10bmvc,zia113drr,pepik12cvpr,pepik12eccv},
people detection~\cite{pishchulin11cvpr,pishchulin12cvpr}, and indoor
scene understanding~\cite{delpero13cvpr}. The success of these
approaches is due to the fact that unlimited amounts of training data
can be generated with lots of variation in viewpoint, shape and
articulation, from just a few models. Existing approaches differ
mostly in the acquisition of appropriate models (3D
scanning~\cite{pishchulin11cvpr,pishchulin12cvpr} vs. manual
design~\cite{liebelt08cvpr,liebelt10cvpr,stark10bmvc,zia113drr,pepik12cvpr,pepik12eccv})
and the rendering output, ranging from close to photo-realistic images
~\cite{pishchulin12cvpr,liebelt08cvpr} to directly rendering
edge-based
features~\cite{stark10bmvc,pishchulin11cvpr,pepik12cvpr,pepik12eccv}. A
special case of data ``generation'' is the borrowing of training
examples from other object classes~\cite{lim11nips}, adapting feature
representations to exploit data from different
domains~\cite{kulis11cvpr} or using decorrelated
features~\cite{hariharan12eccv}.
While our model based on local correlation structures
(Sect.~\ref{sec:approach_model_one}) can leverage  CAD models
of the object class, this is done for the sole purpose of
deriving correspondences based on 3D geometry, and does not include
object appearance.

\myparagraph{Transfer learning.}
The second way is to first condense available training data into a
model, and then reusing that model in the context of another learning
task. While there is a vast amount of literature on this kind of
transfer learning, most approaches focus on object or image
classification rather than
detection~\cite{miller00cvpr,fink2004nips,bart2005bmvc,bartCVPR05,thrun2006nips,zweig07iccv,lampertCVPR09,rohrbach10cvpr,berg10eccv}.
In terms of detection, approaches range from shape-based object class
representations relying on manually-annotated parts in a single
view~\cite{stark09iccv} over HOG- (\cite{dalal05cvpr}) templates with
fixed layout~\cite{aytar11iccv} to full-fledged deformable part
models~\cite{gao12eccv} that share low-level feature
statistics. Common to all these approaches is that they are agnostic
about the inherent 3D nature of the object class detection problem; in
contrast, our models explicitly leverage 3D
(Sect.~\ref{sec:approach_model_one}) or viewpoint
(Sect.~\ref{sec:approach_model_two}) information. This is a key
difference in particular to~\cite{gao12eccv}, which focuses entirely
on low-level features that can be universally transferred.
%
Since part of our evaluation is performed on object
categories of a fine granularity (individual car-models), we
acknowledge that this is of course connected to work in fine-grained
categorization~\cite{yao11cvpr,farrell11iccv,stark12bmvc}. In contrast
to these works, however, the focus of ours is on learning multi-view
detectors from scarce training data, rather than classifying cropped
images according to a fine-grained class taxonomy.

\section{Multi-view transfer learning}
\label{sec:approach}

We consider the scenario of transfer learning for object models. The
goal is to train an object detection model for a 
\target class for which only very few labeled examples are
available. However we have access to an existing (or several) object
model for a similar (or the same) object class, the 
\source models. The main intuition that guides our approach is that if we
extract common regularities shared by both object classes, then this
in turn can be used to devise better \target models. In the case of
object detection on HOG~\cite{dalal05cvpr} we reason that although the
actual feature distribution may differ, there are similarities
in how the features deform under transformations such as viewpoint
changes.

\myparagraph{Preliminaries.} 
 More
formally, let us denote by $\sourcew$ the parameters of a \source
model. Specifically in the case of multi component detectors we have
$\sourcew =[w_1^s,\ldots,w_C^s]$, where the individual $w_i^s$ denote
different components of the models. As we are interested in the
multi-view setting, the components represent specific viewpoints in our
case. Given $\sourcew$ and a few $N_t$ labeled examples of a \target
class $\{x_i,y_i; i\in\{1,\ldots,N_t\}\}$, the goal is to derive a
detection model $\targetw$. This is implemented via the regularized
risk functional, which has been used for multi-task learning in \cite{Evgeniou:2005:LMT:1046920.1088693}
\begin{equation}
  \targetw = \argmin_{\w} J(\w) + \sum_{i=1}^{N_t}l(\w, x_i, y_i),
  \label{eq:svmprior}
\end{equation}
consisting of a regularization $J(\w)$ and a data fit term,
here the empirical loss $l$ on the training data
points. The data term is
standard and we may use loss functions 
such as structured losses or simpler losses like the Hinge loss for 
classification. In addition to the data term we regularize the model
parameters with $J$, that is derived using information from the
\source model. 
We use a transfer learning objective based
on~\cite{Evgeniou:2005:LMT:1046920.1088693} where the same regularizer
is proposed in the context of multi-task learning. The regularizer is
quadratic and of the form
\[J(\w) = \w^{\top}K_s\w.\] We distinguish between different
choices of $K_s$, implementing different possibilities to transfer
knowledge from the \source model. When $K_s=\mathbf{I}$, the
objective \eqref{eq:svmprior} reduces to the standard \noprior
case. In the following, we will explore three different variants for
the knowledge transfer matrix $K_s$.


\subsection{Learning sparse correlation structures}
\label{sec:approach_model_one}

Let us denote with $w$ an appearance filter of one viewpoint component
in the entire set of parameters $\w$. For simplicity we will simply
refer to this as $w$ without using sub- or superscripts. This filter
is of size $n\times m\times L$. It has spatial extent of
$n\times m$ cells, and $L$ are filter values computed in each
cell ($L=32$ in~\cite{felzenszwalb10pami}). We denote each cell $j$ as a vector
$w_j\in\mathbb{R}^L$. We implement different versions of the transfer
learning objective \eqref{eq:svmprior} using a graph Laplacian
regularization approach (\cite{Evgeniou:2005:LMT:1046920.1088693}, Sect 3.1.3)
by choosing 
\[
J(\w) = \w^{\top}K_s \w = \w^{\top} (I-\lambda\Sigma_s) \w,
\]
where $\Sigma_s$ encodes correlations between different cells in the
model. The matrix $\Sigma_s$ is of size $P \times P$, with $P$ being
the total number of model parameters. To distinguish between different
choices for the structure of $\Sigma_s$ we denote with $\sim_n$ a
relationship of type $n$ between two cells in $w$. With ``type'', for
example we can refer to a spatial relation among cells, such as
horizontal neighbors, vertical neighbors, etc. This defines a set of
cell pairs $\mathbb{P}_{n} = \{(w_j,w_k)|j\sim_n k\}$ in the model
that satisfy relation $\sim_n$. From the set $\mathbb{P}_{n}$ one can
compute cross covariances for different types
\begin{equation}
  \label{eq:cov}
  \Sigma_n = \sum_{j\sim_n k} (w_j-\bar{w})(w_k-\bar{w})^{\top},
\end{equation}
where $\bar{w}=\frac{1}{|\mathbb{P}_n|}\sum_j w_j$ is the mean of the
set of cells. The full $P\times P$ matrix $\Sigma_s$ is then
constructed from the smaller $L\times L$ block matrices $\Sigma_n$
(details below). This results in a sparse $\Sigma_s$, as the
number of cell pairs satisfying a relation is small compared to the
total number of possible cell pairs.

\myparagraph{Single view correlation structures (\SVcorr).}
Originally proposed in~\cite{gao12eccv}, \SVcorr aims at capturing
generic neighborhood relationships between HOG cells within a single
template (i.e., a single view). This implements a specific choice for
$\sim_n$. \SVcorr focuses on 5 specific relation types: horizontal
($\sim_h$), vertical ($\sim_v$), upper-left diagonal ($\sim_{d1}$) and
upper-right diagonal ($\sim_{d2}$) cell relations. In addition,
\SVcorr captures self-correlations of the same cell
$\sim_{cell}$. For a given relation type $\sim_n$, \SVcorr takes
into account all cell pairs in the template which satisfy the relation
$\sim_n$, to compute each of the different cross-covariances
$\Sigma_h$, $\Sigma_v$, $\Sigma_{d1}$, $\Sigma_{d2}$ and
$\Sigma_{cell}$.


\myparagraph{Multi-view correlation structures (\MVcorr).}  We extend
\SVcorr to encompass multi-view knowledge transfer. In our model
different components $w$ of $\w$ correspond to different viewpoints
of the object class. Different components are very related since they
 have a common cause in the geometric structure of the three
dimensional object. Therefore, the goal of \MVcorr is to capture the
across-view cell relations in addition to the single view cell
relation introduced by \SVcorr. For that purpose, we learn a new,
across-view relation type $\sim_{mv}$, capturing cell relationships
across different views.

In order to establish cell relationships across viewpoints, we use a
3D CAD model of the object class of interest (or a generic 3D
ellipsoid with proper aspect ratio in case we do not have a CAD model
available for a class), which provides a unique 3D reference frame
across views. The alignment between learned and 3D CAD models is
achieved by back-projecting 2D cell positions onto the 3D surface,
assuming known viewpoints for the learned models and fixed perspective
projection. 
%
%
We then establish cell relationships between cells that back-project
onto the same 3D surface patch in neighboring views, and learn
$\Sigma_{mv}$.

After computing the different cross-covariances $\Sigma_n$ for both
\SVcorr and \MVcorr from the \source models, we 
construct the $\Sigma_s$ matrix, which is further on used as a
regularizer in the \target model training
(Eq.~\ref{eq:svmprior}). $\Sigma_s$ uses the learned cell-cell
correlations of different types from the \source models, to guide the
training of the \target model. In order to
construct $\Sigma_s$, we first establish pairs of cells $(w_i, w_j)$
in the \target model which satisfy a certain relation type $\sim_n$
(e.g. neighbors across views) and then we populate the corresponding
entries in $\Sigma_s$, $\Sigma_s^{i,j}$ with $\Sigma_n$. We apply this
procedure for all cell relation types defined in \SVcorr and \MVcorr.

\subsection{Learning dense multi-view correlation structures (\MVcovmat)}
\label{sec:approach_model_two}

\MVcorr and \SVcorr capture correlation structures among model cells
that satisfy certain cell relations (2D neighboring cells, 3D object
surface) resulting in a sparse graph encoded by $\Sigma_s$. In the
following, we extend this limited structure to a dense graph, that
potentially captures relationships among all cells in the model. We
will refer to this model as \MVcovmat.

Let's assume we are given a set of $N$ source models
\textbf{$\{\w^s_1,\ldots,\w^s_N\}$}, for example by training several
models using bootstrapping. Then, we compute the
un-normalized covariance matrix $\Sigma_{emp}$
\begin{equation}\label{eq:mvcovmat}
\Sigma_{emp} = \frac{1}{N}\sum_{i=1}^N \textbf{$\w^s_i$}  {\textbf{$\w^s_i$}}^{\top}
\end{equation}
which is a rank $N$ matrix.  We set $\Sigma_s=\Sigma_{emp}$.



\begin{figure}[!t]
  \centering
\setlength{\tabcolsep}{3pt}
\begin{center}
  \begin{tabular}{ccc|c}
    \includegraphics[height=1.2cm]{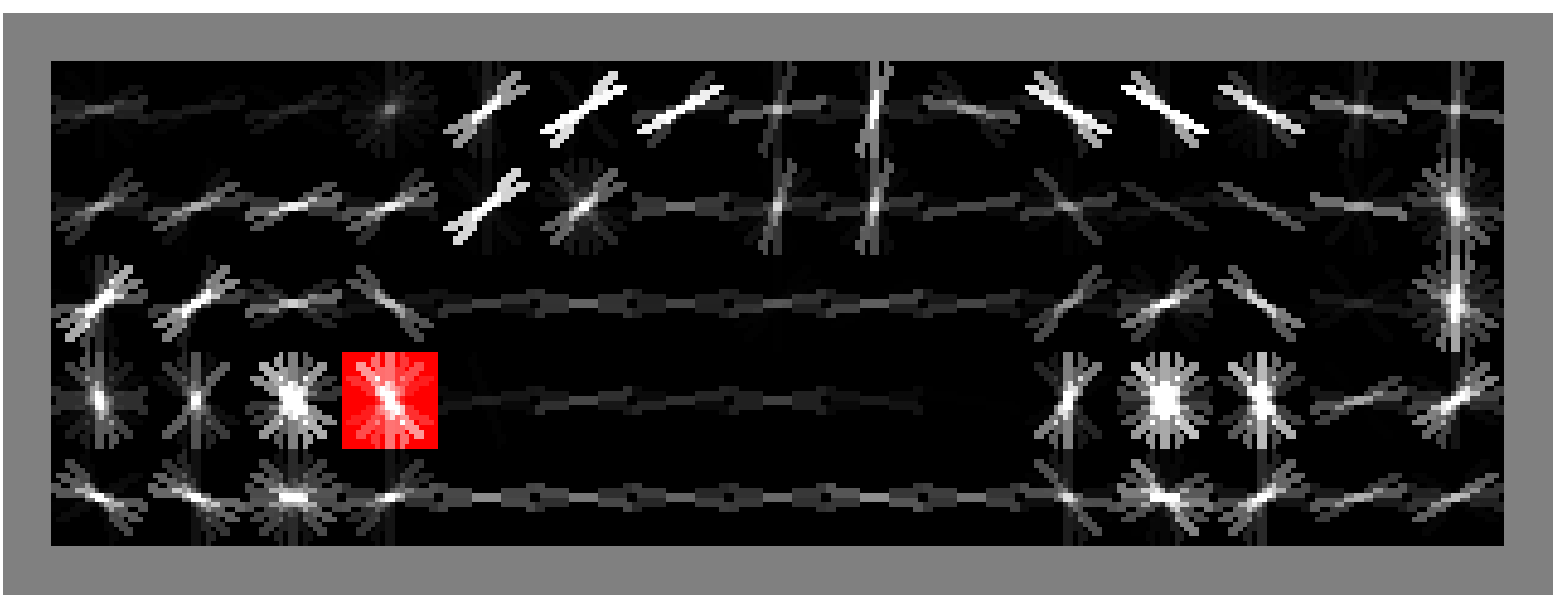}&
    \includegraphics[height=1.2cm]{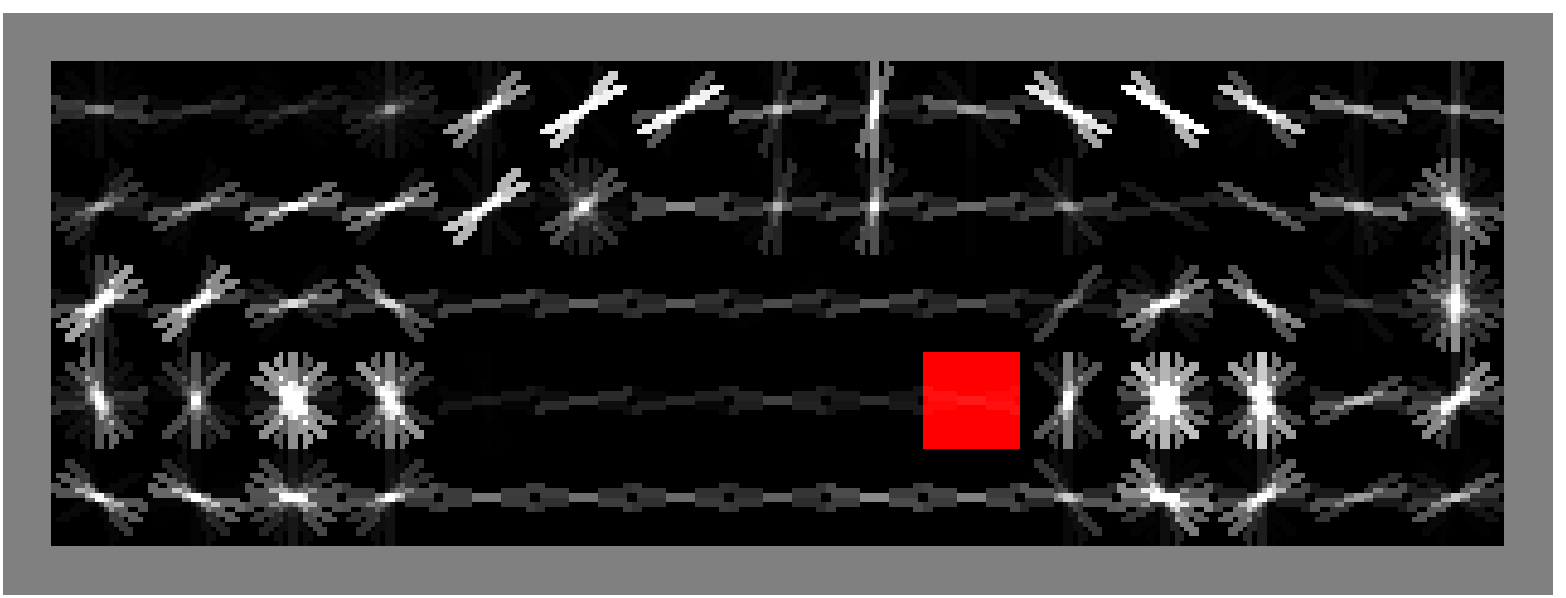}&
    \includegraphics[height=1.2cm]{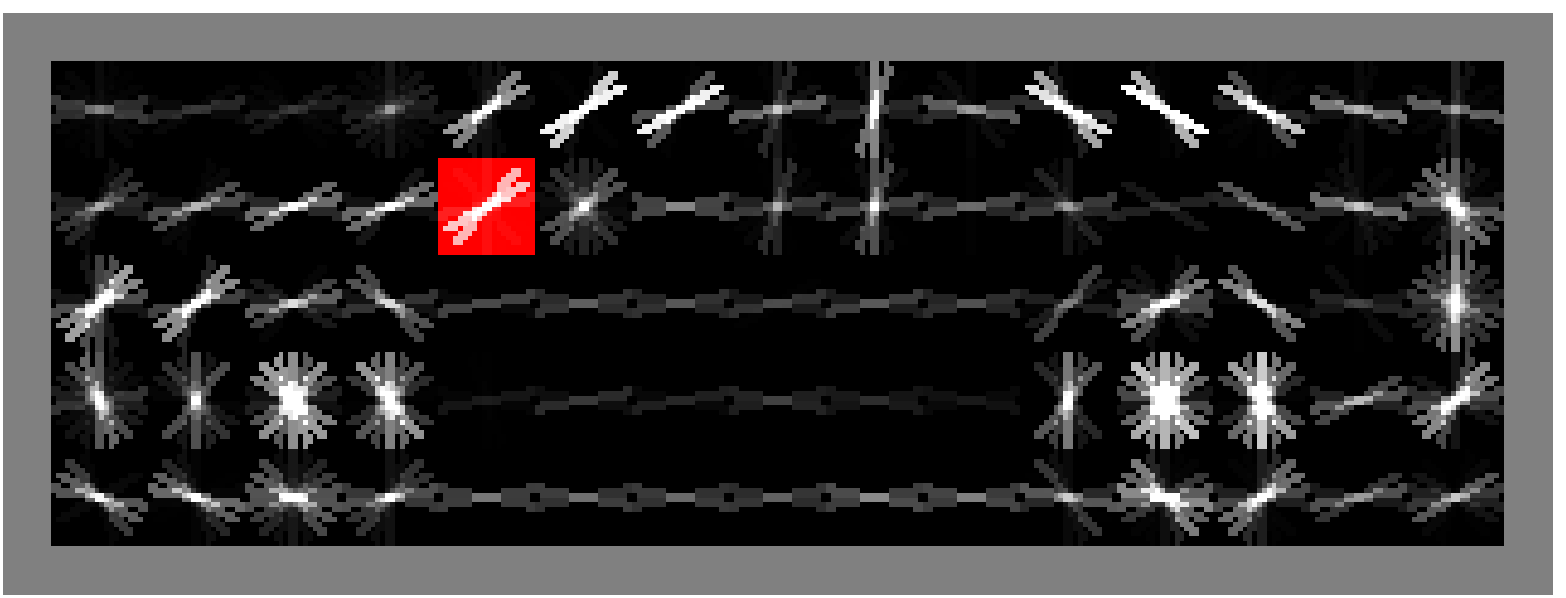}&
  \includegraphics[height=1.3cm]{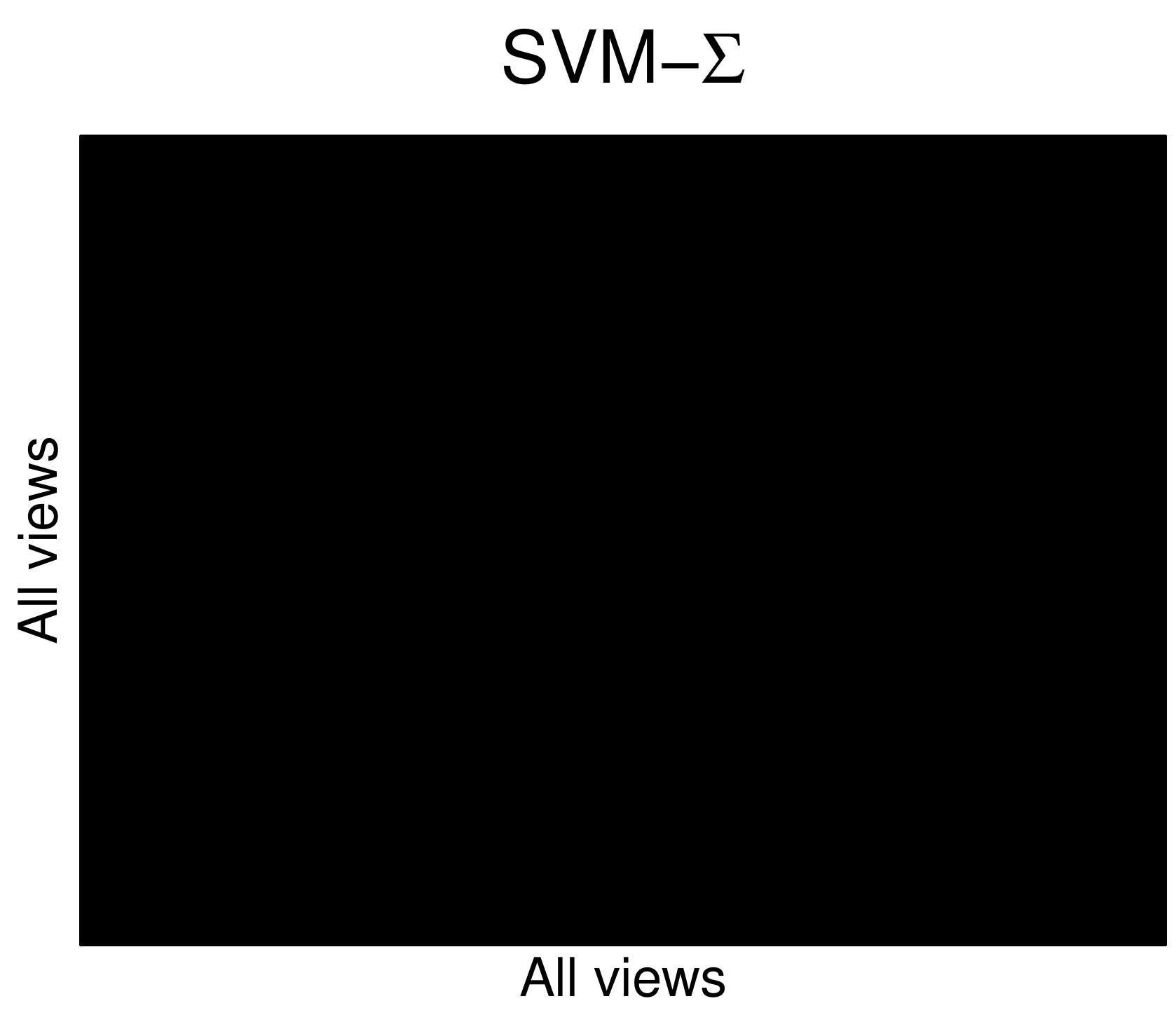}
  \includegraphics[height=1.3cm]{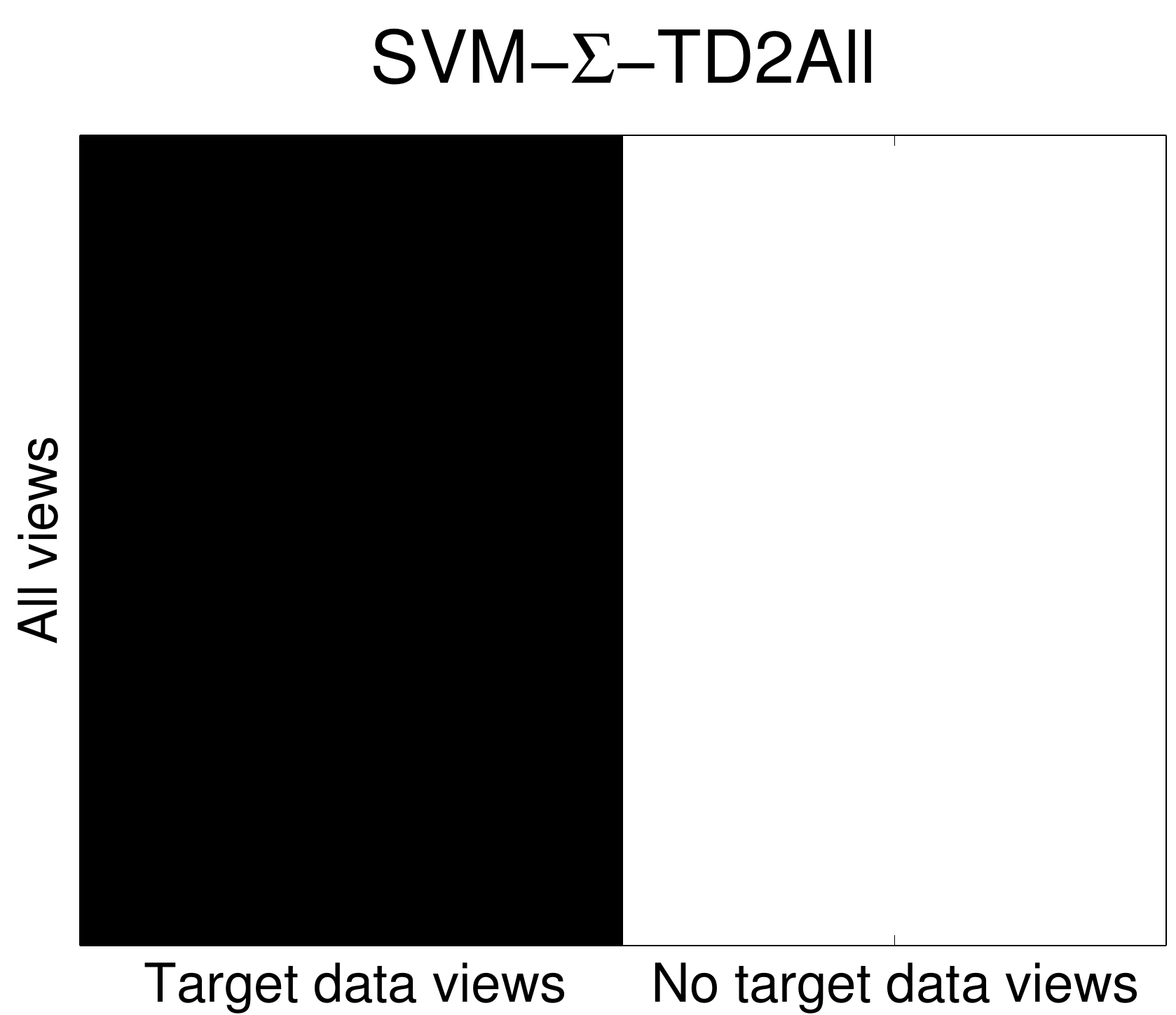}\\
    \includegraphics[height=1.2cm]{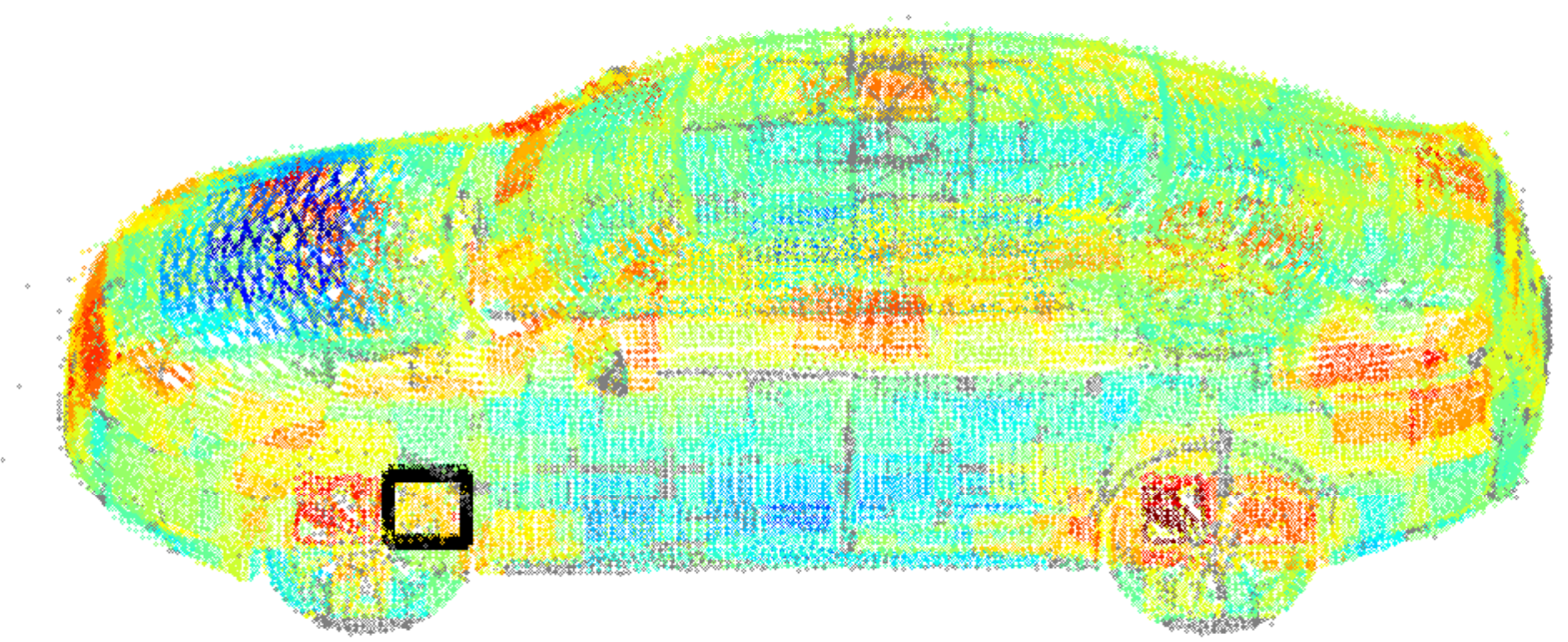}&
    \includegraphics[height=1.2cm]{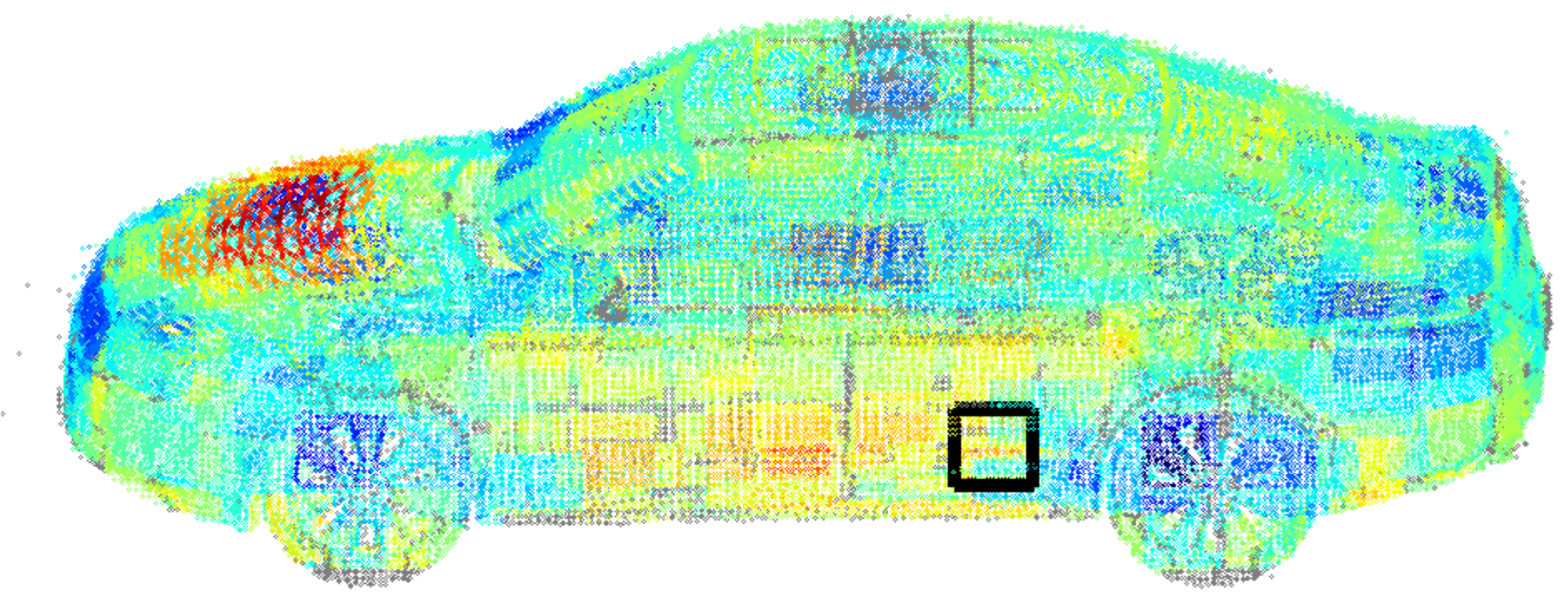}&
    \includegraphics[height=1.2cm]{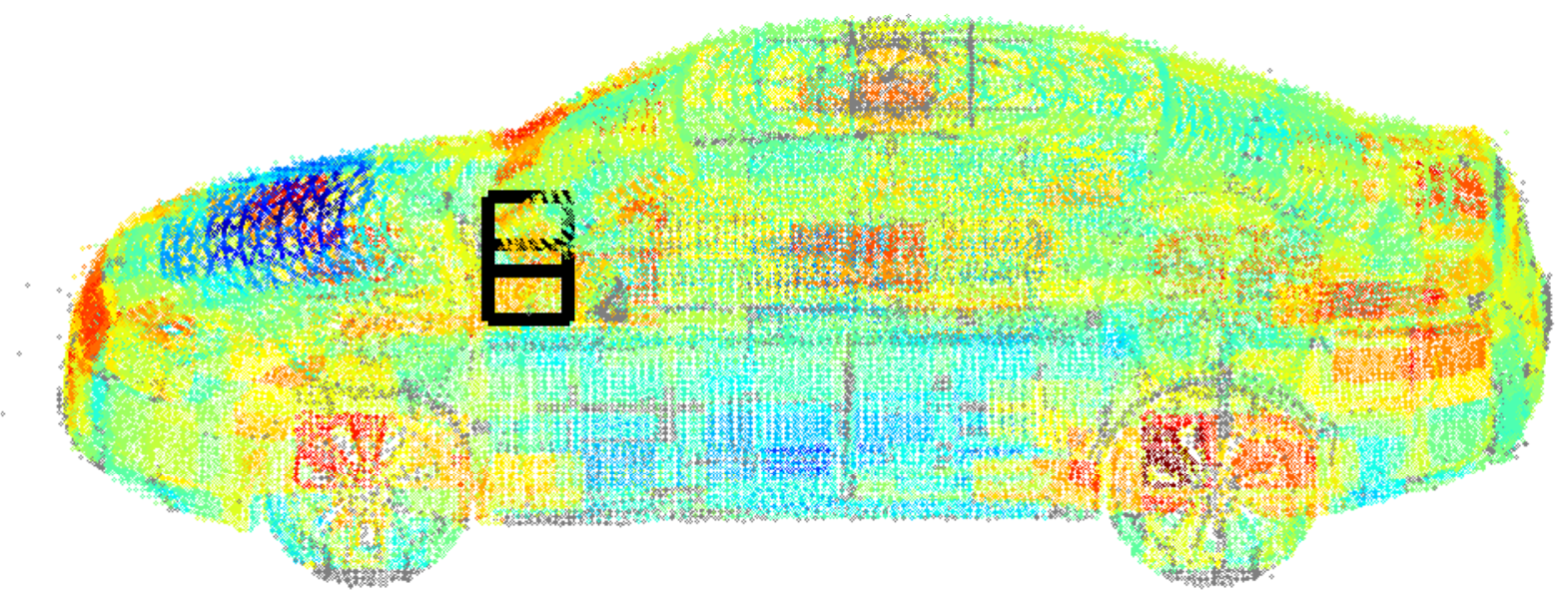}&
    \includegraphics[height=1.3cm]{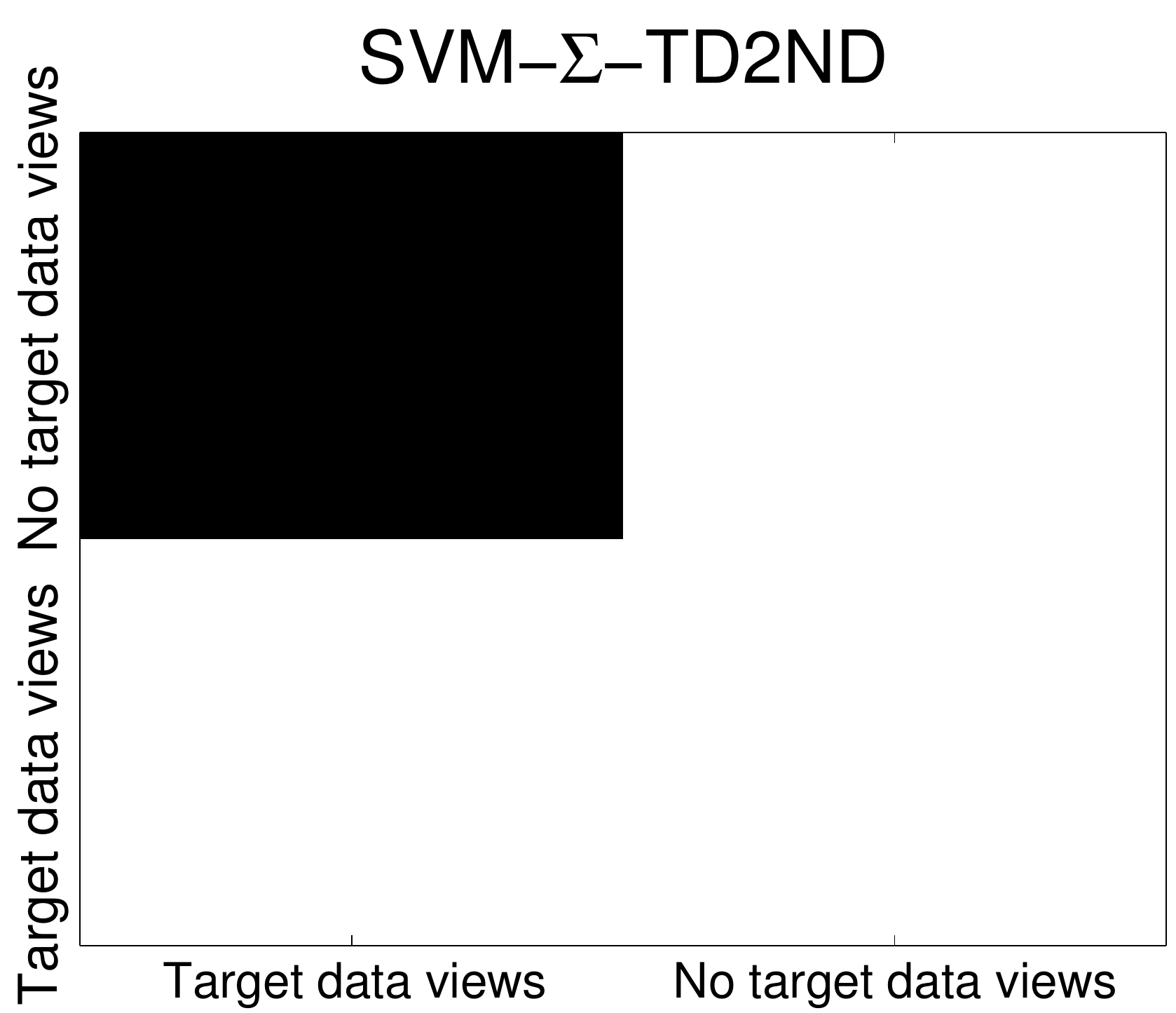}
    \includegraphics[height=1.3cm]{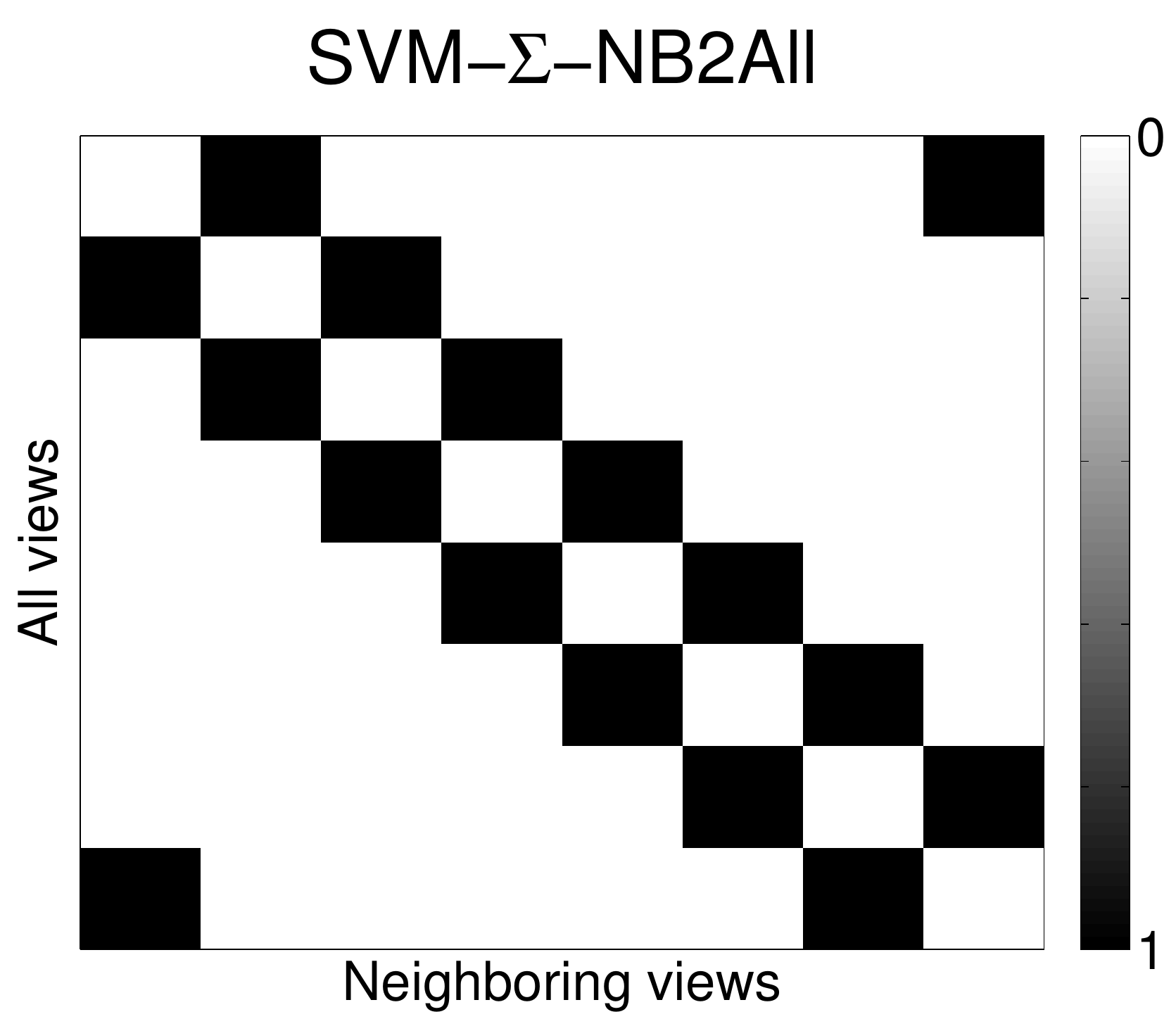}\\
  \end{tabular}
\end{center}
\caption{ (Left) Learned priors visualized in 3D (for a {\em reference}
    cell). Red indicates the {\em reference} cell. The black cube
  indicates the {\em reference} cell back-projected into 3D. (Right) \MVcovmat versions.}
  \label{fig:prior1}
\end{figure}

This variant of $\Sigma_s$ (\MVcovmat) is dense and captures
correlations of all types among all cells across all viewpoints in the
model. Unlike \MVcorr and \SVcorr, which are rather generic in nature
(e.g., all pairs of horizontal pairs in the template are considered
when learning $\Sigma_h$), \MVcovmat can capture very specific and local
cell correlations, within a single template (view) and across
views. Fig.~\ref{fig:prior1}~(left) visualizes a heatmap of the
strength of the learned correlations for \MVcovmat between given
reference cells (red squares, black cubes) and all other cells,
back-projected onto the 3D surface of a car CAD model. Note that the
heatmaps indeed reflect meaningful relationships (e.g., the front wheel
surface patch shows high correlation with back wheel patches).

While \MVcovmat is a symmetric prior (as the correlations are computed
across all views in the model), we also consider the case where the
\target training data distribution is sparse over viewpoints.
We address this by sparser, asymmetric variants of \MVcovmat that
connect only certain views with each other, by zeroing
out parts of the $\Sigma_s$ using an element-wise multiplication with
a sparse matrix $S$ as $S\circ\Sigma_s$. Several choices of $S$ are
depicted in Figure~\ref{fig:prior1} (right). We distinguish between
the following asymmetric priors: \MVcovmatTDtoND, where we transfer
knowledge from views for which we have \target data to views with no
\target training data, \MVcovmatTDtoALL with transfer from views with
\target data to all views, and \MVcovmatNBtoALL where we transfer from
every viewpoint to its neighboring viewpoints.

\subsection{Learning a \target model using the learned $K_s$ matrix}
We perform model learning (Eq.~\ref{eq:svmprior}) by first doing a
Cholesky decomposition of $K_s=U^{\top}U$.  This allows us to define
feature and model transformations: $\tx=U^{-\top}\x$ and
$\tw=U\w$. Using these transformations, one can show that
$\w^{\top}K_s\w=\tw^{\top}\tw$ and $\tw^{\top}\tx=\w^{\top}\x$, which
means we can learn a \target model by first, transforming the features
and the models using $U$, then training a model via a standard
\noprior solver in the transformed space, and in the end transforming
back the trained model.  In the \SVcorr case, to be compatible
with~\cite{gao12eccv}, we perform eigen decomposition instead of
Cholesky.

\section{Experiments}
\label{sec:exps}
In this section, we carefully evaluate the performance of our
multi-view priors.
First (Sect.~\ref{sec:exp_comparison}), we provide an in-depth
analysis of different variants of the \MVcorr and \MVcovmat priors in
a controlled training data setting, by varying the viewpoint
distribution of the training set. We learn \target models using a few
\target training examples plus our priors and compare them to using
the \SVcorr prior proposed in~\cite{gao12eccv} and using standard
\noprior. We perform the analysis on two tasks, 2D bounding box
localization and viewpoint estimation on the 3D Object Classes
dataset~\cite{savarese07iccv}, demonstrating successful knowledge
transfer even for cases in which there is no training data for $3/4$
of the viewing circle.
%
Second (Sect.~\ref{sec:exp_multiview_detection}), we highlight the
potential of our \MVcovmat priors to greatly improve the performance
of simultaneous 2D bounding box localization and viewpoint estimation
in a realistic, uncontrolled data set of challenging street scenes
(the tracking benchmark of KITTI~\cite{geiger12cvpr}).
%

For computational reasons, we restrict ourselves to the
root-template-only version of the DPM~\cite{voc-release4} as the basis
for all our models in Sect.~\ref{sec:exp_comparison}, but consider the
full, part-based version for the more challenging and realistic
experiments in Sect.~\ref{sec:exp_multiview_detection}. In all cases,
the $C$ parameter is fixed to $0.002$~\cite{felzenszwalb10pami} for all tested
methods. We set $\lambda = 0.9/e_{max}$,
where $e_{max}$ is the biggest eigenvalue of $\Sigma_s$. We
empirically verified that this always resulted in a positive definite
matrix $K_s$, which makes Eq.~\ref{eq:svmprior} a convex optimization
problem.

\subsection{Comparison of multi-view priors}
\label{sec:exp_comparison}
\begin{figure}
\centering
  \includegraphics[width=0.45\textwidth]{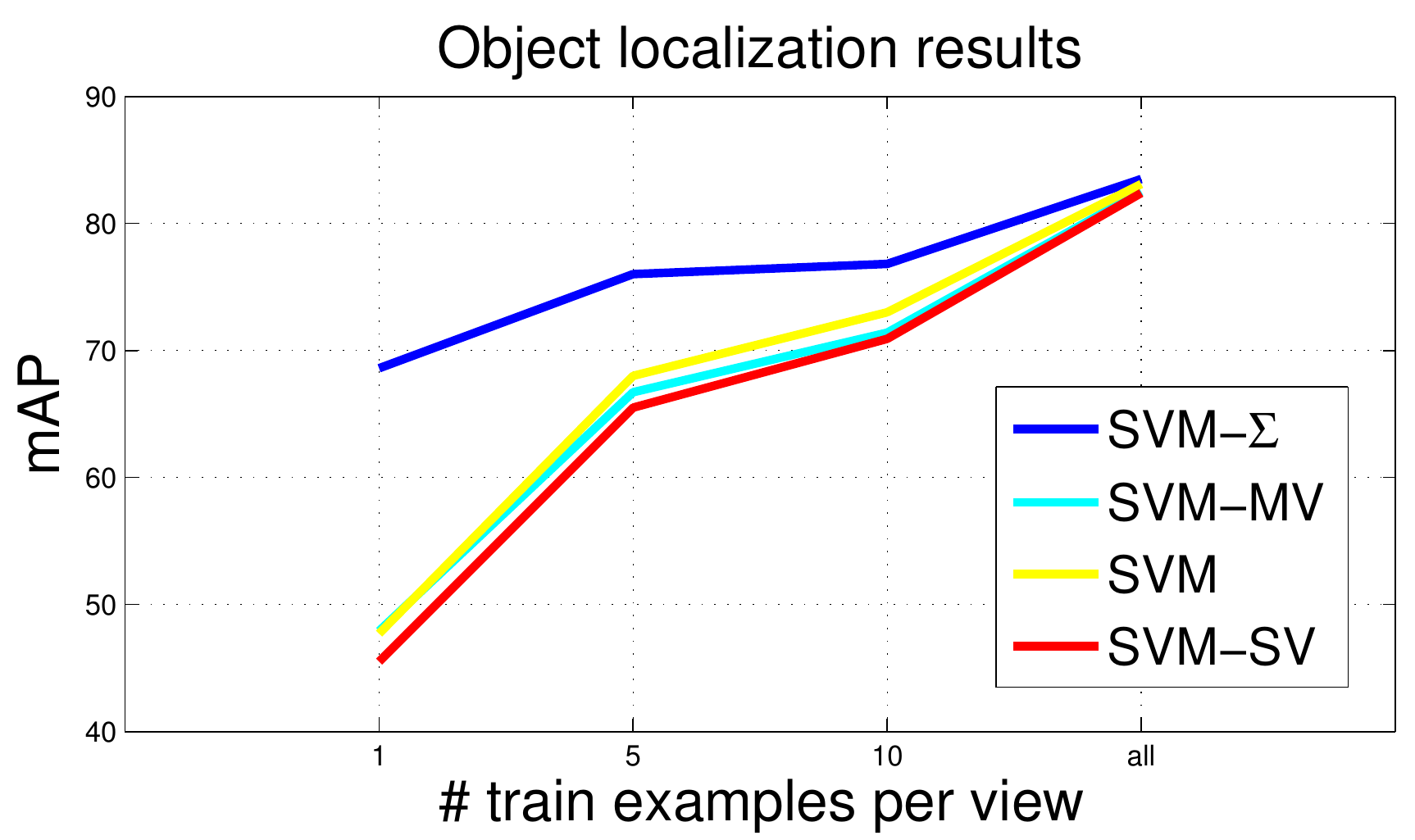}
  \includegraphics[width=0.45\textwidth]{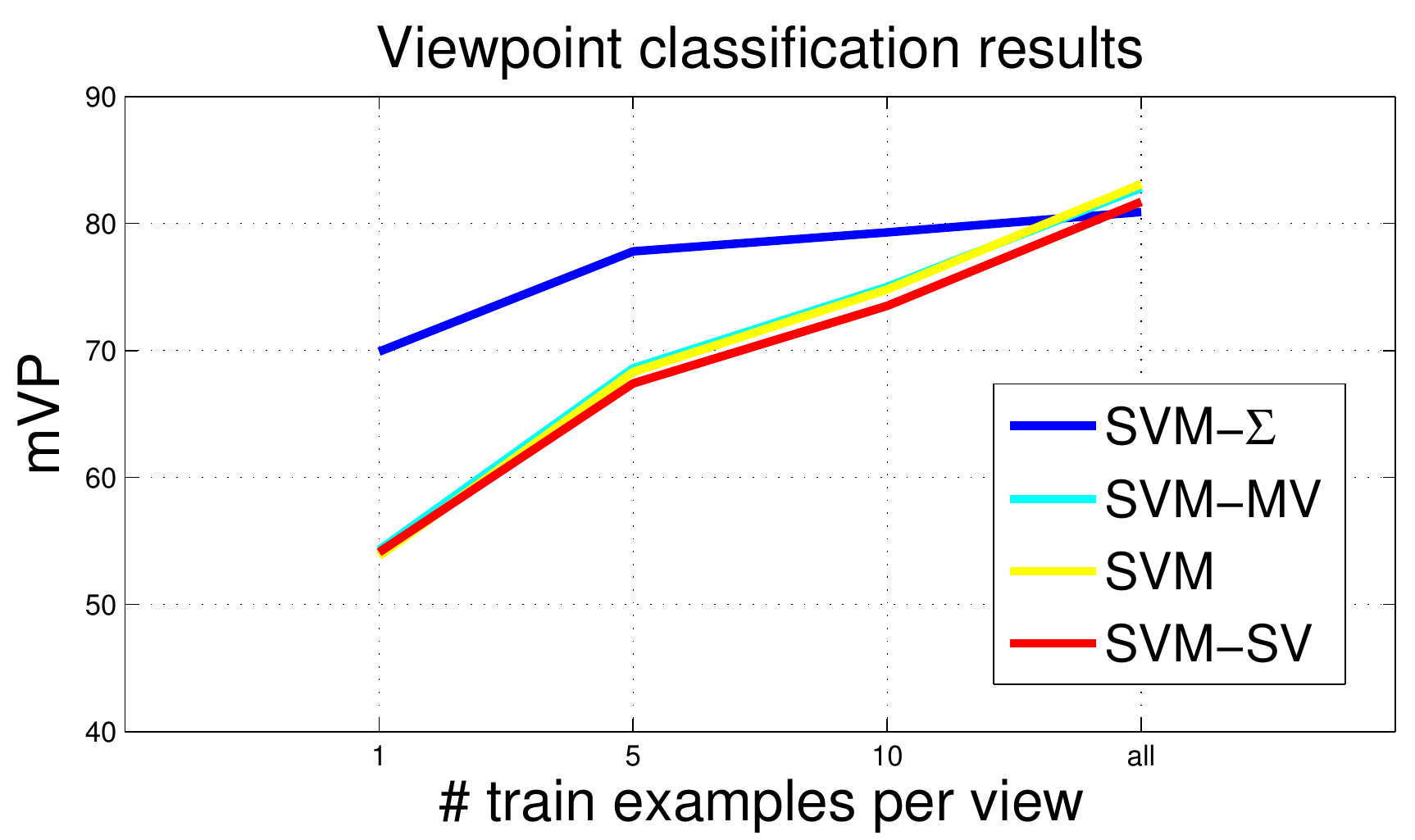}
  \caption{{\small 2D BB localization (left) and viewpoint estimation
     (right) on 3D Object Classes~\cite{savarese07iccv}.}}
  \label{fig:detresults}
\end{figure}
%
\begin{table}
\centering
\scriptsize
\setlength{\tabcolsep}{1pt}
  \begin{tabular}{|c|c|c|c|c||c|c|c|c||c|c|c|c|}
    \hline
              & {\scriptsize \SVcorr}& {\scriptsize \MVcorr} & {\scriptsize \MVcovmat} & {\tiny \noprior} & ~\cite{lopez11iccv} & ~\cite{gu10eccv} & $\begin{matrix}\text{{\tiny DPM-hinge+VP}}\\ \text{~\cite{pepik12cvpr}} \end{matrix}$ & $\begin{matrix}\text{{\tiny DPM-VOC+VP}}\\ \text{~\cite{pepik12cvpr}} \end{matrix}$ & \cite{liebelt10cvpr} & ~\cite{zia113drr} & ~\cite{payet11iccv} & ~\cite{glasner11iccv} \\
\hline
    car       & 99.6 / 92.9          & 99.8 / 95.0 & 99.8 / 92.5             & 99.8 / 95.0            & 96.0 / 89.0         & -/-              & 99.6 / 92.5                             & 99.8 / 97.5                           & 76.7/70.0            & 90.4/84.0         & -/86.1              & 99.2/85.3             \\
    bicycle   & 88.8 / 87.6          & 89.9 / 87.6 & 96.7 / 92.2             & 90.1 / 87.9            & 91.0 / 88.0         & -/-              & 98.6 / 93.0                             & 98.8 / 97.5                           & 69.8 / 75.5          & -/-               & -/80.8              & -/-                   \\
    iron      & 94.9 / 94.7          & 96.4 / 96.1 & 90.6 / 88.8             & 97.0 / 95.5            & 53.0 / -            & -/-              & 93.3 / 86.3                             & 96.0 / 89.7                           & -/-                  & -/-               & -/-                 & -/-                   \\
    cell.     & 51.0 / 82.2          & 51.2 / 82.3 & 53.7 / 80.9             & 51.1 / 81.5            & 43.0 / -            & -/-              & 62.9 / 65.4                             & 62.4 / 83.0                           & -/-                  & -/-               & -/-                 & -/-                   \\
    mouse     & 61.3 / 74.7          & 61.2 / 70.8 & 61.4 / 69.8             & 62.5 / 72.5            & 41.0 / -            & -/-              & 73.1 / 62.2                             & 72.7 / 76.3                           & -/-                  & -/-               & -/-                 & -/-                   \\
    shoe      & 93.9 / 81.4          & 93.4 / 87.3 & 94.7 / 86.2             & 94.8 / 86.5            & 78.0 / -            & -/-              & 97.9 / 71.0                             & 96.9 / 89.8                           & -/-                  & -/-               & -/-                 & -/-                   \\
    stapler   & 71.5 / 69.2          & 72.6 / 70.2 & 74.2 / 70.2             & 74.2 / 70.2            & 32.0 / -            & -/-              & 84.4 / 62.8                             & 83.7 / 81.2                           & -/-                  & -/-               & -/-                 & -/-                   \\
    toaster   & 94.4 / 70.6          & 95.3 / 72.8 & 97.2 / 66.7             & 95.2 / 75.6            & 54.0 / -            & -/-              & 96.0 / 50.0                             & 97.8 / 79.7                           & -/-                  & -/-               & -/-                 & -/-                   \\
\hline
 mAP          & 81.9 / 81.7          & 82.5 / 82.8 & 83.5 / 80.9             & 83.1 / 83.1            & 61.0 / 79.2         & - / 74.2         & 88.2 / 72.9                             & 88.5 / 86.8                           & -/-                  & -/-               & -/-                 & -/-                   \\
\hline
\end{tabular}
\caption{{\small Comparison to state-of-the-art on 3D Object
    Classes~\cite{savarese07iccv}.}}
\label{tab:3dobjects}
\end{table}
We start by comparing the different multi-view priors on the 3D Object
Classes dataset~\cite{savarese07iccv} (a widely accepted
multiview-benchmark with balanced training and test data from 8
viewpoint bins, for 9 object classes), 
in two sets of experiments. In the first set, we use the same
number $k$ of \target training examples per view (multi-view {\em k-shot} learning).
In the second set, we exclude certain viewpoints completely from the
training data ($k = 0$), keeping only a single example from each of the other
viewpoints 
(sparse multi-view  {\em k-shot} learning). In both
cases, the test set requires detecting objects seen {\em from the
 entire viewing circle}.
%
%
For each class, our priors are trained using bootstrapping, from 5 \source
models (each trained from 15 randomly sampled examples per view). The
final \target model for a class is obtained by using $k$ training
examples from that class plus the respective prior
.

\myparagraph{Multi-view {\em k-shot} learning.}
Fig.~\ref{fig:detresults} plots 2D BB localization (left)
and viewpoint estimation (right) performance for \noprior, \SVcorr,
\MVcorr, and \MVcovmat, varying the number $k \in \{1,5,10,all\}$ of
\target training examples per view, averaged over 5 randomized runs.
%
We make the following observations. First, we see that \MVcovmat
outperforms all other methods by significant margins for restricted
training data ($k \in \{1,5,10\}$), for both 2D BB localization (by
at least $20.1\%$, $8.0\%$ and $3.8\%$, 
respectively) and viewpoint
estimation (by $15.6\%$, $9.2\%$ and $4.3\%$ 
). Second, the benefit of
\MVcovmat increases with decreasing number of training examples,
saturating for $k = all$. And
third, \SVcorr~\cite{gao12eccv} and \MVcorr apparently fail
to convey viewpoint-related information beyond what can be learned
from the $k$ target examples alone, performing on par with \noprior.

As a sanity check, Tab.~\ref{tab:3dobjects} relates the complete
per-class results for all methods and $k = all$ (rightmost curve
points in Fig.~\ref{fig:detresults}) to the state-of-the-art. Despite
not using parts, our models in fact outperform previously reported
results~\cite{gu10eccv,liebelt10cvpr,zia113drr,lopez11iccv,payet11iccv,glasner11iccv}
with and without priors, except for ~\cite{pepik12cvpr}
based on DPM~\cite{felzenszwalb10pami} with parts. As parts
obviously improve performance, we add them in
Sect.~\ref{sec:exp_multiview_detection}.

\begin{figure}
\centering
  \includegraphics[width=\textwidth]{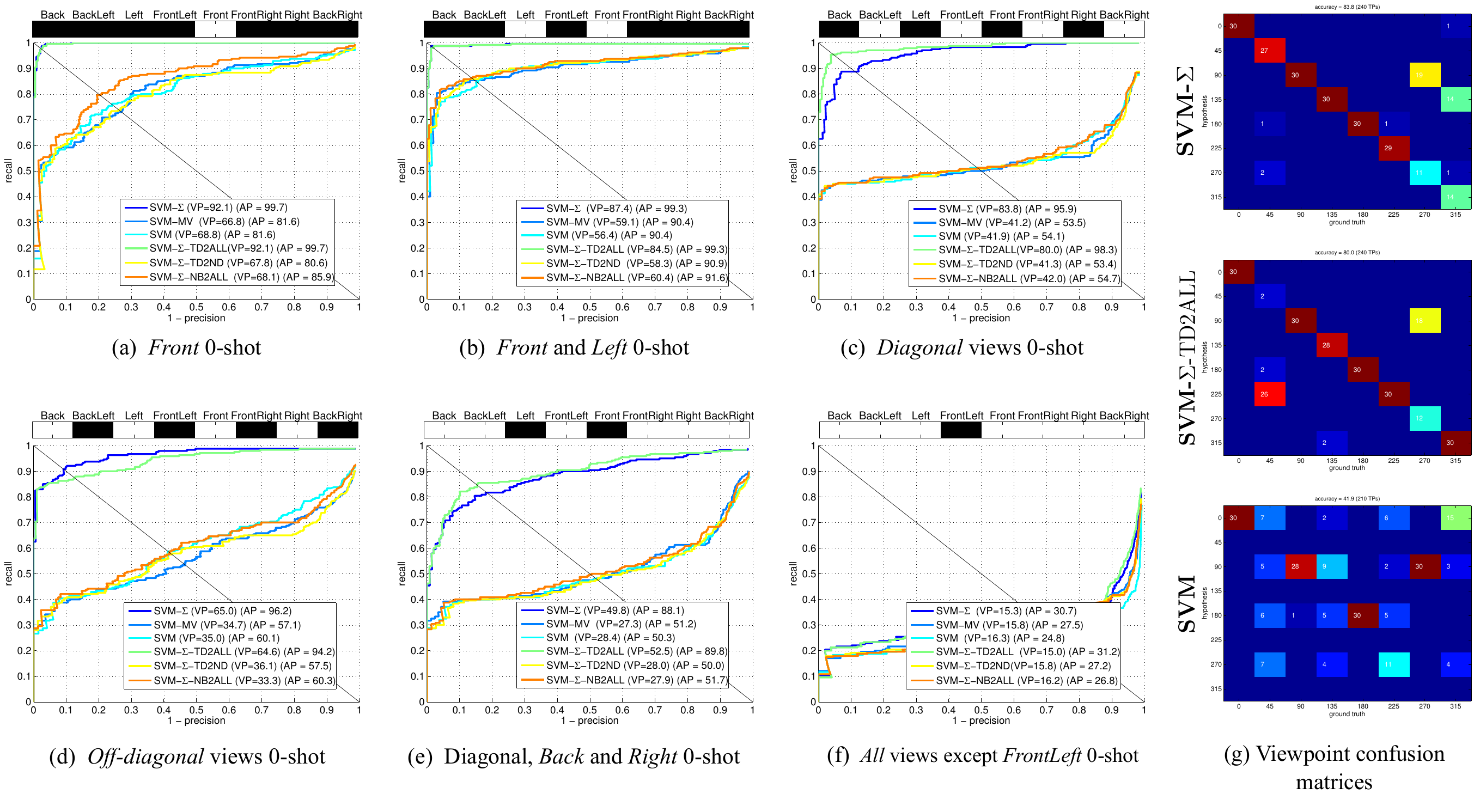}
  \caption{{\footnotesize 3D Object
      Classes~\cite{savarese07iccv}. Unbalanced multi-view {\em
        0-shot} experiments (on {\em cars}) with no training data for
      (a) {\em Front}, (b) {\em Front} and Left, (c) {\em
        Off-diagonal} views, (d) {\em Diagonal} views, (e) 1 training
      example for Left and {\em Front} views, (f) 1 example for {\em
        Front-left} view.  (g) VP confusion matrices for the 0-shot
      {\em Diagonal} case. Bars on top indicate (with black) which viewpoints are
      used in training for each experiment.}}
  \label{fig:unbalanced_results}
\end{figure}
\myparagraph{Sparse multi-view {\em k-shot} learning.}
We move on to a more challenging setting in which (single) training
examples are only available for selected views, but not for
others. Successful localization and viewpoint estimation thus depends
on prior information that can be ``filled in'' for the missing
viewpoints. Fig.~\ref{fig:unbalanced_results} plots precision-recall
curves for the car class and six
different settings of increasing difficulty, not having training data
for just one view (front) (a), not for two views (front and
left) (b), not for four views (diagonal) (c), off-diagonal (d), not
for six views (diagonal, back and right) (e), and not for all views
except front-left (f). Average precision and viewpoint
estimation results are given in plot legends. We compare the
performance of \noprior,
\MVcorr, \MVcovmat, and three further variations of \MVcovmat that
restrict the structure of the prior covariance matrix (Sect.~\ref{sec:approach_model_two}),
namely, \MVcovmat-TD2ALL, \MVcovmat-TD2ND, and \MVcovmat-NB2ALL.

In Fig.~\ref{fig:unbalanced_results}~(a) to (f), we observe that two
methods succeed in transferring information to up to 6 unseen viewpoints 
(\MVcovmat, dark blue, and \MVcovmat-TD2ALL, green), with APs
ranging from an impressive $99.7$ to $88.1\%$ and VPs ranging from $92.1$
to $49.8\%$ for \MVcovmat). This observation is confirmed by the
confusion matrices in Fig.~\ref{fig:unbalanced_results}~(g): both
\MVcovmat and \MVcovmat-TD2ALL exhibit a much stronger diagonal
structure than \noprior. Understandably, performance deteriorates
for just one observable viewpoint
(Fig.~\ref{fig:unbalanced_results}~(f); AP drops to $30.7\%$, VP to
$15.3\%$). \MVcorr (light blue) provides  an advantage over
\noprior (cyan) only for extremely little data
(Fig.~\ref{fig:unbalanced_results}~(e), (f)), where it improves AP by
$0.9\%$ and $2.7\%$.

\myparagraph{Summary.}
We conclude that different kinds of priors (\SVcorr, \MVcorr, and
variations of \MVcovmat) vary drastically in their ability to convey
viewpoint-related information. Notably, we observe only minor
differences between \SVcorr, \MVcorr, and \noprior, but large gains in
both 2D bounding box localization and viewpoint estimation for
\MVcovmat.

\subsection{Leveraging multi-view priors for object detection}
\label{sec:exp_multiview_detection}
Having verified the ability of our \MVcovmat priors to transfer
viewpoint information for scarce and unbalanced training data in
Sect.~\ref{sec:exp_comparison}, we now move on to an actual, realistic
setting, which naturally exhibits the dataset statistics that we
simulated earlier (see Fig.~\ref{fig:carstat},~\ref{fig:cartypesstat}
and~\ref{fig:carmodelstat}). Specifically, we focus on the {\em car}
object class on the KITTI street scene dataset~\cite{geiger12cvpr}
(and the tracking benchmark subset),
consisting of 21 sequences (8,008 images, corresponding to $579$ car
tracks and $27,300$ annotated car bounding boxes) taken from a driving
vehicle. We use $5$ sequences for training and the rest for testing.
%
Due to the car-mounted camera setup, the distribution of viewpoints
for car objects is already heavily skewed towards back and diagonal
views (cars driving in front of the camera car or being parked on the
side of the road, see Fig.~\ref{fig:carstat}). This becomes even more severe when considering
more fine-grained categories, such as individual {\em car-types} (we
distinguish and annotate $7$: stat. wagon, convertible, coupe, hatchback,
minibus, sedan, suv) and {\em car-models} ($23$ in total)
\footnote{The annotations will be made publicly
 available upon publication.}.

\myparagraph{Evaluation criteria.} Average precision (AP) computed
using the Pascal VOC~\cite{everingham06voc} overlap criterion, based
solely on bounding boxes (BB) has been widely used as an evaluation
measure. Since the ultimate goal of our approach is to enable
simultaneous object localization and viewpoint estimation (both are
equally important in an autonomous driving scenario), and in line with~\cite{geiger12cvpr}, we report performance for two combined measures
(jointly addressing both tasks) in addition to AP and
VP. Specifically, \apvpd allows a detection $\hat{y}$ to be a true
positive detection if and only if the viewpoint estimate $\hat{y}^v$
is the same as the ground truth $y^v$. The second measure, \apvpc
assigns a weight $\hat{w}=(180^{\circ}-|\angle(\hat{y}^v, y^v)|) /
180^{\circ}$ to the true positive detection based on how well it
aligns with the ground truth viewpoint. Also in line
with~\cite{geiger12cvpr}, we report results for non-occluded objects.




\begin{table}
  \centering
  \scriptsize
  \setlength{\tabcolsep}{3pt}
  \begin{tabular}{cc}
      \begin{tabular}{|c|c|c|c|c|}
        \hline
        \multicolumn{2}{|c|}{prior}            & \multicolumn{3}{c|}{AP / VP}                                  \\
        \hline
        method                                 & dataset        &           {\em base} & {\em car-type} & {\em car-model} \\
        \hline
        \multicolumn{2}{|c|}{{\scriptsize \noprior (KITTI+ 3D obj.)}} & 87.1 / 69.3           & - / -                 & - / -                \\        
        \multicolumn{2}{|c|}{{\scriptsize \noprior (KITTI)}}          & 86.6 / 68.7           & 88.7 / 70.9          & 83.3 / 62.0           \\
        \hline                                                                                                                  
        \MVcovmat                              & 3D objects           & 90.7 / 71.9           & \textbf{91.6} / 75.1 & 87.5 / 73.9           \\
        \MVcovmat                              & KITTI                & 90.7 / 71.9           & 90.1 / 75.1          & 89.4 / \textbf{75.6}  \\
        \MVcorr                                & 3D objects            & 90.2 / 72.6           & 90.3 / 71.9          & 82.9 / 63.2           \\
        \MVcorr                                & KITTI                & 89.2 / 73.1           & 88.5 / 71.1          & 76.5 / 66.5           \\
        \SVcorr                                & 3D objects            & 90.7 / 71.9           & 86.5 / 70.4          & 76.6 / 65.8           \\
        \SVcorr                                & KITTI                & 86.9 / 71.4           & 85.8 / 70.8          & 76.5 / 66.5           \\
        \hline
      \end{tabular}
      & 
      \begin{tabular}{|c|c|c|}
        \hline
         \multicolumn{3}{|c|}{\apvpd~/ \apvpc}   \\
         \hline
 
        {\em base}       & {\em car-type}                 & {\em car-model}                                                                                                                                                                                        \\
         \hline
         53.6 / 67.0     & - / -                          & - / -                                                                                                                                                                                                  \\
         53.3 / 65.8     & 58.1 / 67.9                    & 40.3 / 55.1                                                                                                                                                                                            \\
         \hline
         61.5 / 70.1     & 65.2 / \textbf{74.1}           & 60.9 / 70.7                                                                                                                                                                                            \\
         61.6 / 70.2     & \textbf{66.1} / 73.5           & 65.2 / 73.4                                                                                                                                                                                            \\
         60.9 / 69.9     & 60.8 / 70.0                    & 41.5 / 55.8                                                                                                                                                                                            \\
         62.1 / 69.8     & 58.8 / 67.6                    & 44.8 / 53.5                                                                                                                                                                                            \\
         61.5 / 70.1     & 55.9 / 65.8                    & 44.3 / 53.1                                                                                                                                                                                            \\
         59.6 / 67.0     & 56.5 / 65.1                    & 44.8 / 53.5                                                                                                                                                                                            \\
         \hline
       \end{tabular}
                                                                                                                                                                                                                                                                   \\
     \end{tabular}
     \caption{{\small Multi-view detection results on KITTI~\cite{geiger12cvpr}.}}
\label{tab:kittifinal}
\end{table}
\myparagraph{Basic-level category transfer.}
We commence by applying our priors to a standard object class detector
setup, in which a detector is trained such that positive examples are
annotated on the level of basic-level categories (i.e., {\em car}),
denoted {\em base} in the following. Tab.~\ref{tab:kittifinal}~(left)
gives the corresponding 2D bounding box localization and viewpoint
estimation results, comparing our priors \MVcorr and \MVcovmat to
\SVcorr and a baseline not using any prior (\noprior). For each, we consider two
variants depending from which data the prior (or the detector
itself for \noprior) has been trained (KITTI, 3D Object
Classes, or both). Note that the respective prior and \noprior
variants use the exact same training data (but in different ways) and
are hence directly comparable in terms of performance.

In Tab.~\ref{tab:kittifinal}~(left, col. {\em base}), we observe that
our priors \MVcorr and \MVcovmat consistently outperform \noprior, for
both 2D BB localization and viewpoint estimation, for both choices of
training data (e.g., \MVcovmat -KITTI with $90.7\%$ AP and $71.9\%$
VP vs. \noprior -KITTI with $86.6\%$ AP and  $68.7\%$ VP). The
performance difference is even more pronounced when considering the
combined performance measures (Tab.~\ref{tab:kittifinal}~(right,
col. {\em base})). \MVcovmat -KITTI achieves $61.6\%$ \apvpd
and $70.2\%$ \apvpc, outperforming \noprior -KITTI ($53.3\%$, $65.8\%$) by
a significant margin. 

Similarly, \MVcovmat -3D Object Classes outperforms \noprior
-KITTI+3D Object Classes in all measures ($90.7\%$ vs. $87.1\%$ AP, $71.9\%$
vs. $69.3\%$ VP, $61.5\%$ vs. $53.6\%$ \apvpd and $70.1$ vs. $67.0\%$
\apvpc). \MVcorr and \SVcorr priors also show promising detection performance,
outperforming the \noprior models in all metrics.

\myparagraph{Fine-grained category transfer.}
Recently, is has been shown that fine-grained object class
representations on the level of sub-categories can improve
performance~\cite{lan13iccv,hoai13cvpr,stark12bmvc}, since they better
capture the different modes of intra-class variation than
representations that equalize training examples on the level of
basic-level categories. Further, these representations lend themselves
to generate additional output in the form of fine-grained category
labels that can be useful for higher-level tasks, such as scene
understanding. In the following, we hence consider two fine-grained
object class representations that decompose {\em cars} into distinct
{\em car-types} or even individual {\em car-models}. Both are
implemented as a bank of multi-view detectors (one per fine-grained
category) that are trained independently, but combined at test time by
a joint non-maxima suppression to yield basic-level category
detections.

Note that the individual fine-grained detectors suffer
even more severely from scarce and unbalanced training data (see Fig.~\ref{fig:cartypesstat} and~\ref{fig:carmodelstat}
  in Sect.~\ref{sec:suppl}) than on
the basic-level (see Fig.~\ref{fig:carstat}) -- this is where our priors come into
play: we train the priors, as before, on the {\em base} level, and use
them to facilitate the learning of each individual fine-grained
detector, effectively transferring knowledge from {\em base} to
fine-grained categories.

Tab.~\ref{tab:kittifinal} gives the corresponding results
in columns {\em car-type} and {\em car-model}, respectively. We
observe: first, performance can in fact improve as a result
of the more fine-grained representation, for both \MVcorr, \MVcovmat
and even \noprior (\noprior-KITTI-car-type
improves AP from $86.6\%$ to $88.7\%$, and VP from $68.7\%$ to
$70.9\%$, and \apvpd from $53.3\%$ to $58.1\%$ and \apvpc from
$65.8\%$ to $67.9\%$ compared to \noprior-KITTI-base). A similar boost
in performance in viewpoint estimation and combined can be seen for
\MVcovmat (\MVcovmat-KITTI-car-type improves VP from $71.9\%$ to
$75.1\%$, and \apvpd from $61.6\%$ to $66.1\%$ and \apvpc from
$70.2\%$ to $73.5\%$ compared to \MVcovmat-KITTI-base; the AP stays
consistently high with $90.7\%$ vs. $90.1\%$).
Second, the level of granularity can be too fine: for almost all
methods, the performance of the fine-grained {\em car-model} drops
below the performance of the corresponding {\em base} detector --
there is just so little training data for each of the car models that
reliable fine-grained detectors can hardly be learned. Curiously,
\MVcovmat-KITTI-car-model can still keep up in terms of localization
($89.4\%$ AP) and even obtains the overall best VP accuracy of
$75.6\%$, which is also reflected in the combined measures ($65.2\%$
\apvpd, $73.4\%$ \apvpc).
Third, \MVcovmat-KITTI-car-type is the overall best method,
outperforming the original baseline \noprior-KITTI-base by impressive
margins, in particular for the combined measures ($90.1\%$
vs. $86.6\%$ AP, $75.1\%$ vs. $68.7\%$ VP, $66.1\%$ vs. $53.3\%$
\apvpd, $73.5\%$ vs. $65.8\%$ \apvpc).
\begin{table}[t]
\centering
\scriptsize
\setlength{\tabcolsep}{3pt}
\begin{tabular}{|c|c|c|c|c|c||c|c|c|c|}
\hline
\multicolumn{2}{|c|}{} & \multicolumn{4}{|c||}{@50 iou} & \multicolumn{4}{|c|}{@70 iou}                                                                                                              \\
\hline
                       &                           & \multicolumn{2}{|c|}{{\scriptsize AP / VP}} & \multicolumn{2}{|c||}{{\scriptsize \apvpd~/ \apvpc}}  & \multicolumn{2}{|c|}{{\scriptsize AP / VP}} & \multicolumn{2}{|c|}{{\scriptsize \apvpd~/ \apvpc}}  \\
\hline

   prior  & dataset & {\scriptsize {\em base}} & {\scriptsize {\em car-type}} & {\scriptsize {\em base}} & {\scriptsize {\em car-type}} & {\scriptsize {\em base}} & {\scriptsize {\em car-type}} & {\scriptsize {\em base}} & {\scriptsize {\em car-type}} \\
 \hline
\multicolumn{2}{|c|}{\noprior(KITTI)} & 90.9 / 74.3              & 93.2 / 75.9                  & 65.2 / 72.1              & 66.8   / 74.9                & 49.9 / 74.2              & 60.0 / 76.8                  & 37.5 / 40.4              & 44.6 / 48.3                  \\
\hline
\MVcovmat & 3D obj. & \textbf{94.8} / 78.6     & 93.4 / \textbf{81.7}         & 72.1 /  78.7             & \textbf{73.0   / 80.6}       & 51.5 / 81.2              & \textbf{64.7 / 83.9}         & 41.9 / 44.3              & \textbf{53.0 / 56.7}         \\
\MVcovmat & KITTI   & 94.8 / 77.2              & 94.3 / 78.3                  & 70.4 / 77.3              & 70.4   / 79.6                & 49.7 / 79.0              & 61.2 / 80.4                  & 39.5 / 41.8              & 47.9 / 53.3                  \\
\hline
\end{tabular}
\caption{{\small Multi-view detection results on KITTI~\cite{geiger12cvpr}. Models have root and 4 parts  per view.}}
\label{tab:kittiparts}
\end{table}

Tab.~\ref{tab:kittiparts}~(left) gives the  results for the best
performing priors of Tab.~\ref{tab:kittifinal} (\MVcovmat-KITTI,
\MVcovmat-3D Object Classes) in comparison to \noprior-KITTI, now 
using parts. As expected, parts result in a general
performance boost for all methods (around $5\%$ for all measures). The
benefit of our priors remains, for both granularity levels {\em base}
and {\em car-type}, in particular for the combined measures:
\MVcovmat-KITTI-base outperforms \noprior-KITTI-base by similarly
large margins as for the no-parts case ($70.4\%$ vs. $65.2\%$ \apvpd,
$77.3\%$ vs. $72.1\%$ \apvpc), and \MVcovmat-KITTI-car-type
outperforms \noprior-KITTI-car-types by ($70.4\%$ vs. $66.8\%$ \apvpd,
$79.6\%$ vs. $74.9\%$ \apvpc).

Tab.~\ref{tab:kittiparts}~(right) applies a tighter overlap
criterion for true positive detections ($0.7$ intersection over
union)~\cite{geiger12cvpr}. Interestingly, this leads to a larger
separation in performance between {\em base} and {\em car-type}
models, in particular in AP: e.g., \MVcovmat-3D Object Classes
improves from $51.5\%$ to $64.7\%$, \MVcovmat-KITTI from $49.7\%$ to
$61.2\%$ and \noprior-KITTI improves from $49.9\%$ to $60.0\%$,
highlighting the benefit of the fine-grained object class
representation in particular for highly precise detection.

\begin{table*}
\centering
\scriptsize
\setlength{\tabcolsep}{3pt}
  \begin{tabular}{|c|c|c|c|c|c|c|c||c|c|c|}
\hline
                & \multicolumn{7}{|c||}{without parts} & \multicolumn{3}{|c|}{with parts}                                                                                                                                                                                             \\
    \hline

    prior       & {\scriptsize \MVcovmat} & {\scriptsize \MVcovmat} & {\scriptsize \MVcorr} & {\scriptsize \MVcorr} & {\scriptsize \SVcorr} & {\scriptsize \SVcorr} & {\scriptsize \noprior} & {\scriptsize \MVcovmat} & {\scriptsize \MVcovmat} & {\scriptsize \noprior} \\
    \hline                                                                                                                                                                                                                                                                               
                & KITTI                     & 3D obj.                   & KITTI                  & 3D obj.                & KITTI                  & 3D obj.                & -                        & KITTI                     & 3D obj.                   & -                        \\
    \hline                                                                                                                                                                                                                                                                               
station wagon   & \textbf{71.2}             & 70.2                      & 64.5                   & 63.6                   & 62.6                   & 61.9                   & 61.9                     & \textbf{82.7}             & 81.9                      & 79.0                        \\
convertible     & \textbf{24.4}             & 24.0                      & 12.9                   & 10.8                   & 13.8                   & 11.7                   & 12.7                     & \textbf{50.7}             & 36.8                      & 12.0                         \\
     coupe      & \textbf{67.5}             & 67.1                      & 63.7                   & 67.0                   & 60.5                   & 57.7                   & 67.1                     & \textbf{79.9}             & 76.6                      & 76.5                         \\
 hatchback      & \textbf{89.8}             & 85.7                      & 66.4                   & 78.2                   & 58.9                   & 65.0                   & 71.0                     & \textbf{95.5}             & 88.0                      & 87.2                         \\
   minibus      & \textbf{31.3}             & 16.8                      & 20.0                   & 18.7                   & 16.3                   & 18.0                   & 18.6                     & \textbf{59.7}             & 42.0                      & 41.4                         \\
     sedan      & \textbf{69.4}             & 53.8                      & 46.7                   & 49.4                   & 37.8                   & 41.8                   & 48.7                     & \textbf{83.8}             & 79.8                      & 66.2                         \\
       suv      & \textbf{19.7}             & 14.7                      & 8.1                    & 7.3                    & 5.2                    & 8.0                    & 8.6                      & 34.5                      & \textbf{35.1}             & 16.4                         \\
\hline                                                                                                                                                                                                                                                                                   
       mAP      & \textbf{53.3}             & 47.5                      & 40.3                   & 42.1                   & 36.4                   & 37.7                   & 41.2                     & \textbf{69.5}             & 62.9                      & 54.1                      \\
\hline
\end{tabular}
\caption{{\small {\em Car-type} detection results on the KITTI~\cite{geiger12cvpr} dataset.}}
\label{tab:kittitypes}
\end{table*}
Lastly, we evaluate the performance of our fine-grained detectors on
the level of the respective fine-grained categories ({\em car-types}),
as independent detection tasks. Tab.~\ref{tab:kittitypes} gives the results without (left)
and with parts (right). Again, our priors \MVcovmat consistently
outperform the baseline \noprior for all individual categories as well
as on average by large margins ($53.3\%$ vs. $41.2\%$ mAP for
\MVcovmat-KITTI without parts, and $69.5\%$ vs. $54.1\%$ with parts).

\paragraph{Summary.}
We conclude that our priors (in particular \MVcovmat) in fact improve
performance for simultaneous 2D bounding box localization and
viewpoint estimation, for different levels of granularity of the
underlying object representation ({\em base}, {\em car-type}, {\em
  car-model}). Notably, our priors allow for robust learning even on
the most fine-grained level of {\em car-models}, where training data
is scarce and unbalanced and \noprior fails. The combination of
fine-grained representation and prior results in a pronounced
performance gain compared to \noprior on the {\em base} level.

\section{Conclusion}\label{sec:conclusion}
In this paper, we have approached the problem of scarce and unbalanced
training data for training multi-view and fine-grained detectors from
a transfer learning perspective, introducing two flavors of learning
prior distributions over permissible detectors, one based on sparse
feature correlations, and one based on the full covariance matrix
between all features. In both cases, we have demonstrated
improved simultaneous 2D bounding box localization and viewpoint
estimation performance when applying these priors to detectors
based on basic-level category representations. In addition, the
second flavor allowed us to learn reliable detectors even for
finer-grained object class representations, resulting in an additional
boost in performance on a realistic dataset of street
scenes~\cite{geiger12cvpr}.


{\scriptsize
\bibliographystyle{ieee}
\bibliography{paper}
}

\pagebreak
\section{Supplemental material}
\label{sec:suppl}
\subsection{Training and test data distributions for {\em car}, {\em car-type} and {\em car-model} category levels}
In this section, we visualize the viewpoint distributions of the
realistic street scene dataset (the tracking benchmark of
KITTI~\cite{geiger12cvpr}) that is the basis of our experiments
in Sect.~\ref{sec:exp_multiview_detection}.To that end, we plot
histograms of the number of car instances for each of 8 viewpoint
(azimuth angle) bins, separately for training (blue) and test (red)
data, and distinguishing between three different
levels of granularity ({\em car}, {\em car-type} and {\em car-model}).

\begin{figure}[!h]
\centering
\includegraphics[height=3cm]{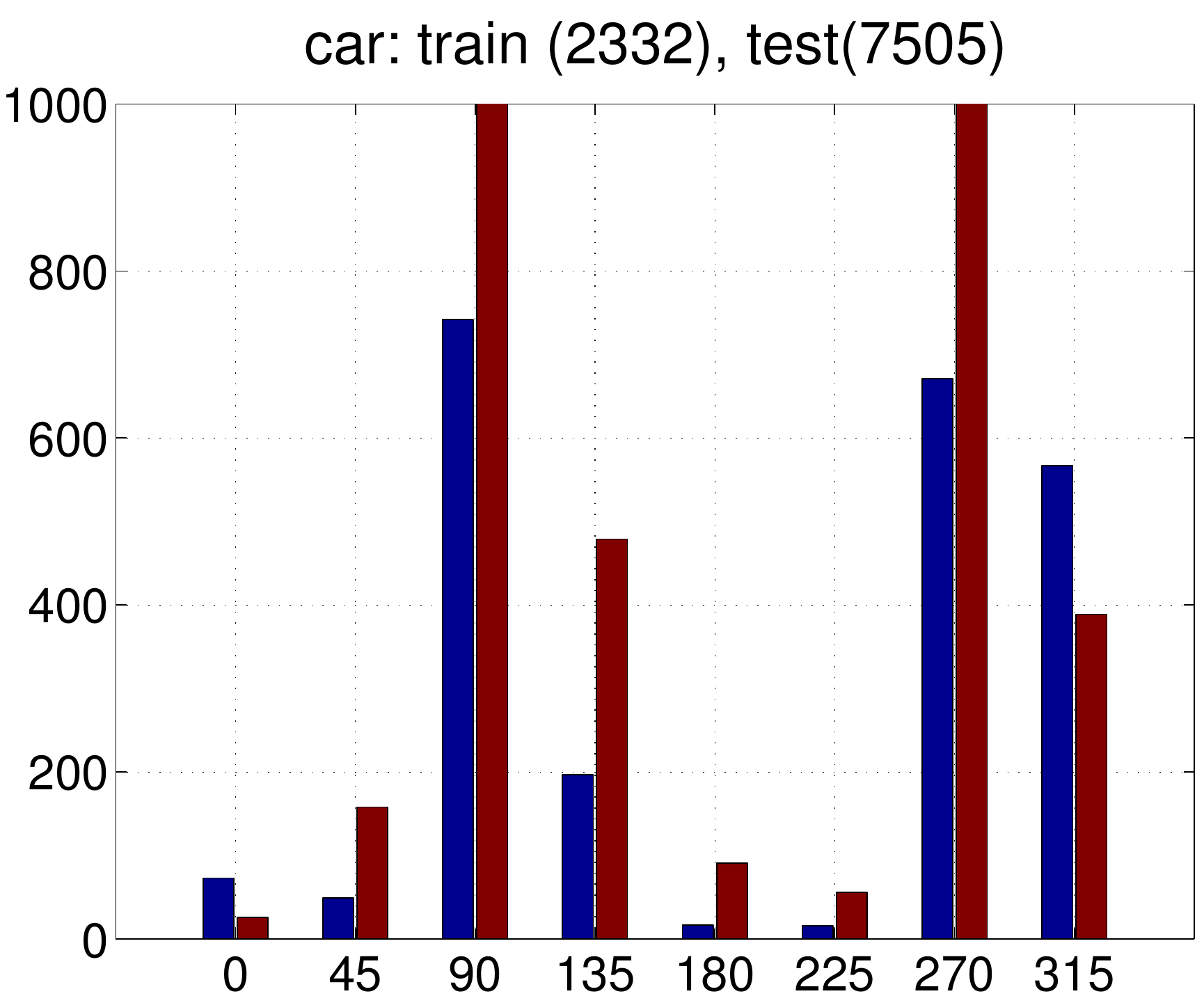}
\caption{{\em Car} train and test statistics over 8 viewpoint bins.}
\label{fig:carstat}
\end{figure}
Fig.~\ref{fig:carstat} shows the data distribution for the
{\em car} class. Notice the unbalanced data distribution across
views. There are two main modes in the viewpoint distribution, at
$90^{\circ}$ and $270^{\circ}$, which represent back and front
views. This behavior is expected, as in the
KITTI dataset~\cite{geiger12cvpr} the images have been taken from a
driving vehicle. On the other hand, the left ($0^{\circ}$) and right
($180^{\circ}$) facing views are poorly represented. With only $16$
right facing training examples, training a robust right-view {\em car}
template is extremely challenging.

\begin{figure}[!h]
\centering
\includegraphics[height=2.5cm]{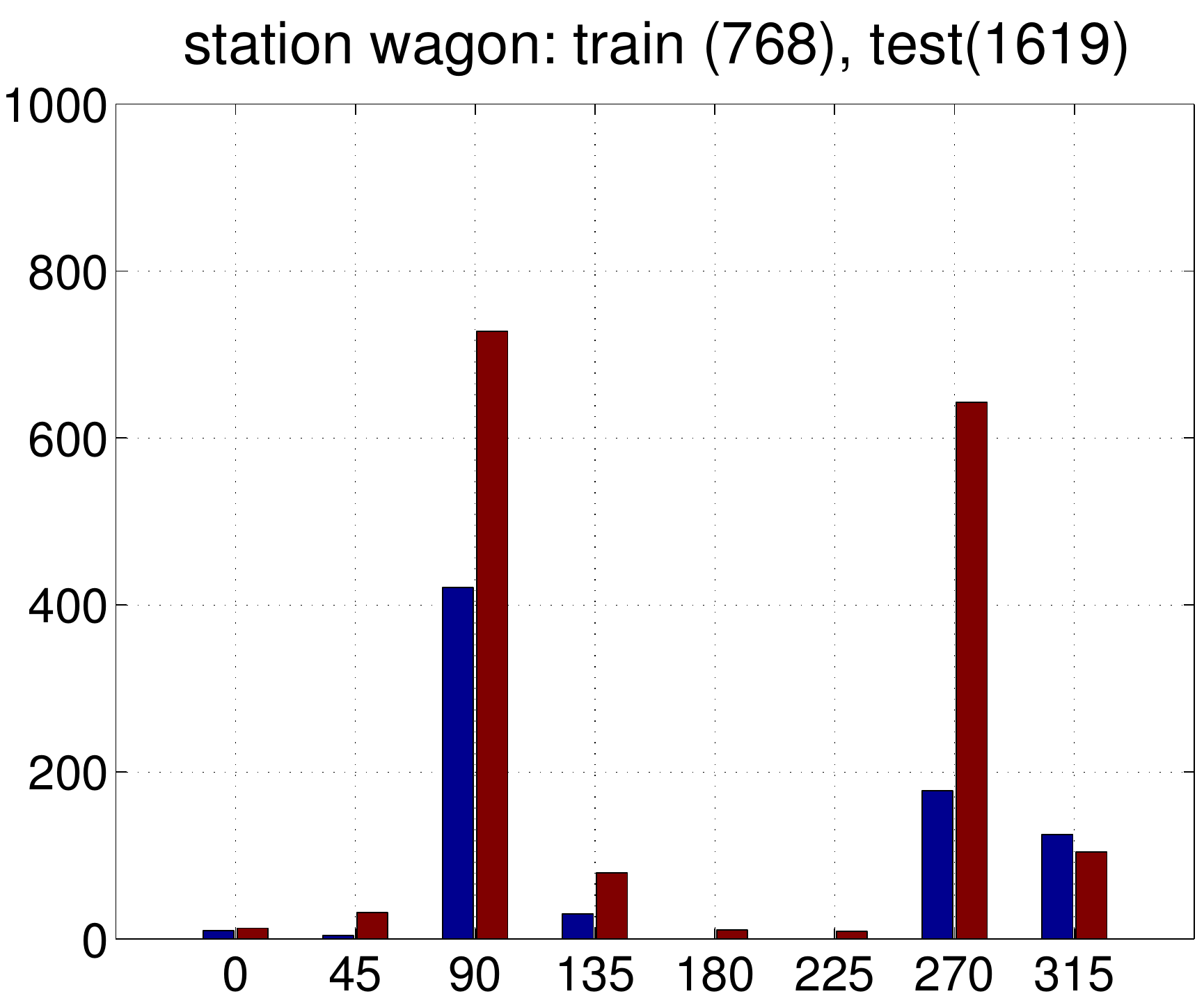}
\includegraphics[height=2.5cm]{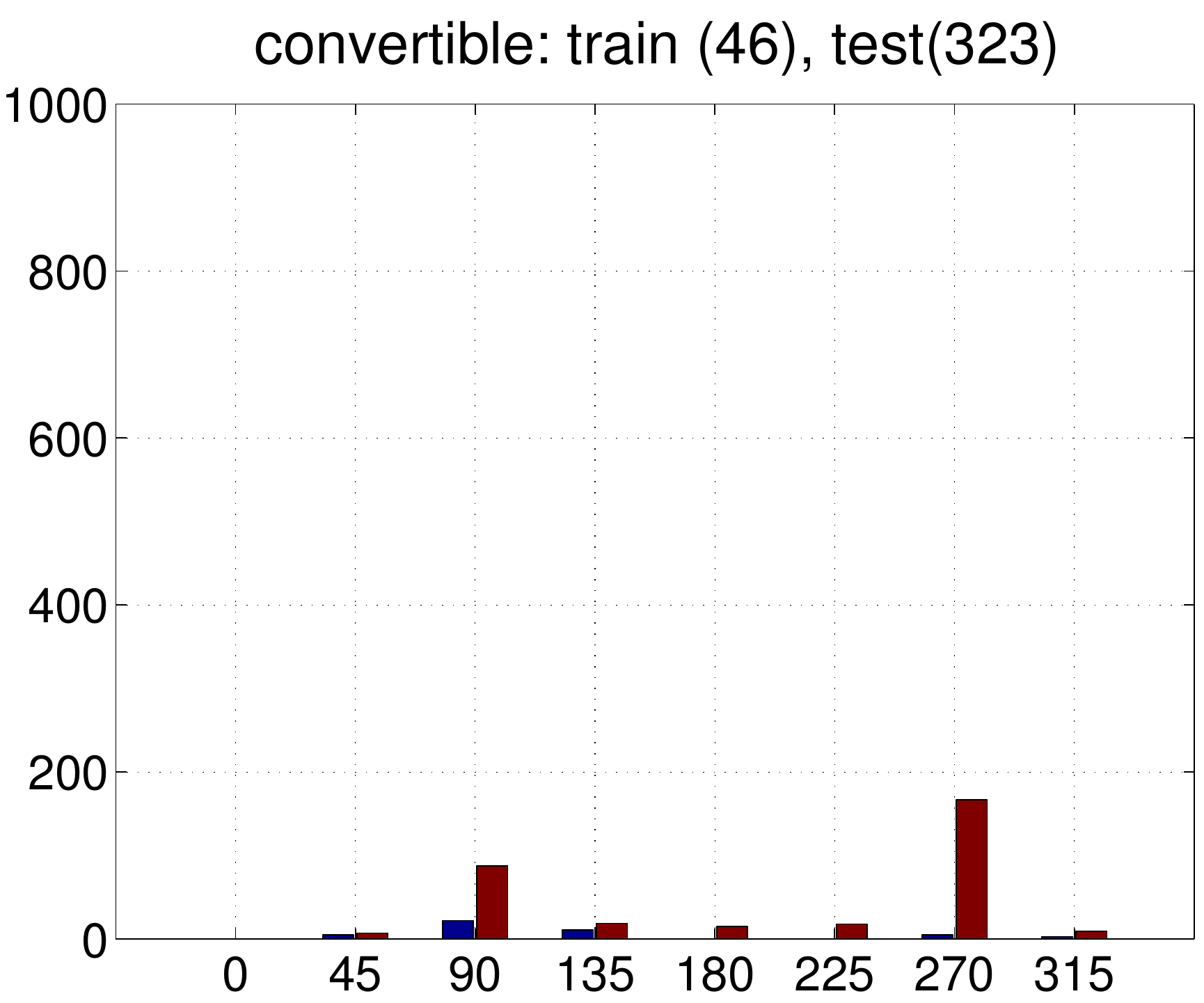}
\includegraphics[height=2.5cm]{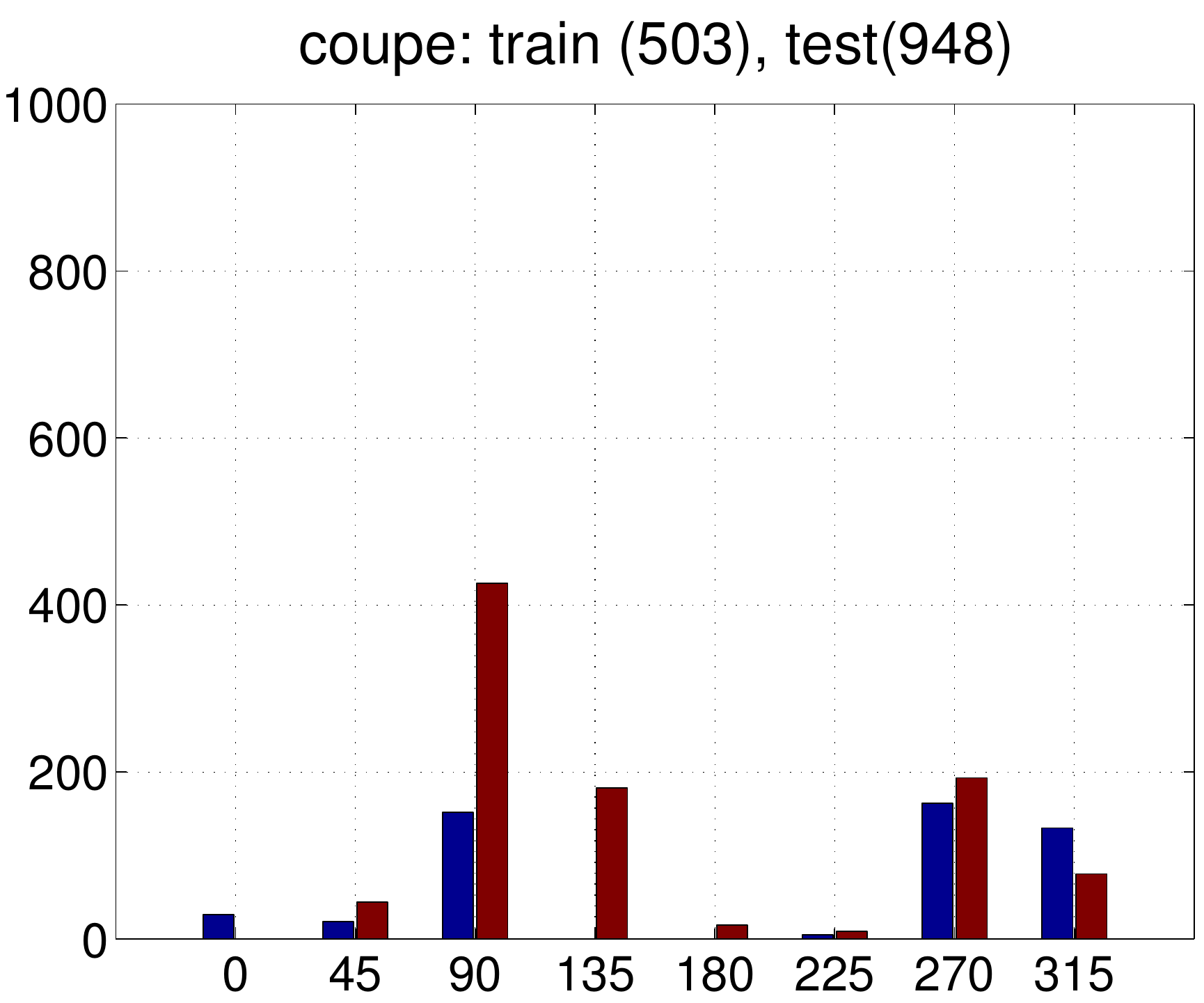}
\includegraphics[height=2.5cm]{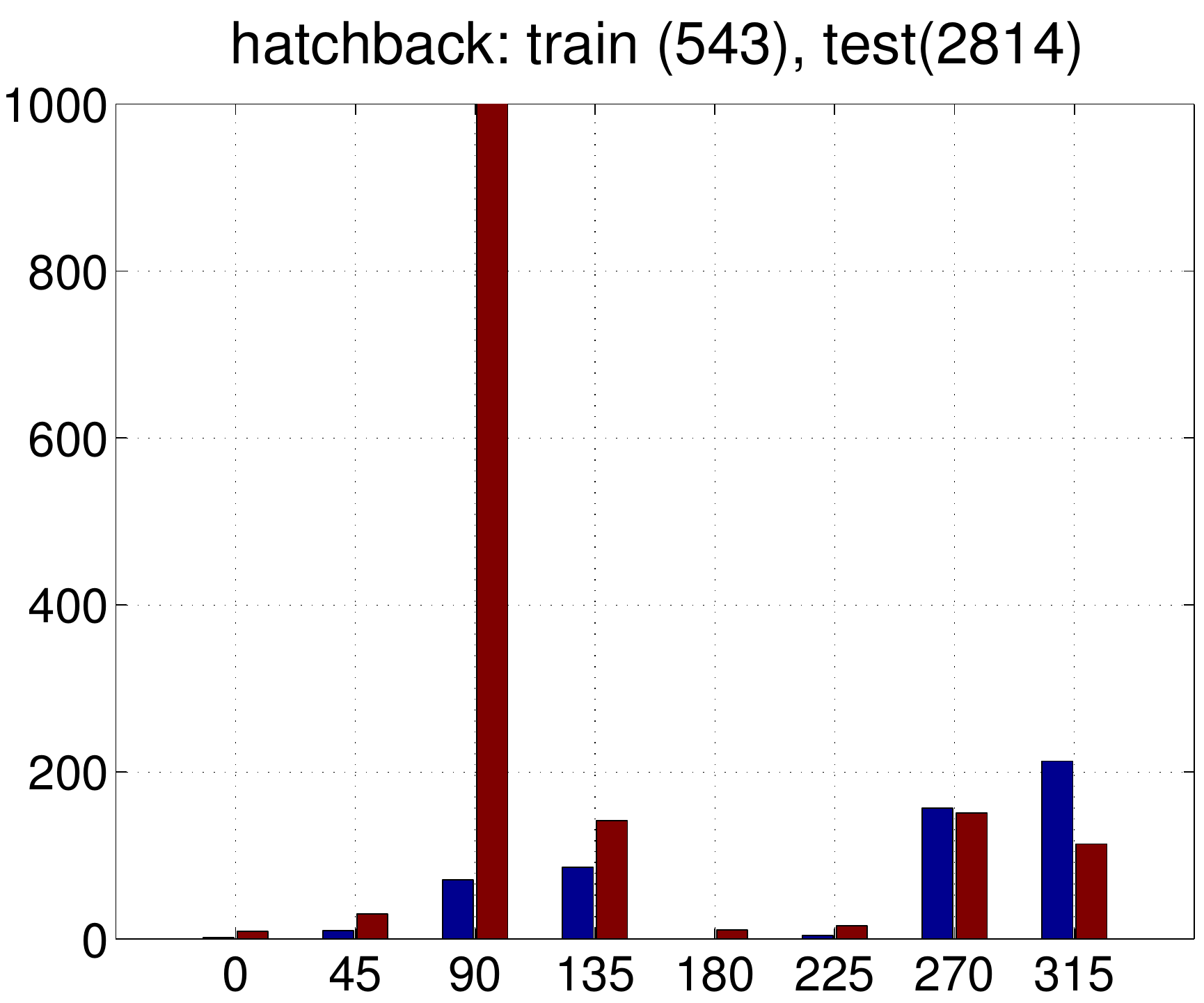}\\
\includegraphics[height=2.5cm]{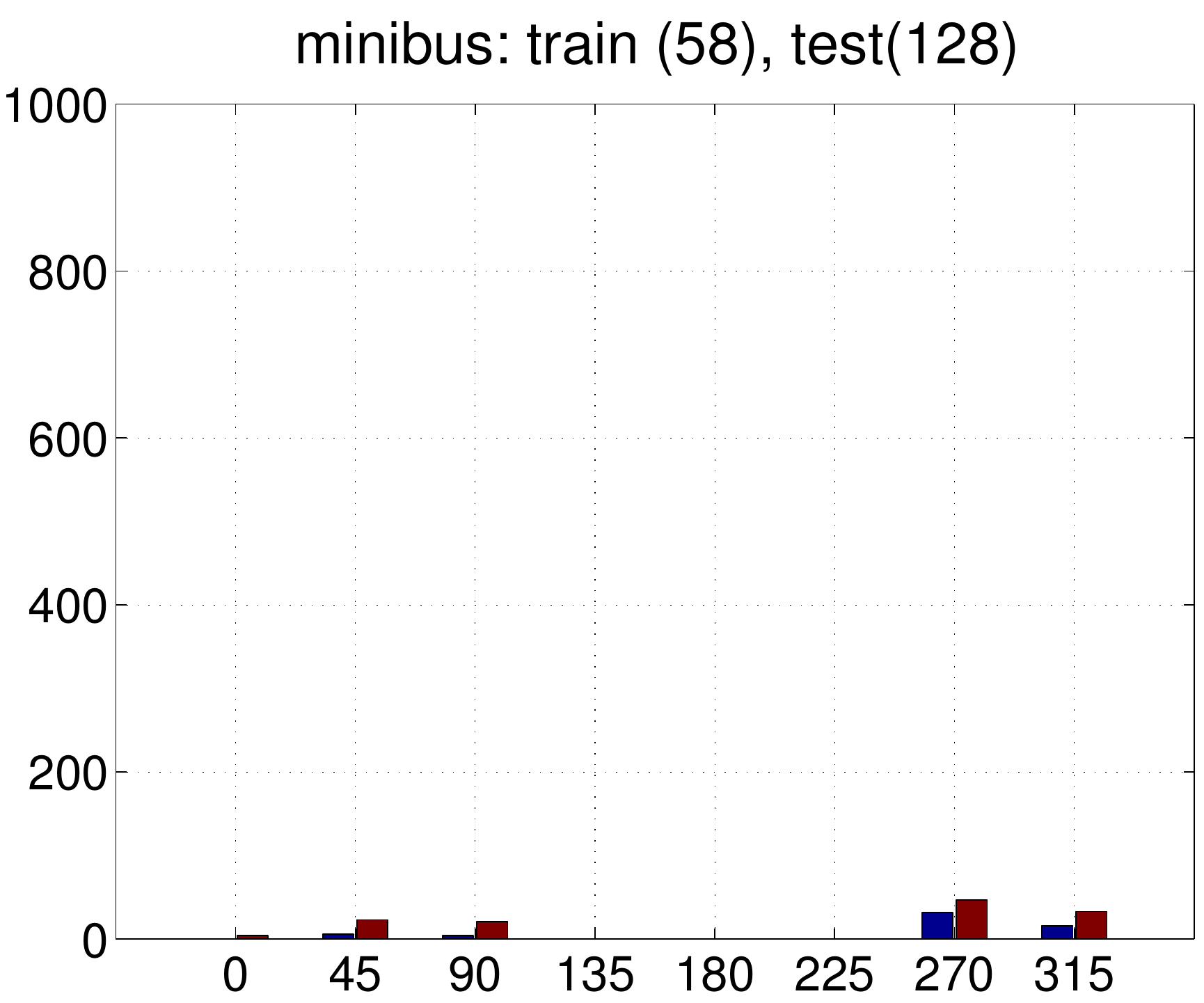}
\includegraphics[height=2.5cm]{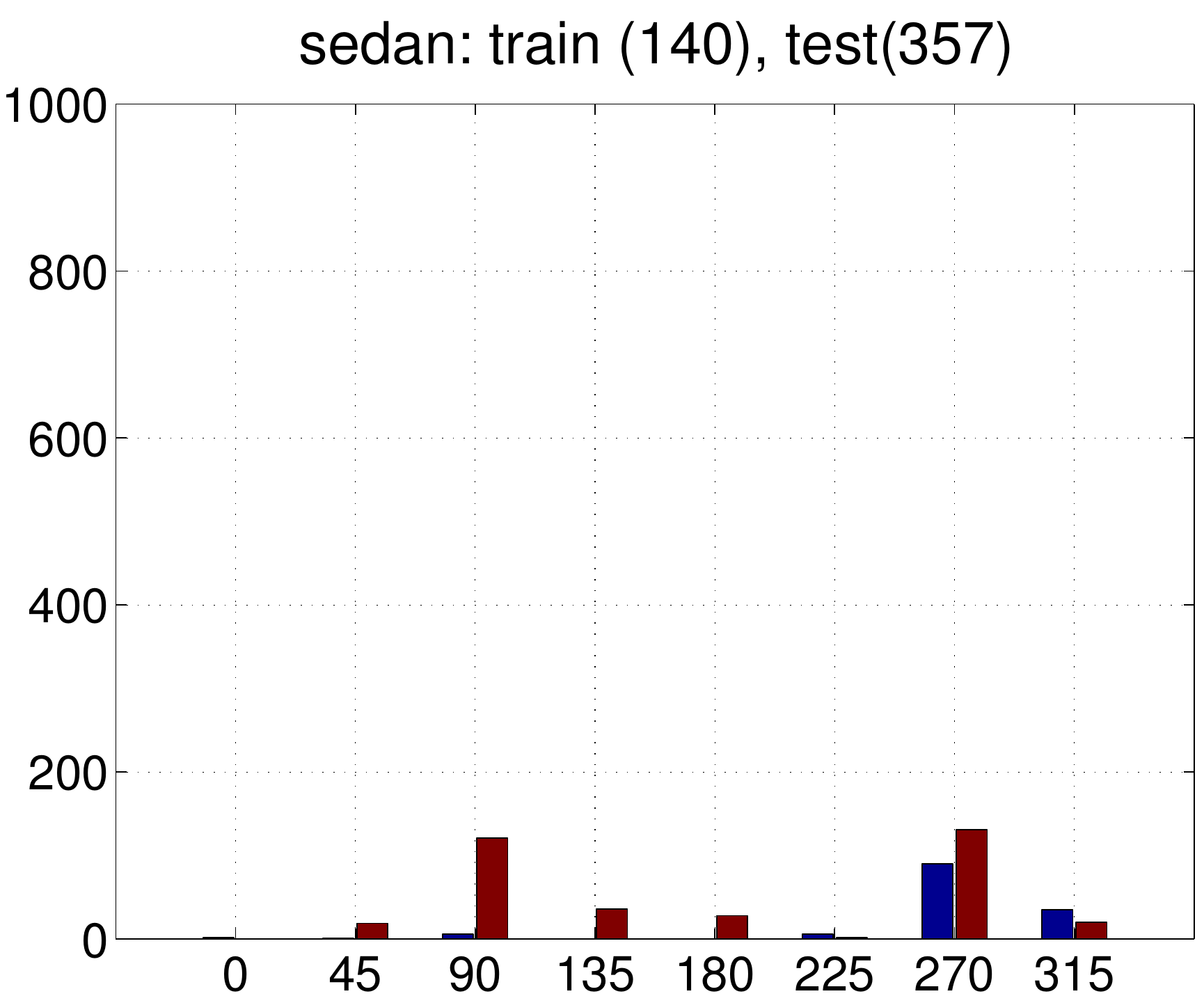}
\includegraphics[height=2.5cm]{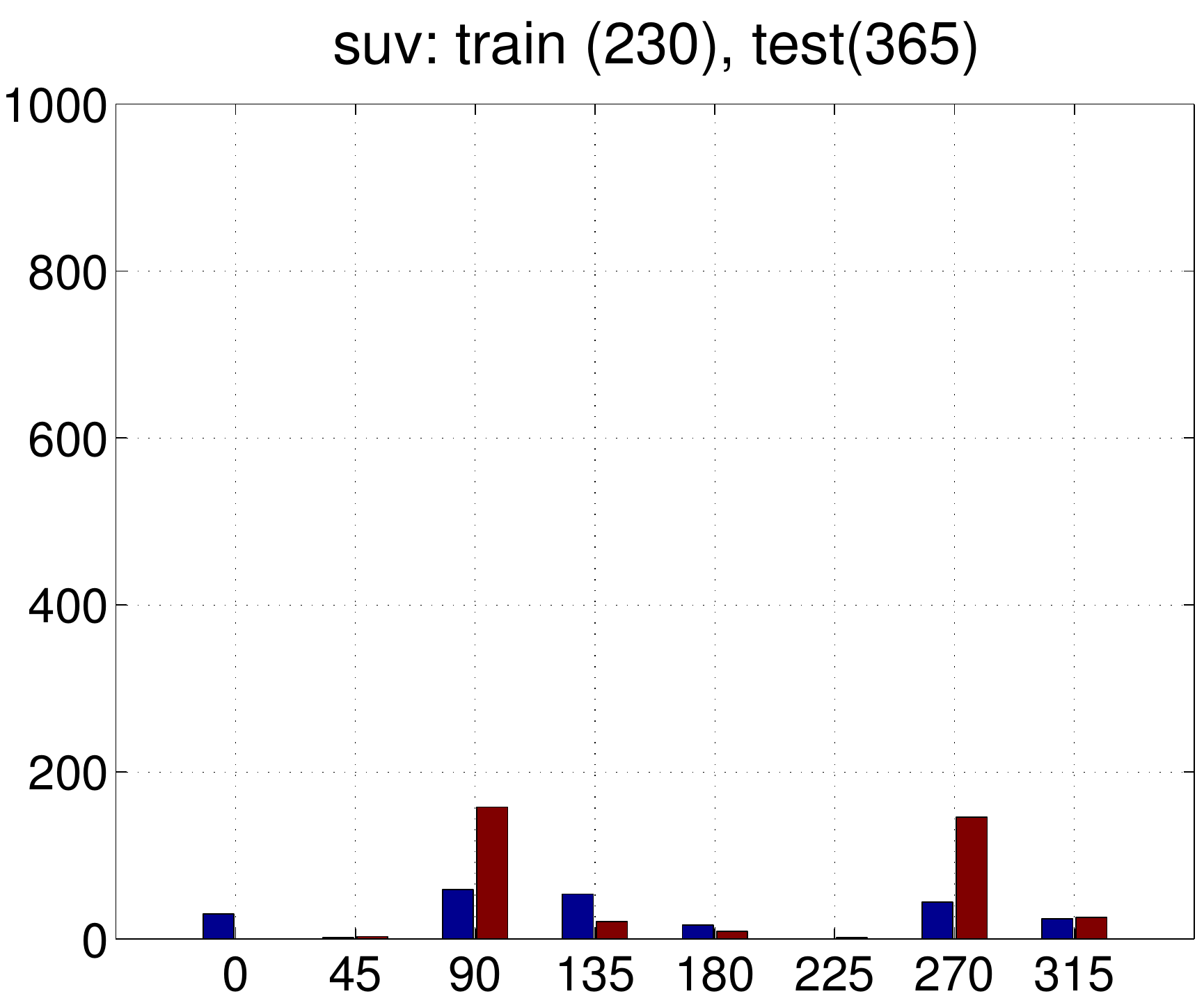}
\caption{{\em Car-types} train and test statistics over 8 viewpoint bins.}
\label{fig:cartypesstat}
\end{figure}
Fig.~\ref{fig:cartypesstat} illustrates the train and test
distributions for each of the $7$ {\em car types} (station wagon,
convertible, coupe, hatchback, minibus, sedan, suv). We observe:
i) the amount of training data available per {\em
  car-type} varies dramatically; classes like {\em station wagon,
  coupe, hatchback} have much more examples than {\em
  convertible, minibus, sedan}, ii) the training data
distributions are skewed. There is not
a single {\em car-type} for which all viewpoints are represented in the training
data, thus learning a full multi-view representation for a {\em
  car-type} is impossible with a standard \noprior framework. And
iii), the train and test distributions differ a lot for
each {\em car-type}. For example, for {\em sedan} and {\em
  convertible} there are viewpoints in the test set which are not
represented in the training set.

\begin{figure}
\centering
\includegraphics[height=2.5cm]{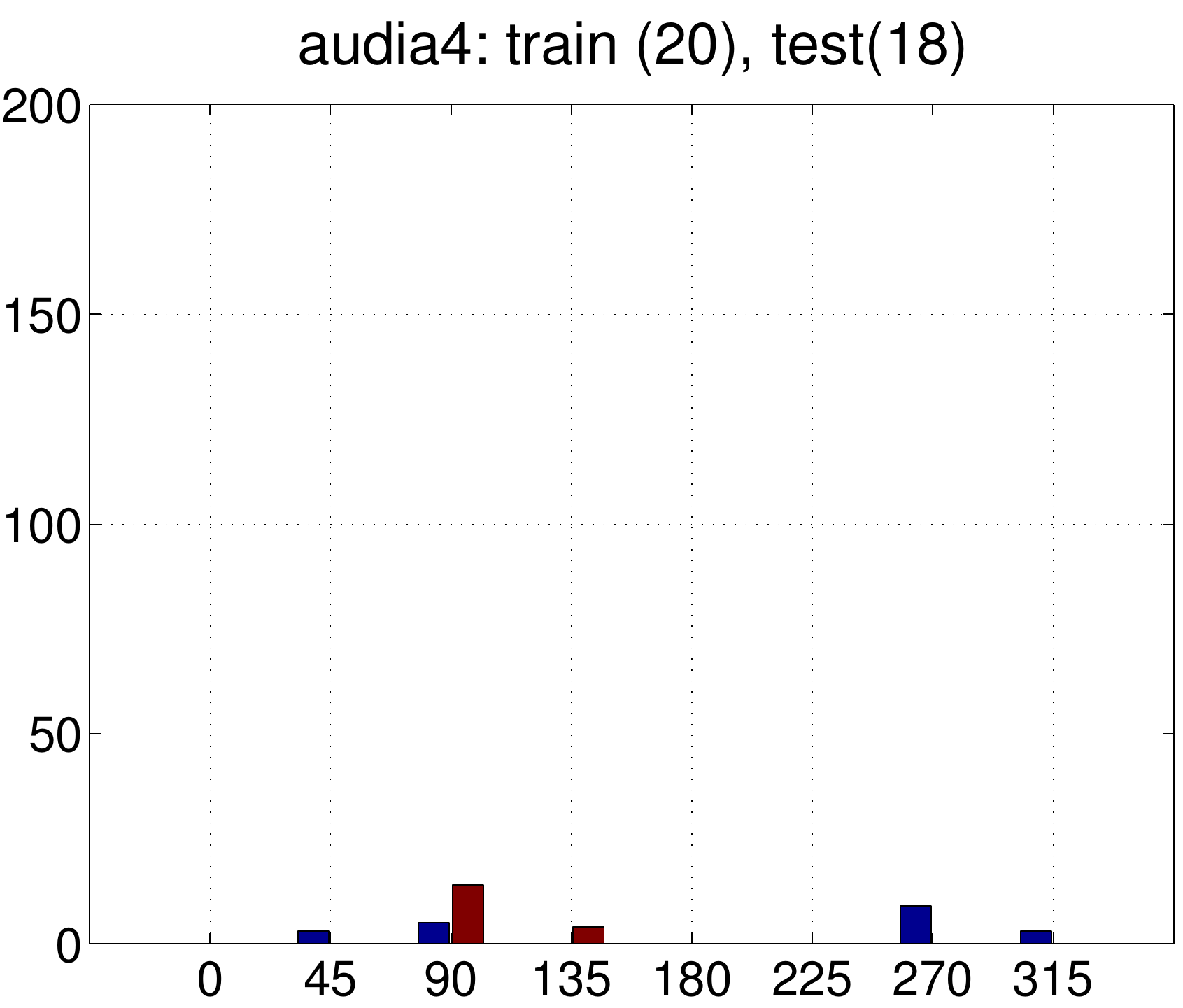}
\includegraphics[height=2.5cm]{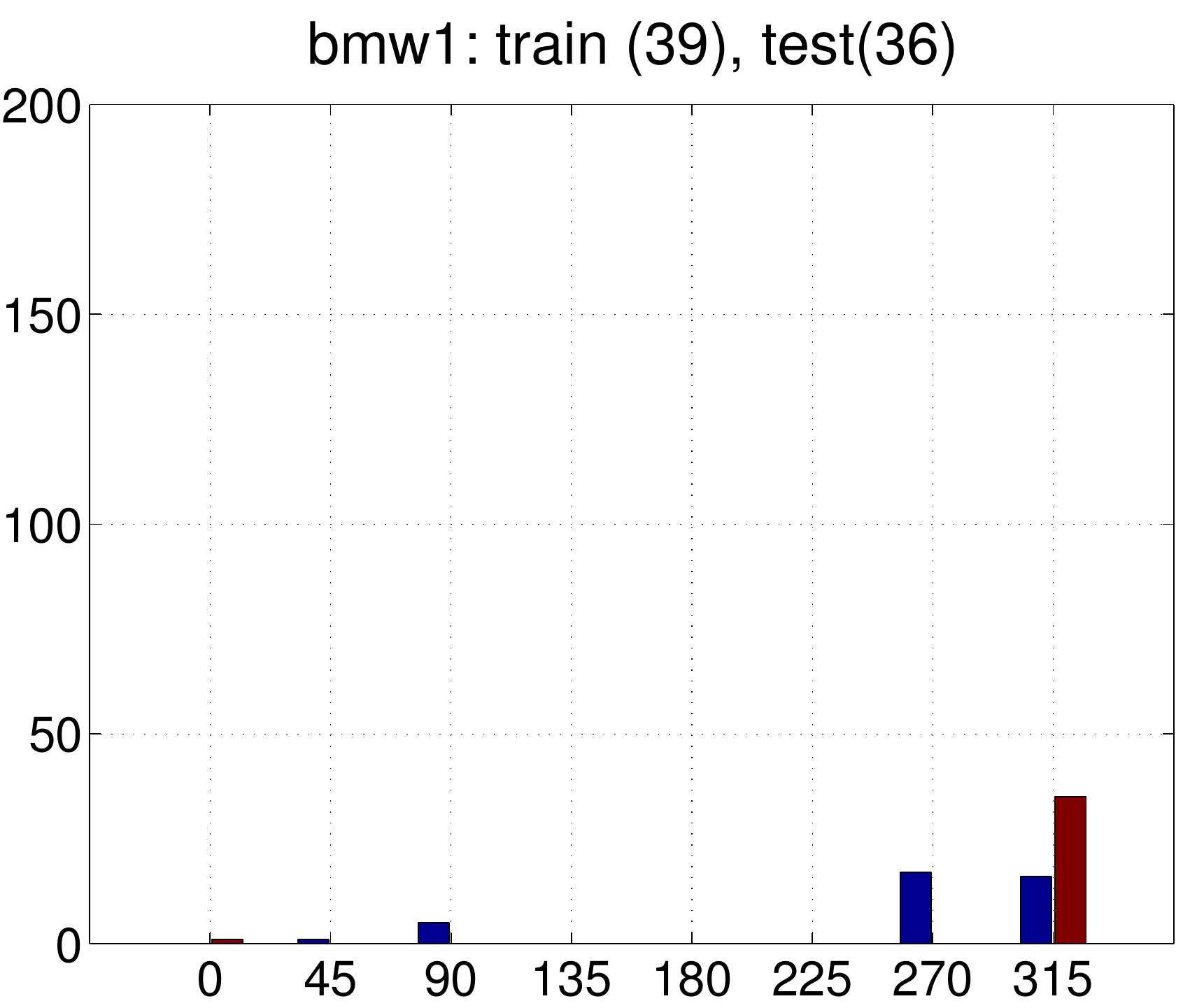}
\includegraphics[height=2.5cm]{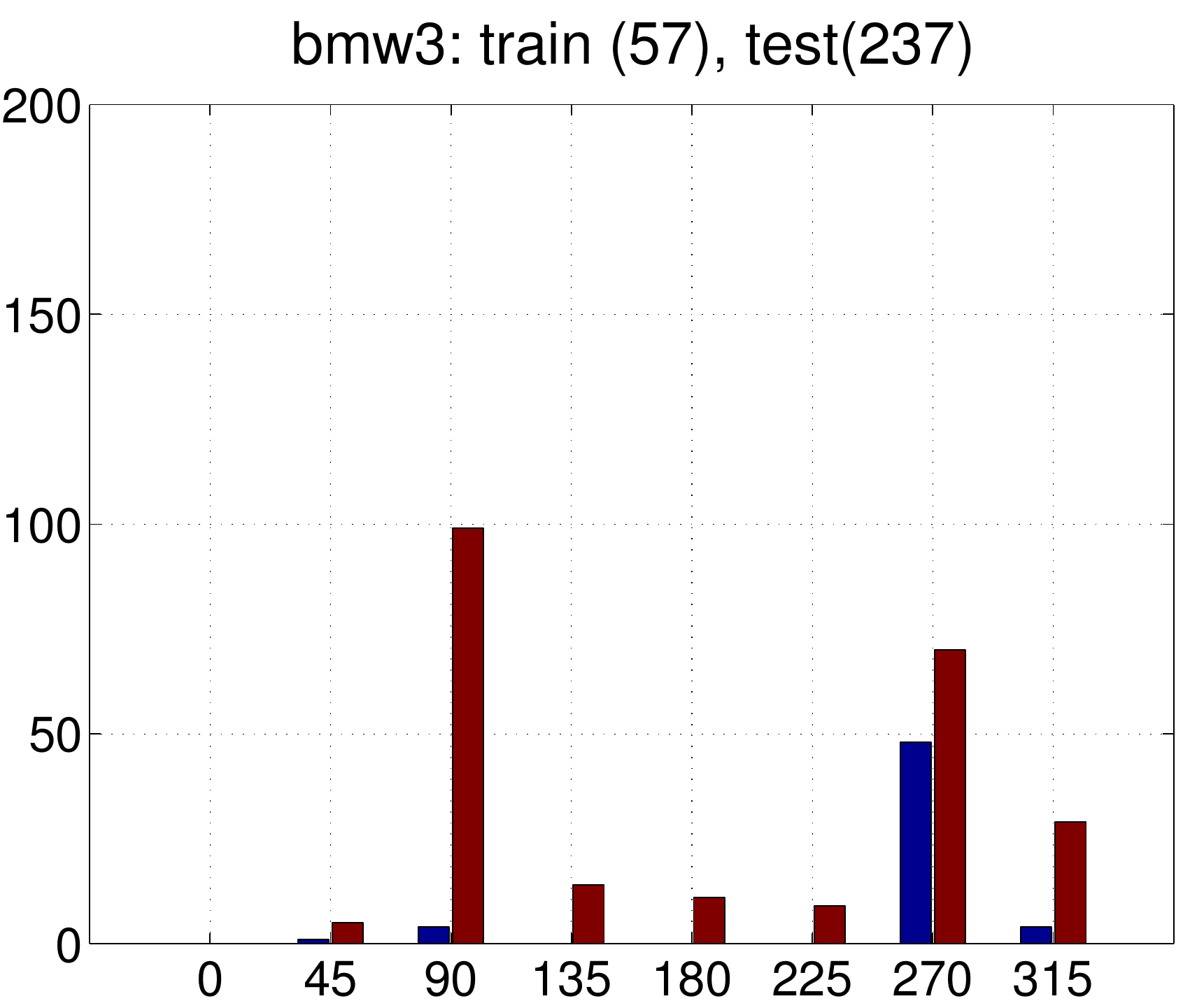}
\includegraphics[height=2.5cm]{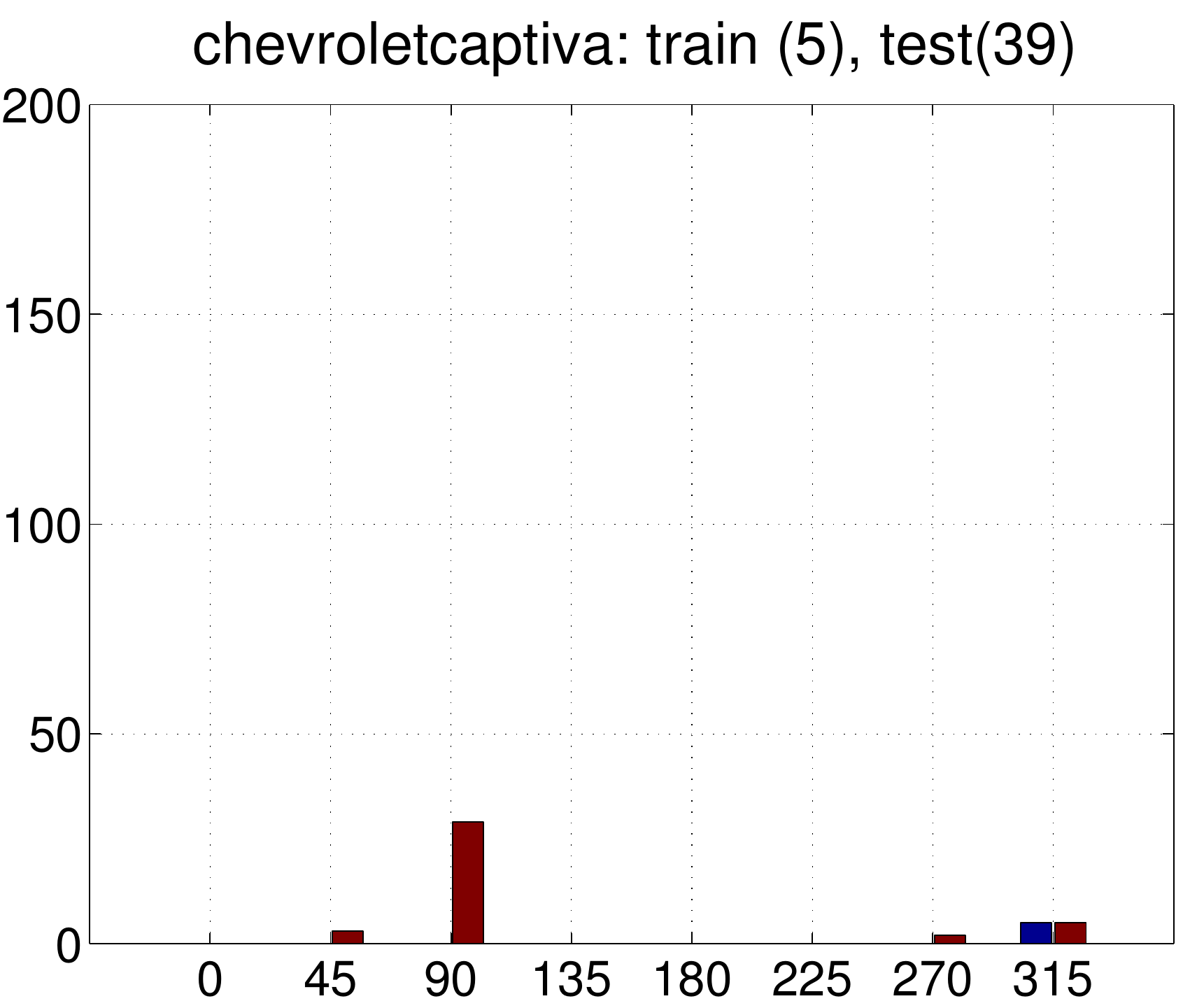}\\
\includegraphics[height=2.5cm]{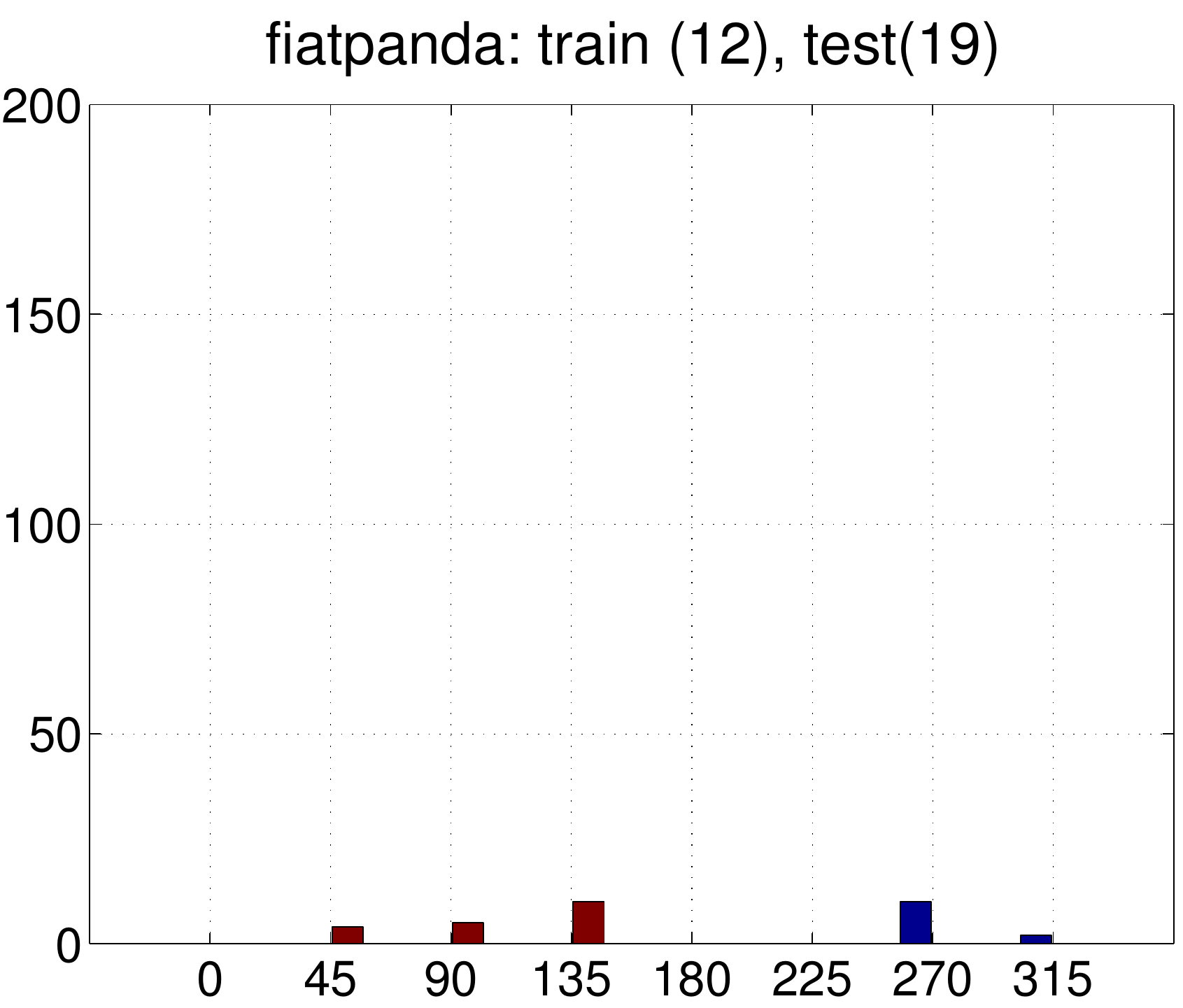}
\includegraphics[height=2.5cm]{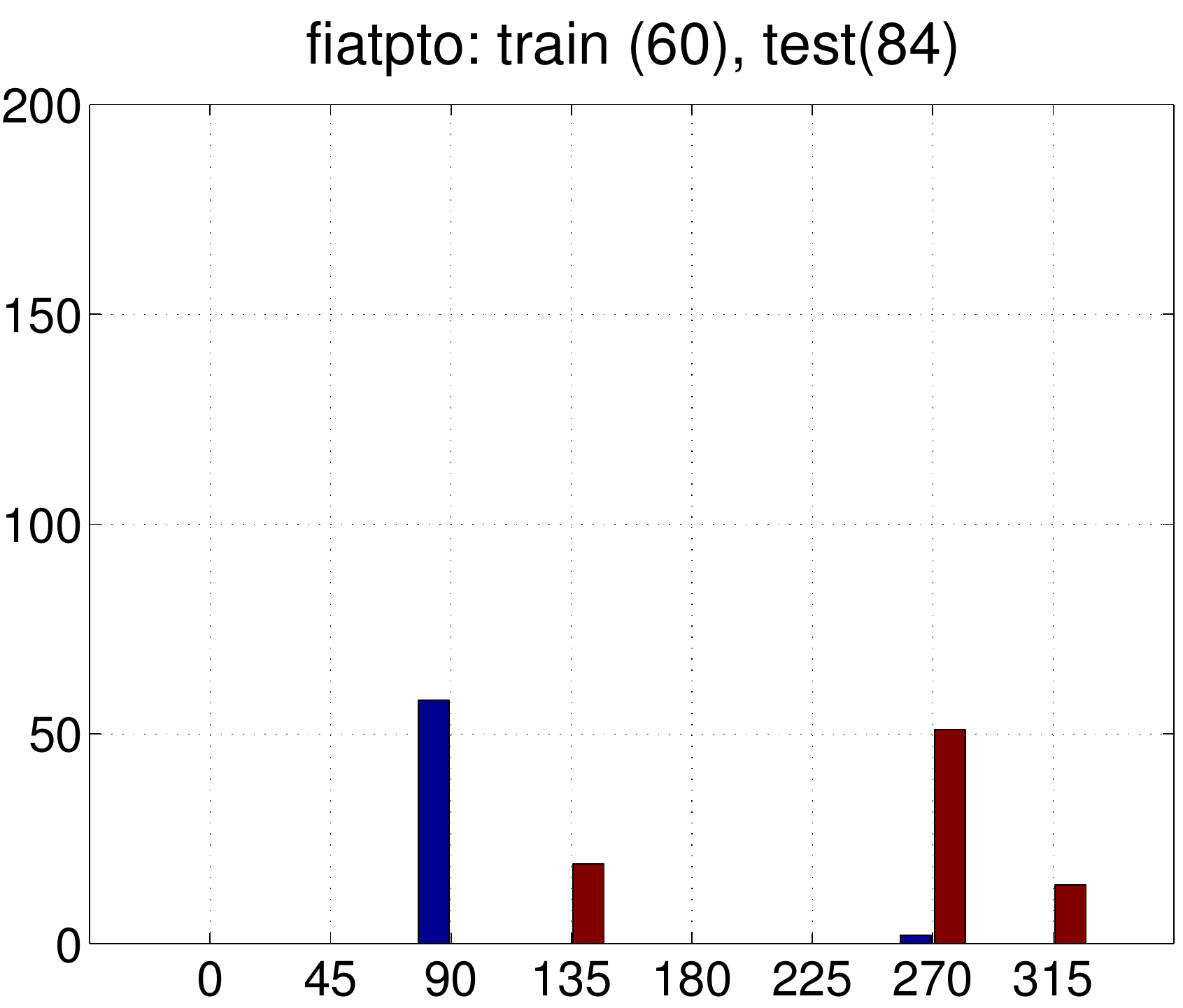}
\includegraphics[height=2.5cm]{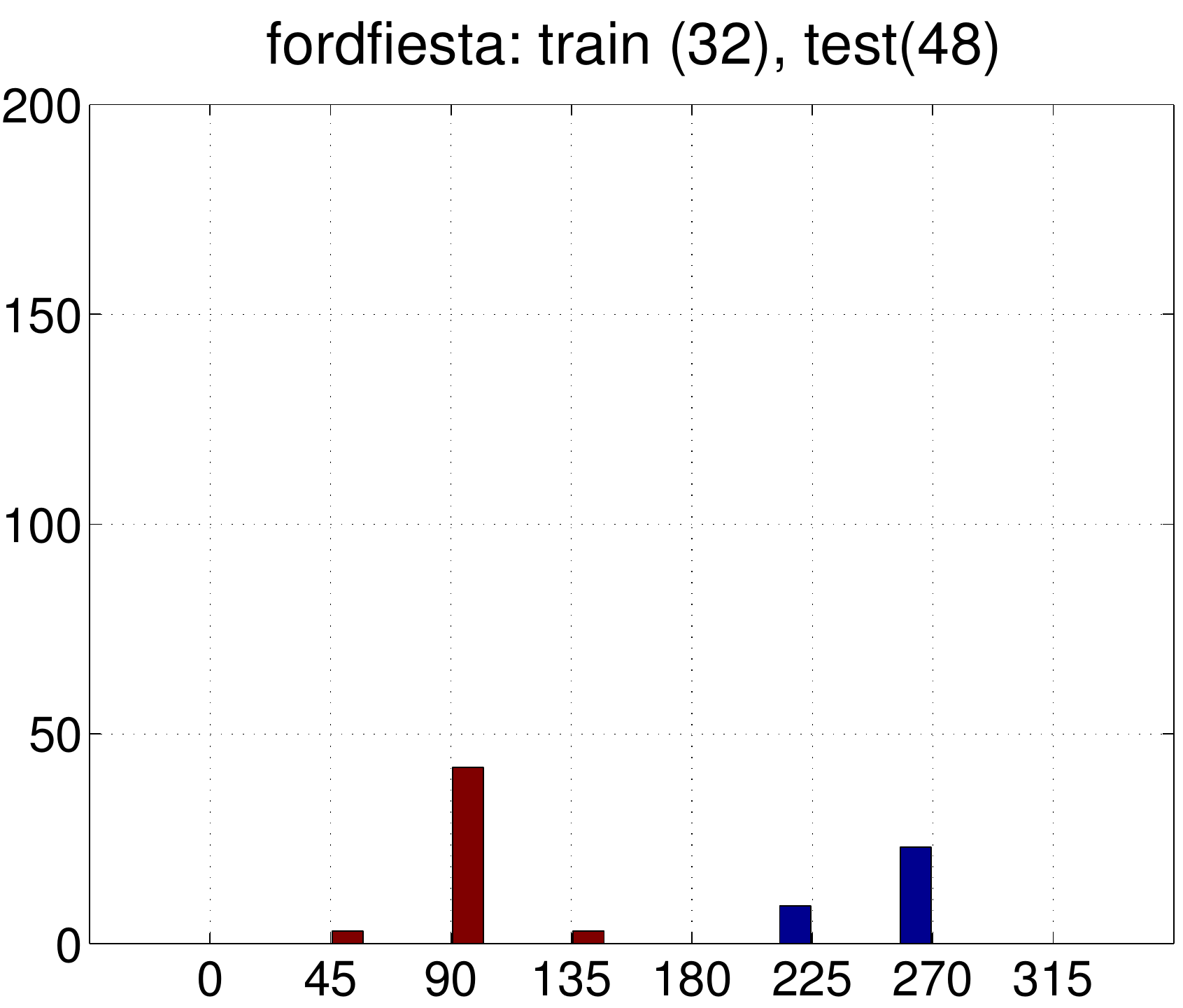}
\includegraphics[height=2.5cm]{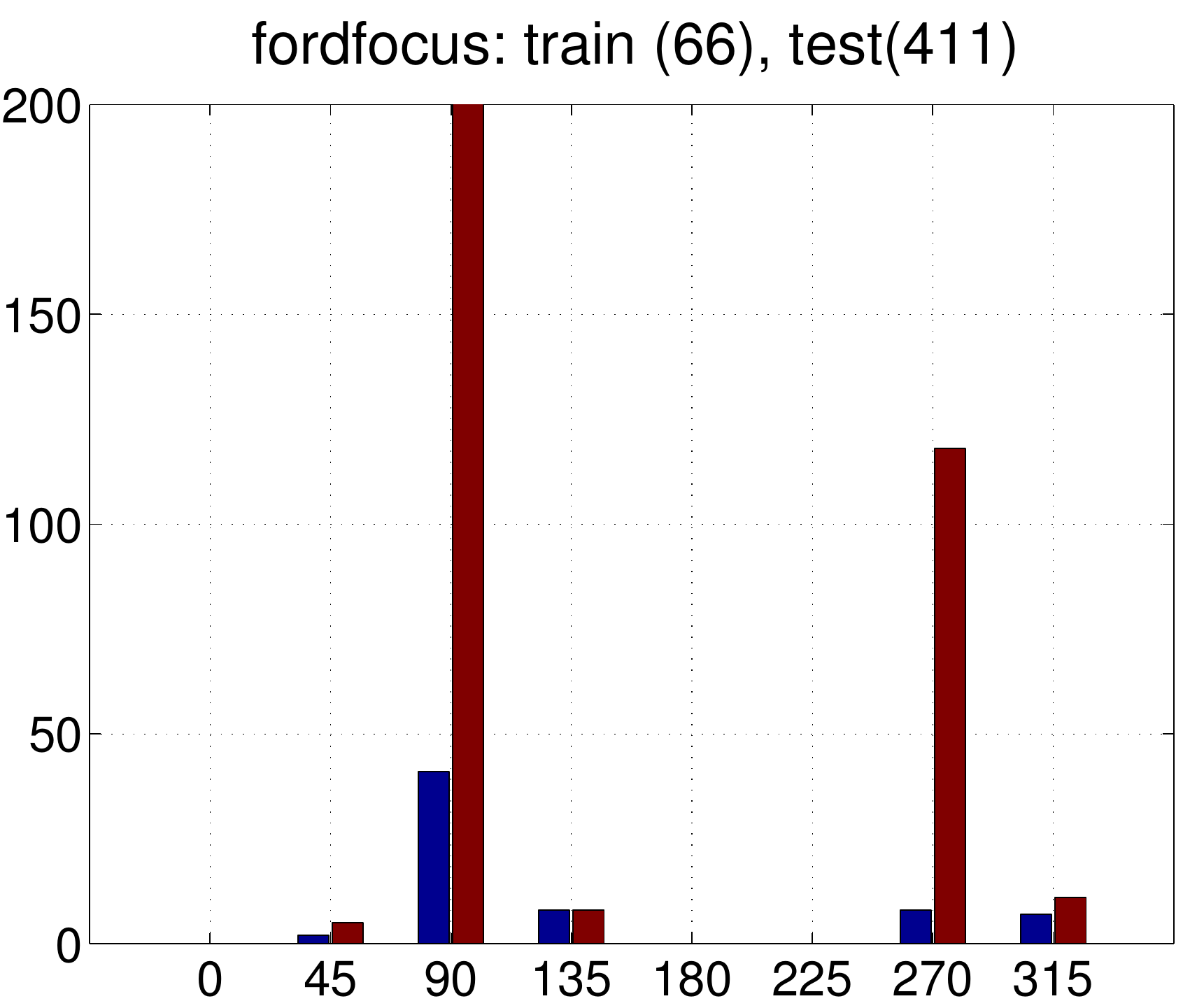}\\
\includegraphics[height=2.5cm]{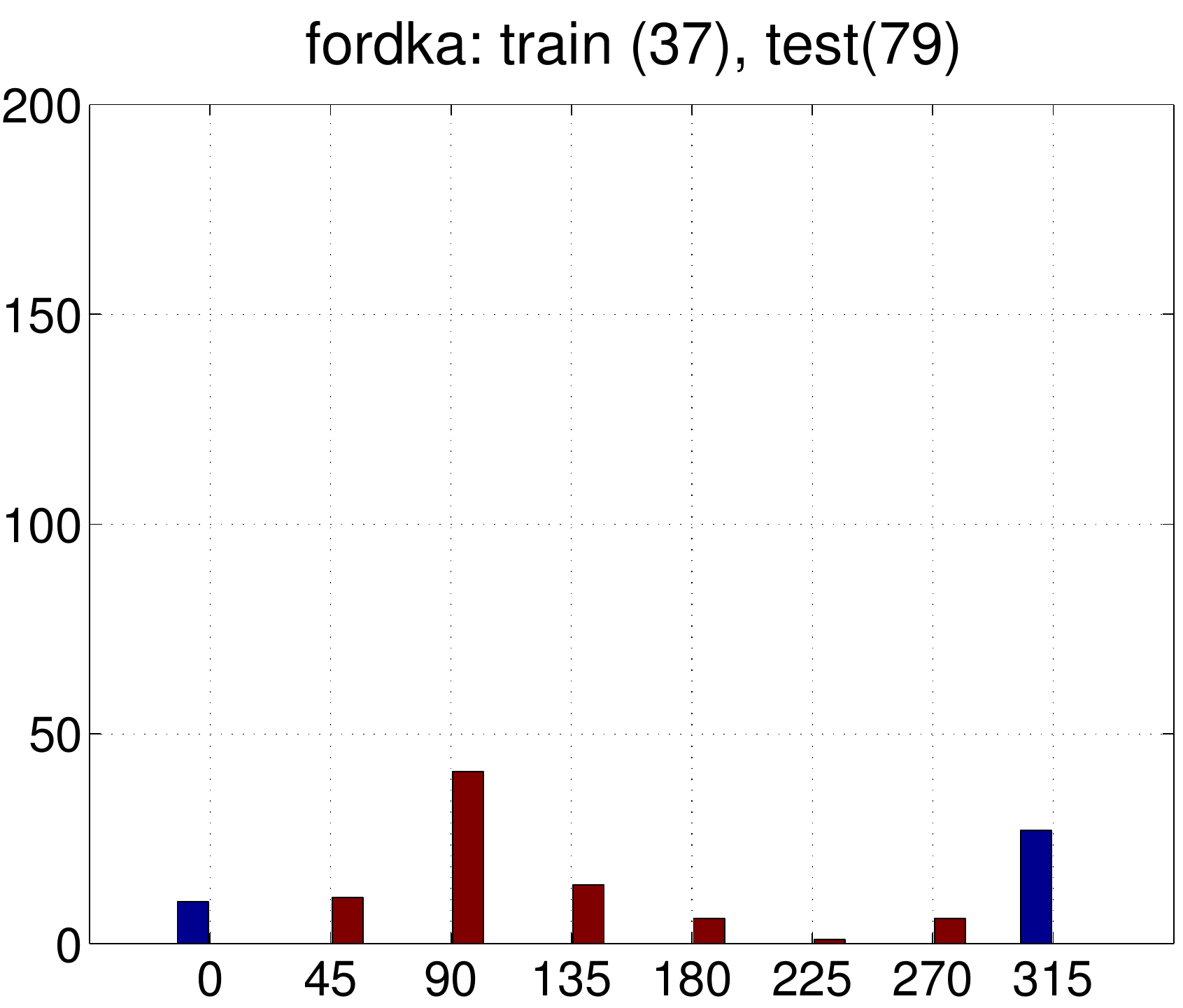}
\includegraphics[height=2.5cm]{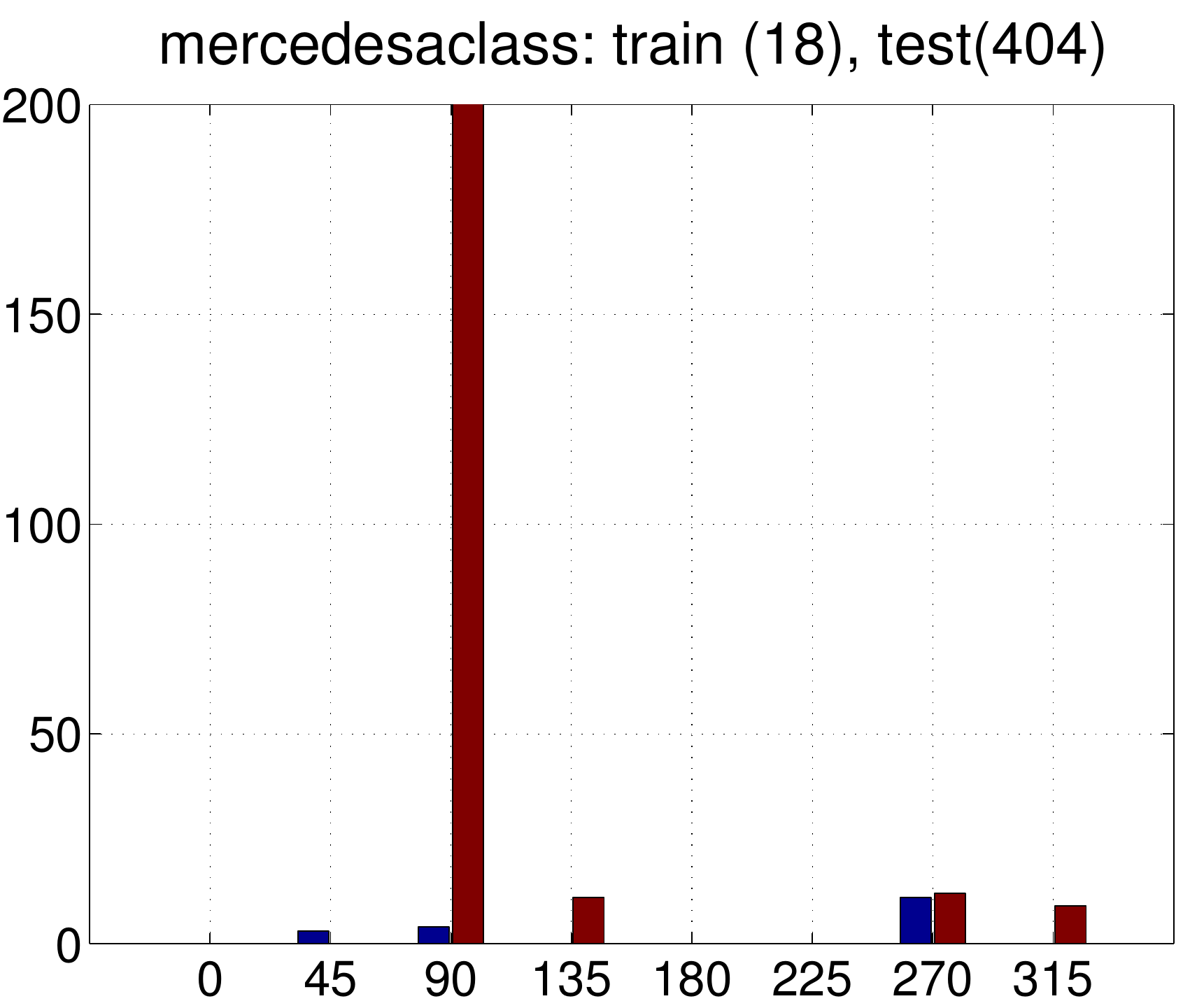}
\includegraphics[height=2.5cm]{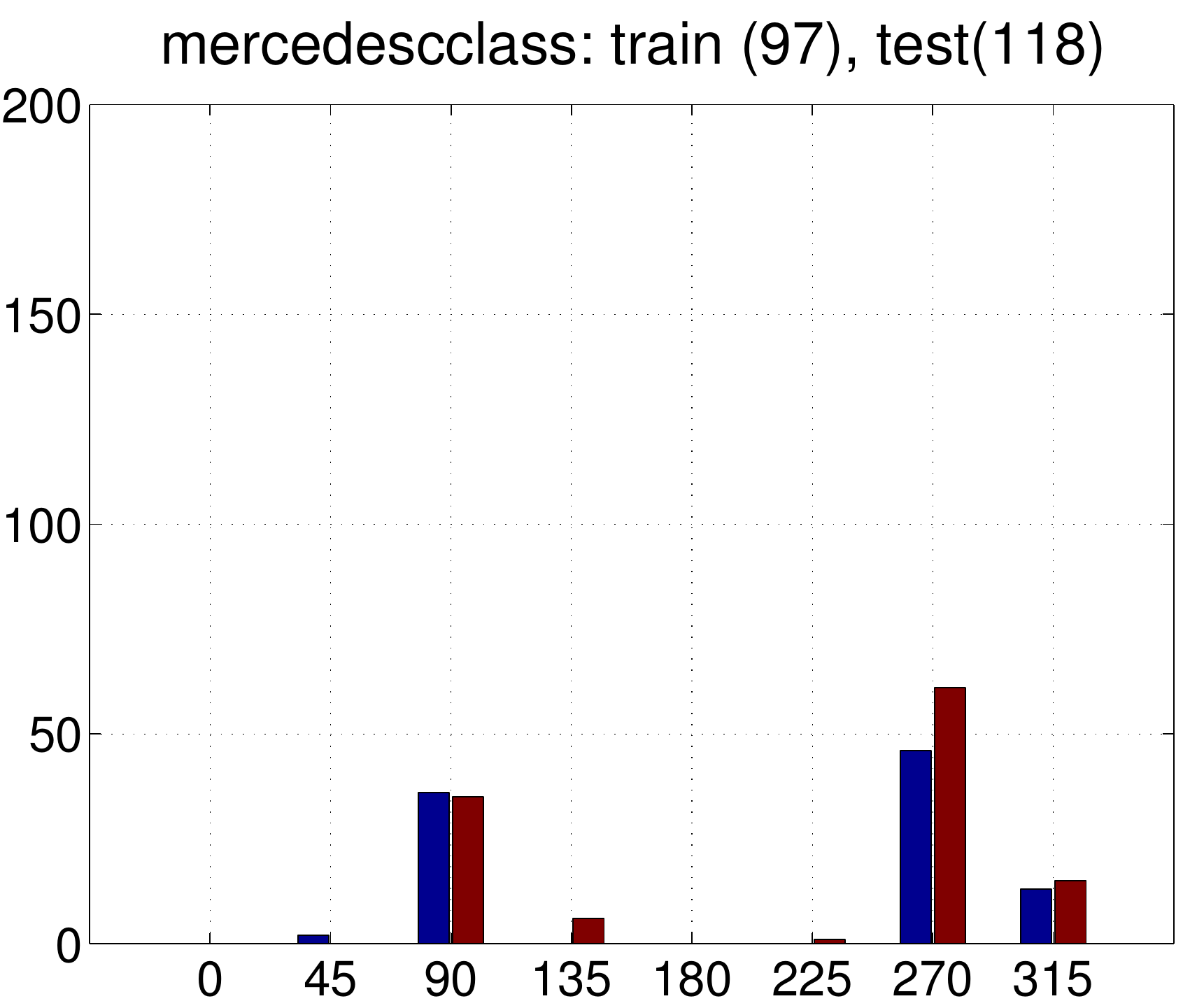}
\includegraphics[height=2.5cm]{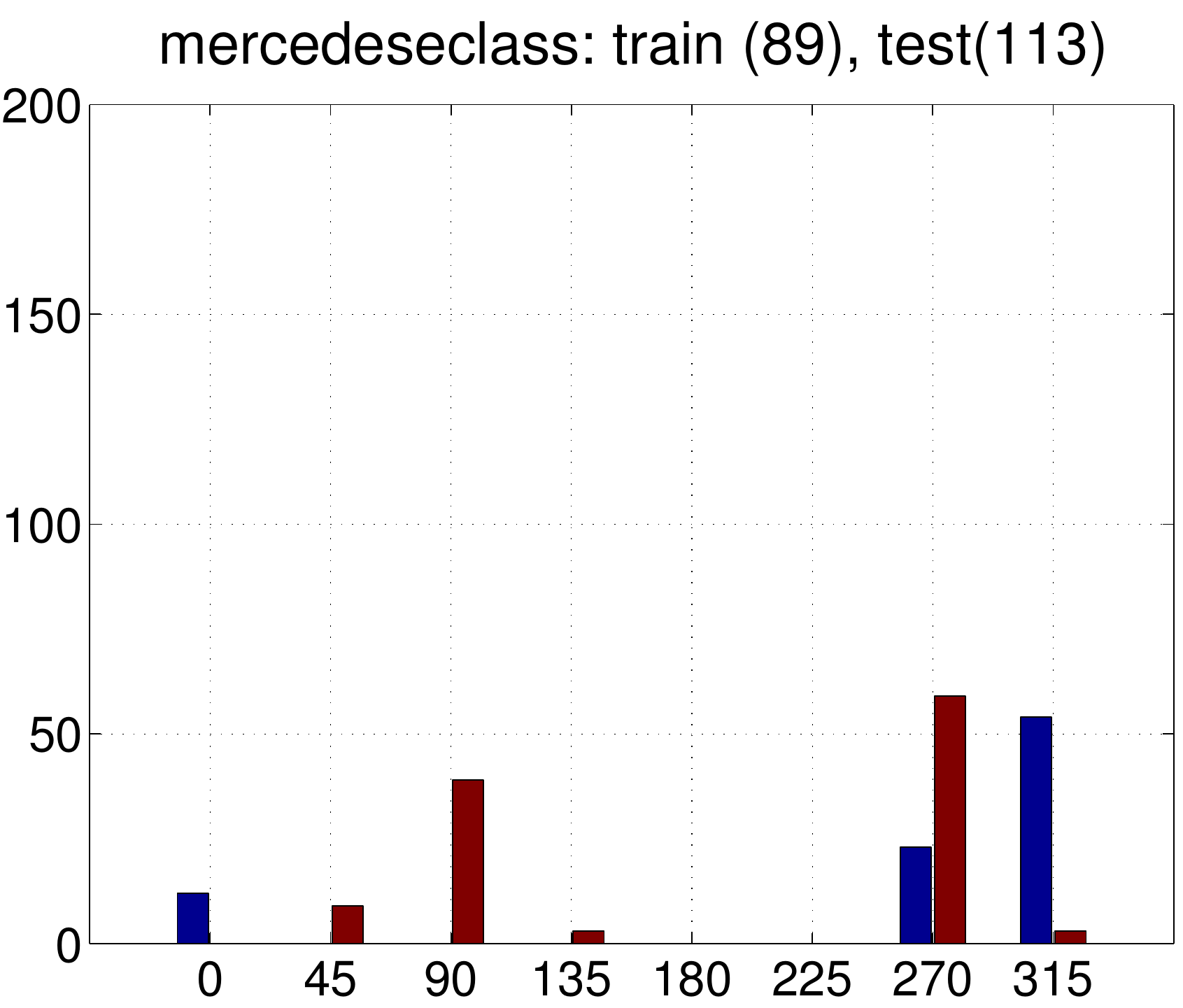}\\
\includegraphics[height=2.5cm]{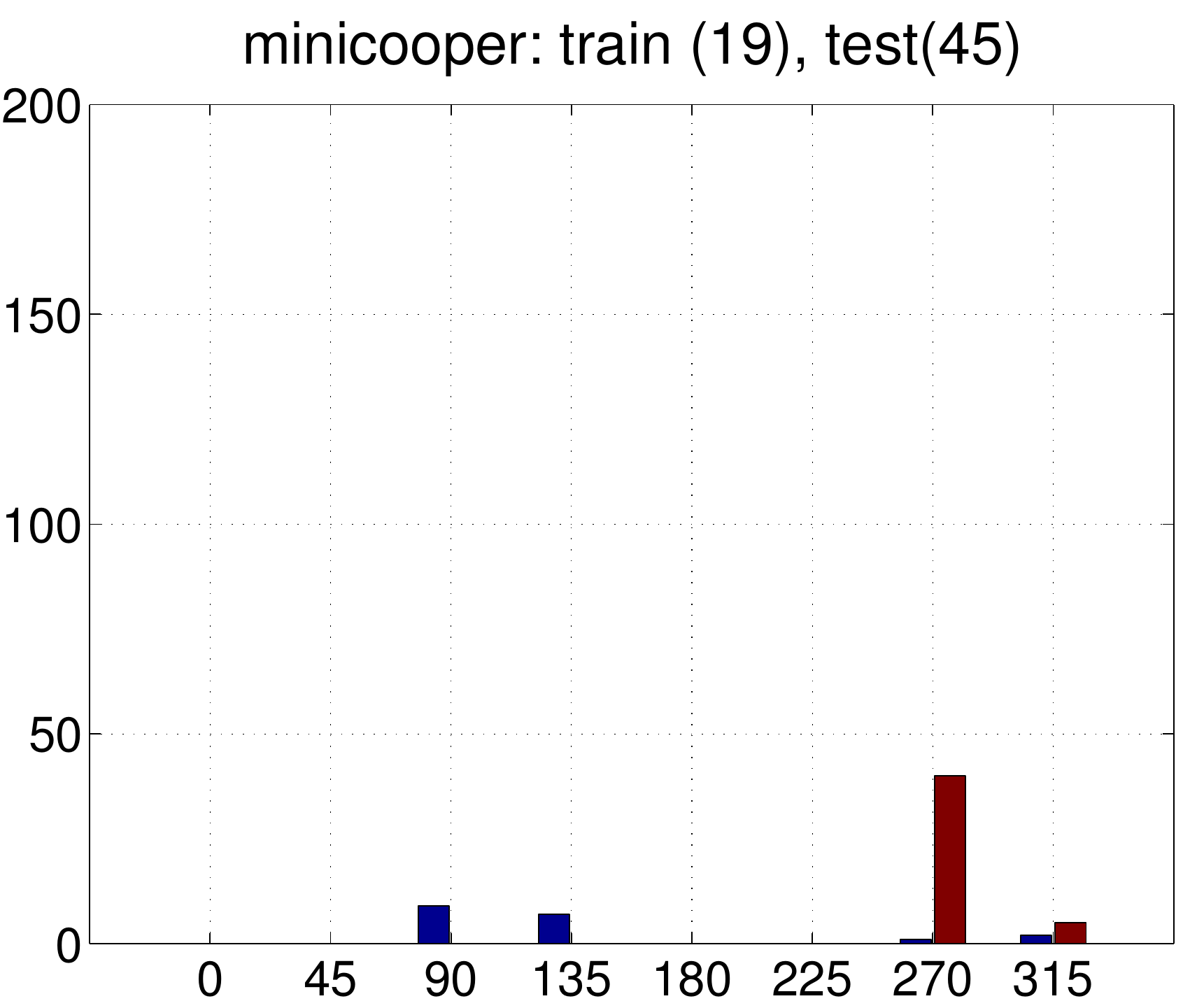}
\includegraphics[height=2.5cm]{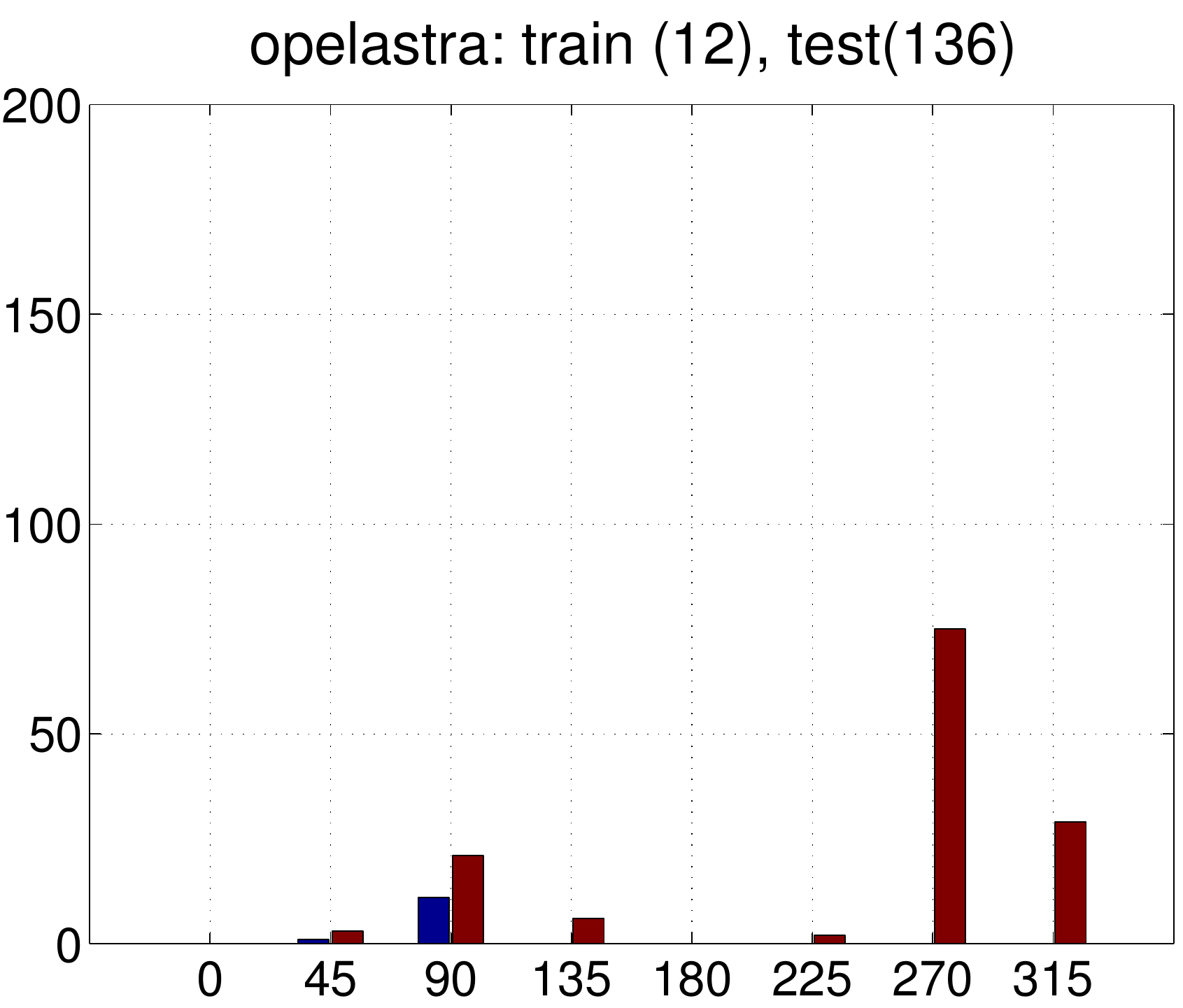}
\includegraphics[height=2.5cm]{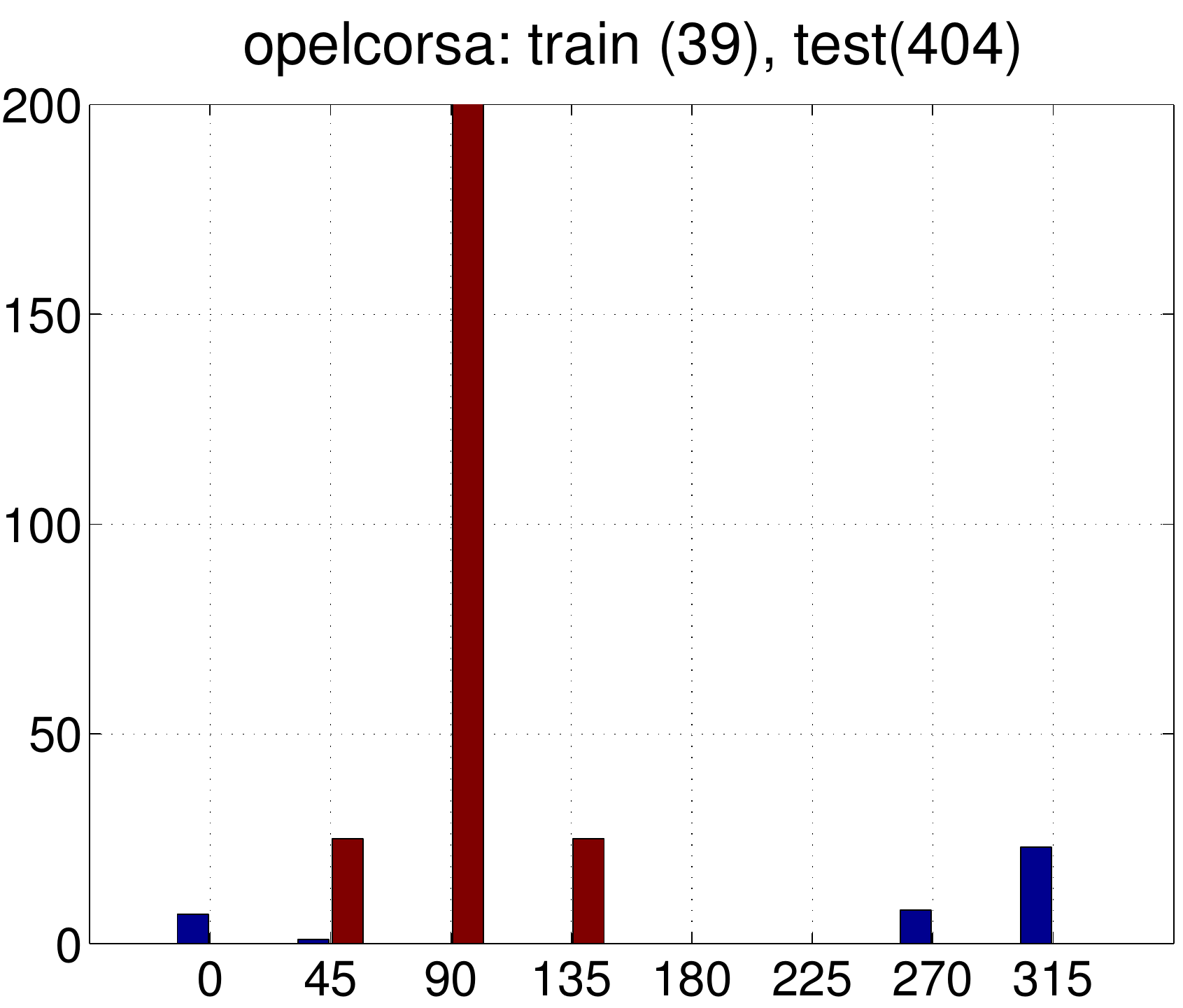}
\includegraphics[height=2.5cm]{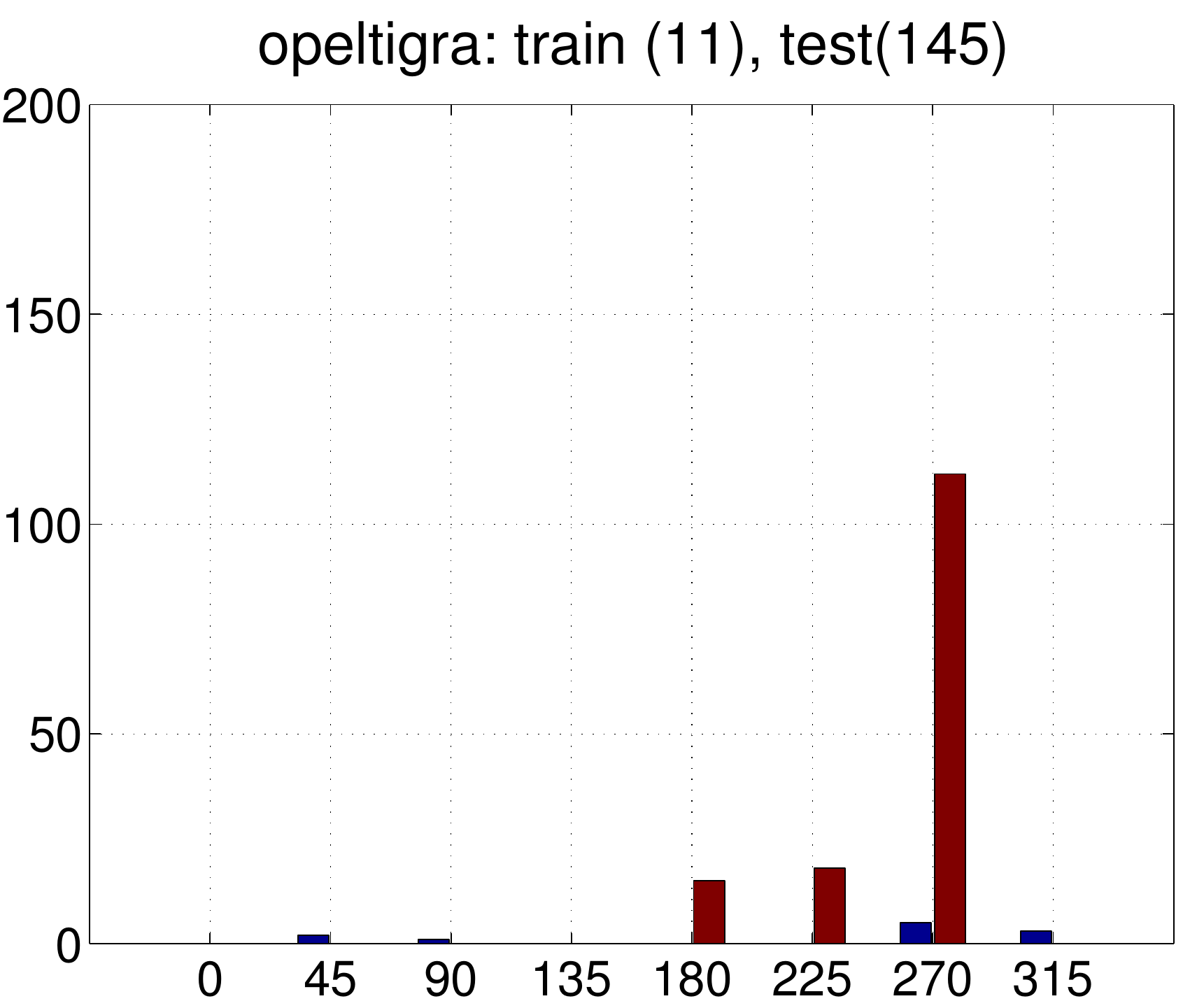}\\
\includegraphics[height=2.5cm]{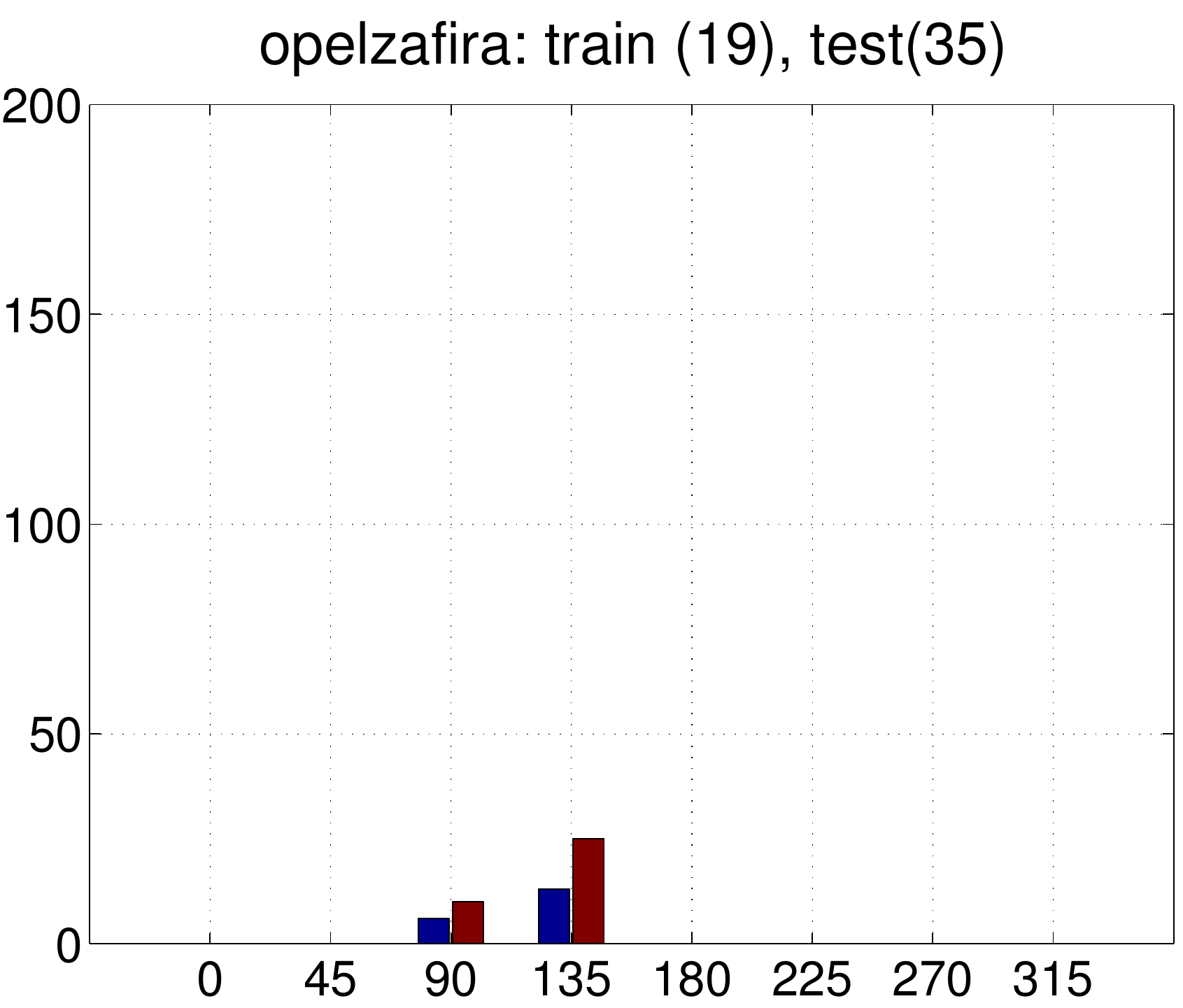}
\includegraphics[height=2.5cm]{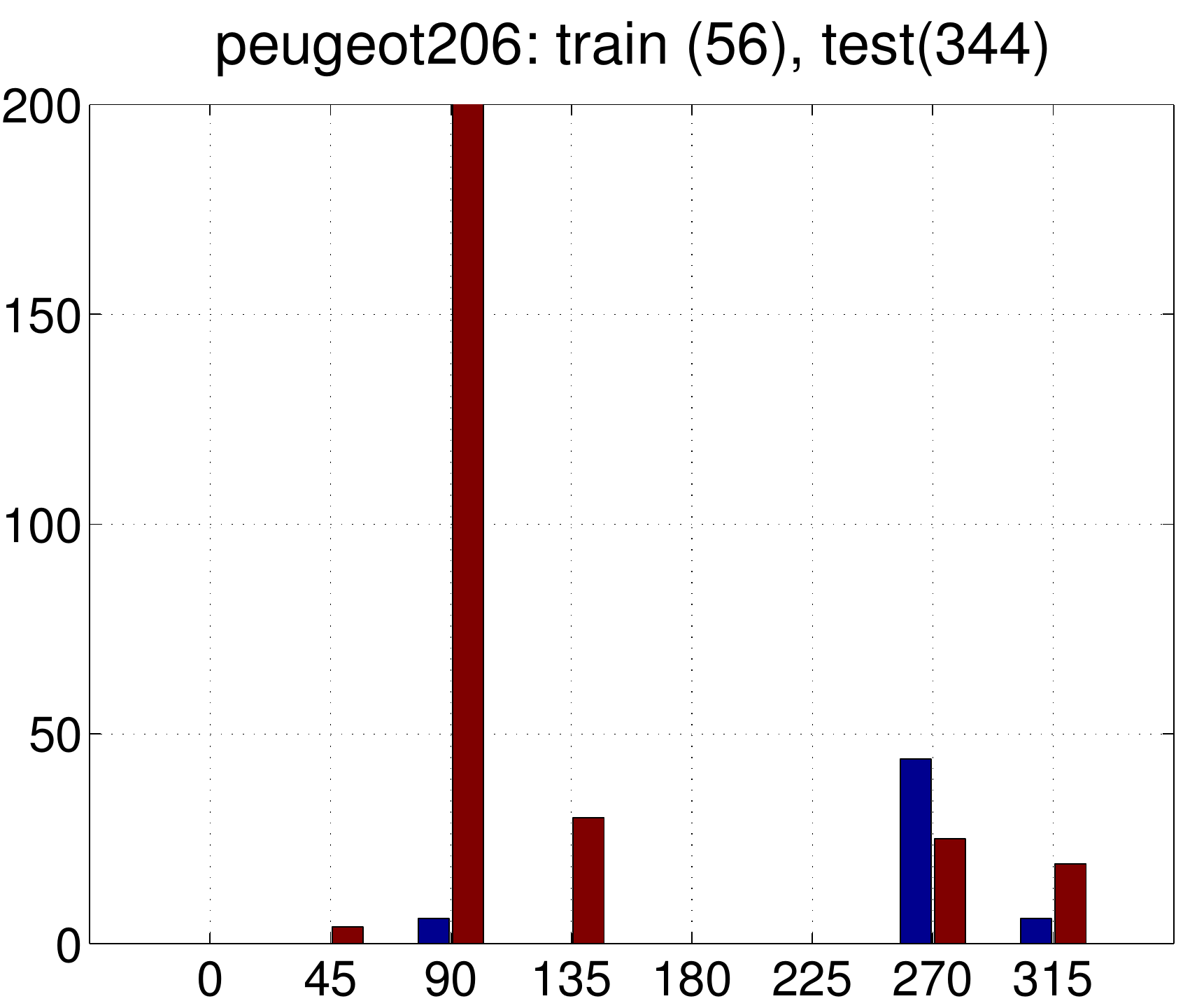}
\includegraphics[height=2.5cm]{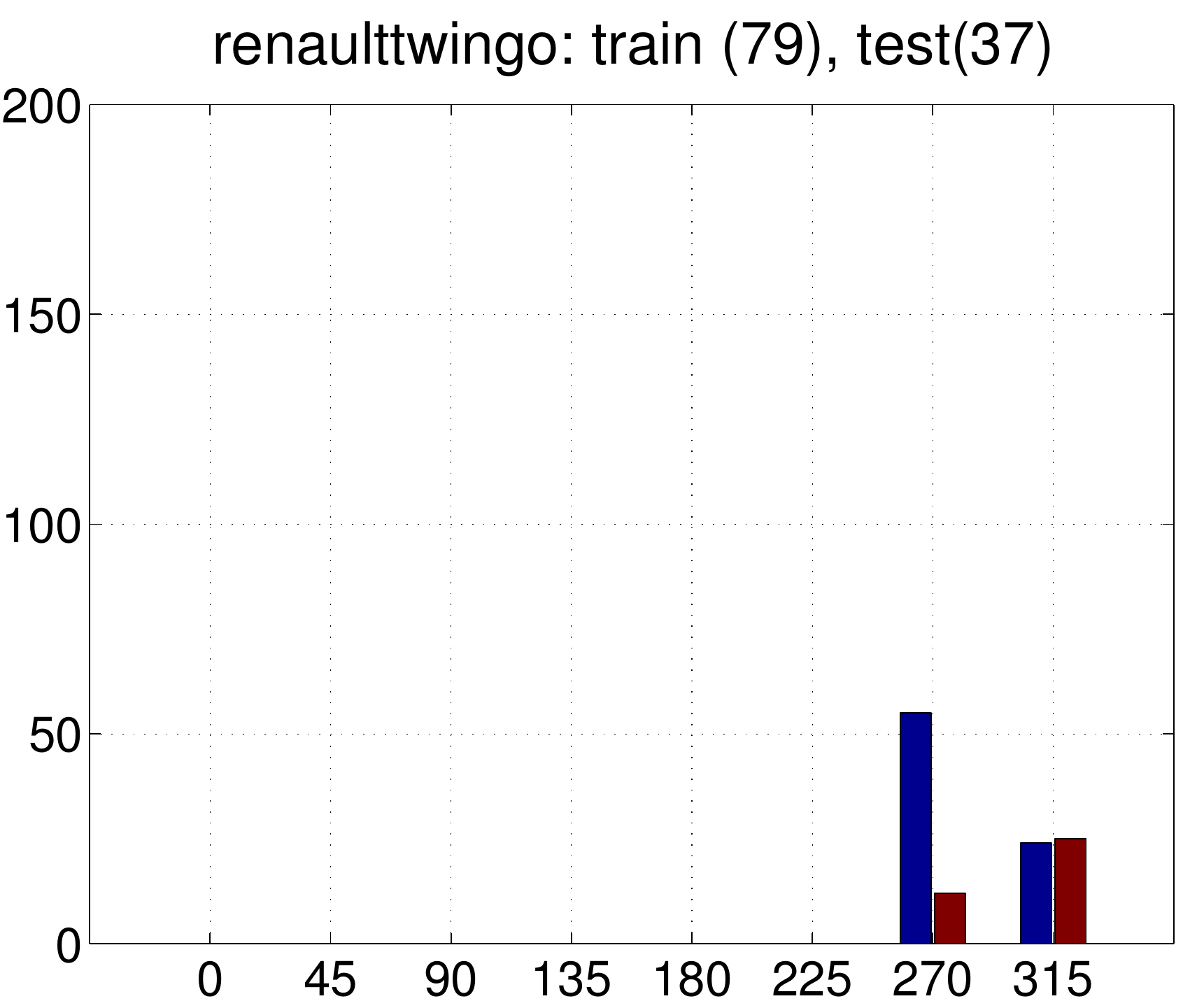}
\includegraphics[height=2.5cm]{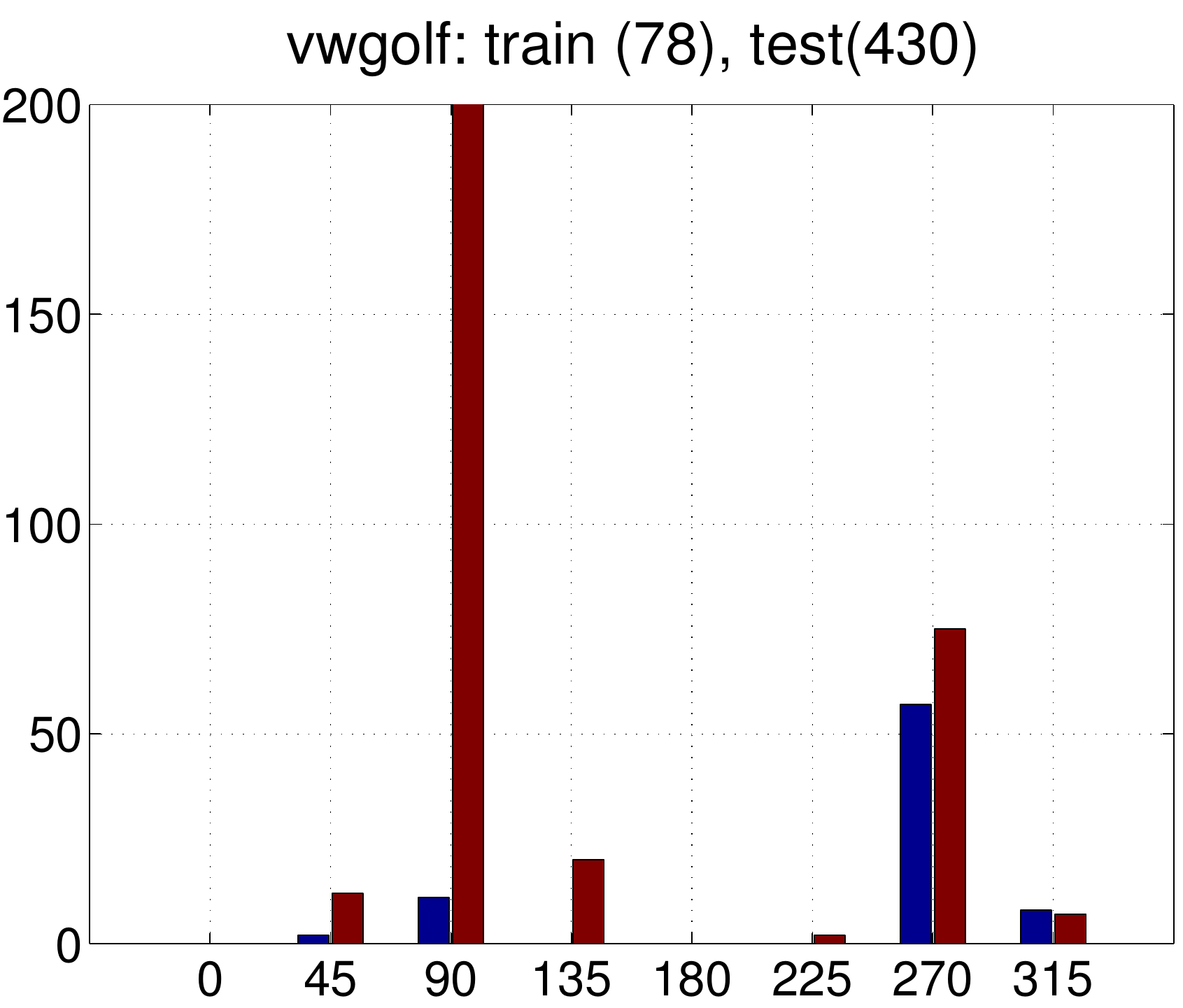}\\
\includegraphics[height=2.5cm]{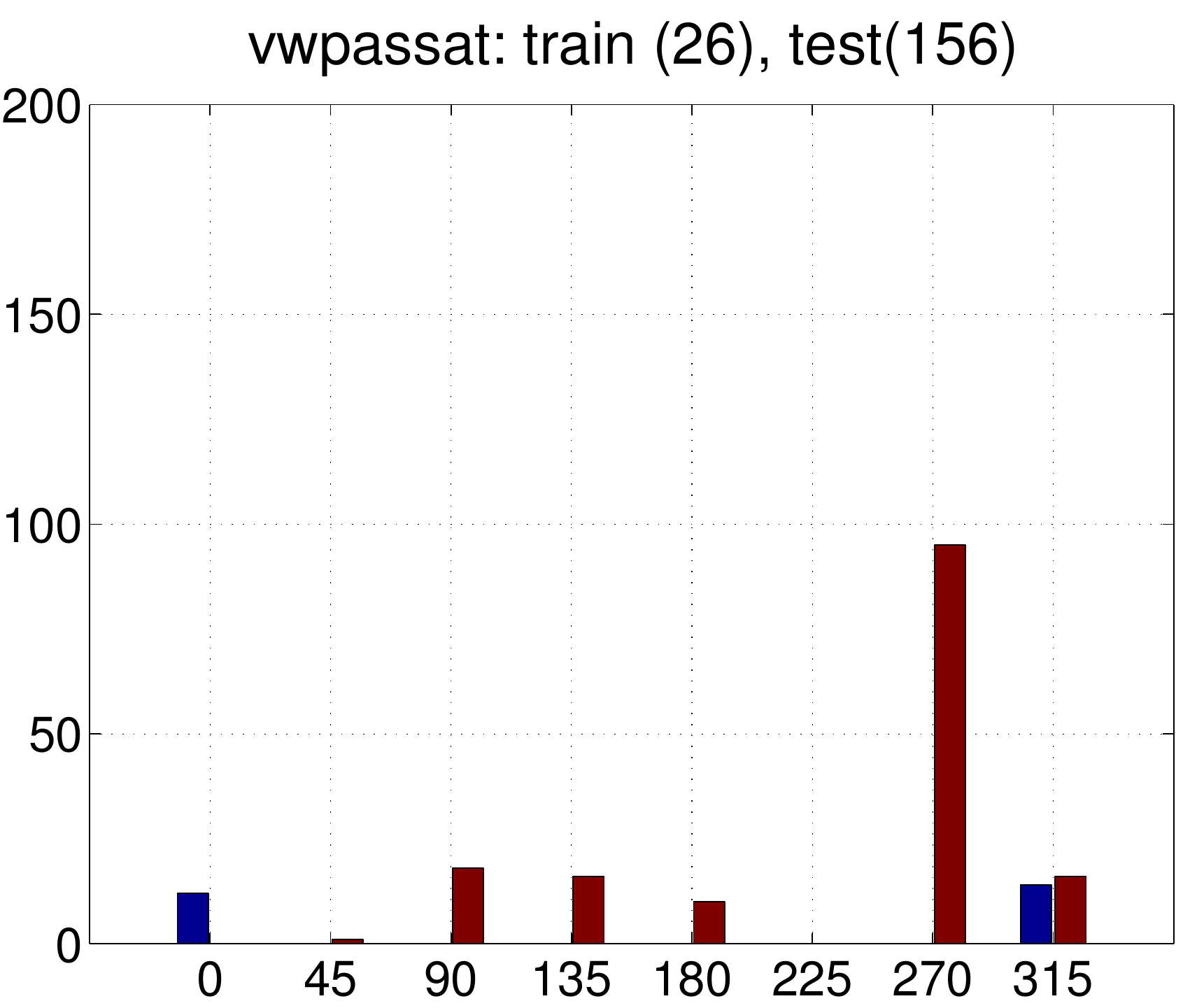}
\includegraphics[height=2.5cm]{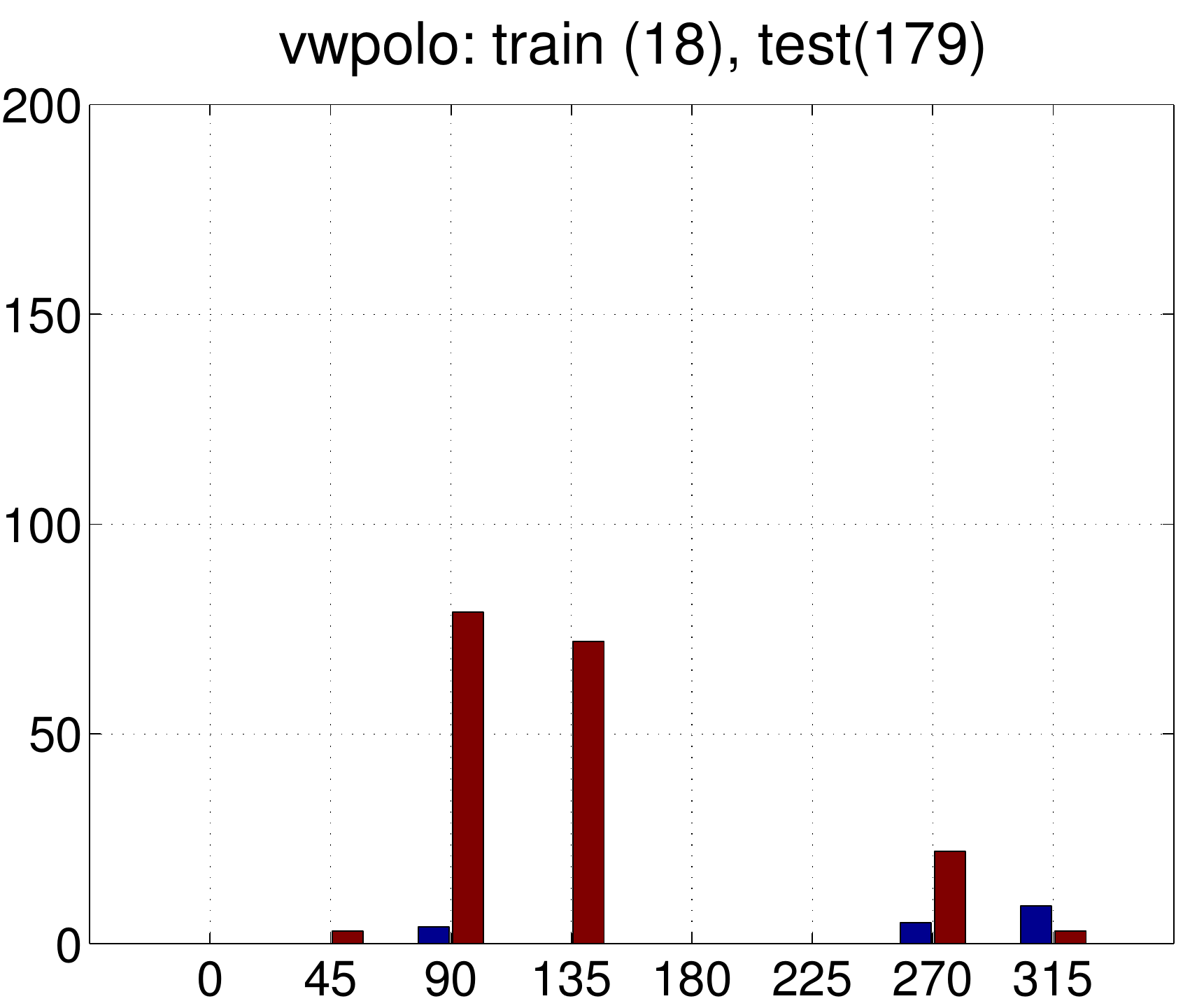}
\includegraphics[height=2.5cm]{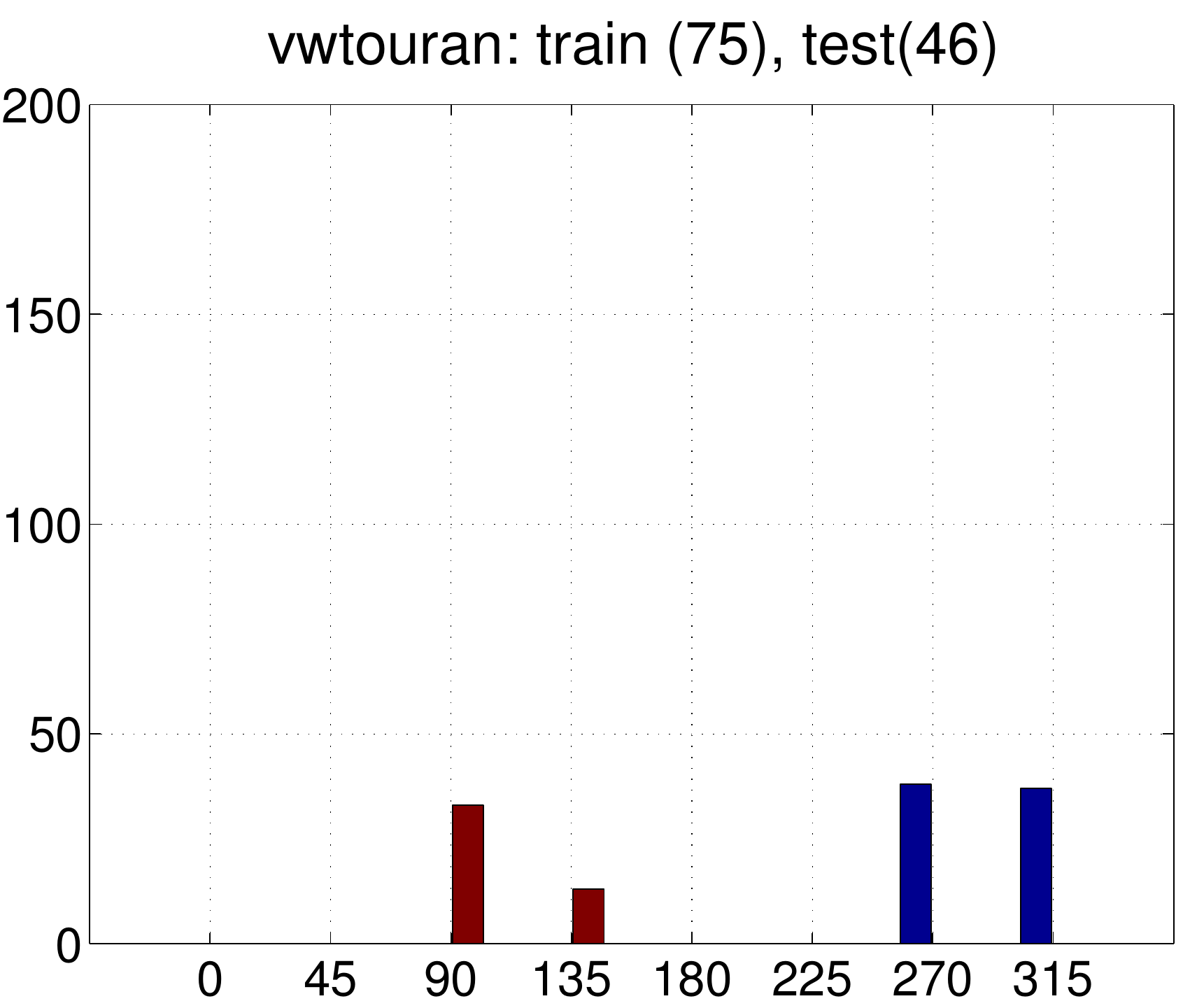}\\
\caption{{\em Car-models} train and test statistics over 8 viewpoint bins.}
\label{fig:carmodelstat}
\end{figure}

Lastly, Fig.~\ref{fig:carmodelstat} illustrates the train and test
distribution for each of the $23$ different {\em car-models} we have
annotated. For most of the {\em car-models}, the viewpoint train and
test distributions differ drastically. For {\em car-models} like {\em
  VW Touran, VW Passat, Opel Corsa, Opel Tigra, Opel Astra, Ford Ka,
  Ford Fiesta, Fiat Punto, Fiat Panda, BMW 1} there is either none or
at most $1$ viewpoint that has both training and test data. This
poses a major challenge for learning a robust multi-view object
detector. In addition, the absolute number of available training
examples for most of the {\em  car-models} is rather small ($12$ for
{\em Fiat Panda}, $19$ for {\em  Mini Cooper}, $12$ for {\em Opel
  Astra} etc.).
In summary, the training and test viewpoint data distributions
tend to be skewed and sparse, especially on the fine-grained category
levels,  
which, in addition to the low numbers of training examples,
represent serious challenges when learning models for the fine-grained
categories.

\section{Prior visualization in 3D}

In this section we visualize in 3D the
\MVcovmat, \MVcorr and the \MVcovmatNBtoALL priors learned for the {\em car}
class on the 3D object classes dataset~\cite{savarese07iccv}. For that
purpose, we sample a cell $(c,r)$ from viewpoint $v$ from the \target
model, which we call {\em reference} cell, and back-project on a 3D CAD
model the learned weights (dependencies) for that particular cell in
the $\Sigma_s$ (see Eq. (3) in the paper) correlation matrix
($K_s=\mathbf{I}-\lambda\Sigma_s$).

\begin{figure*}
\centering
\rotatebox{90}{model}\quad\quad
  \includegraphics[height=1.4cm]{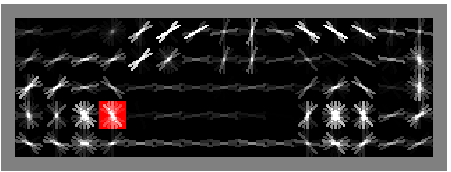}
  \includegraphics[height=1.4cm]{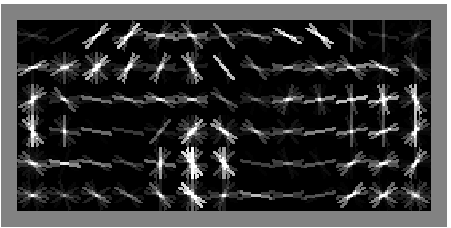}
  \includegraphics[height=1.4cm]{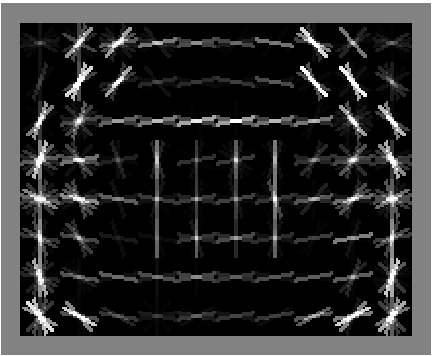}
  \includegraphics[height=1.4cm]{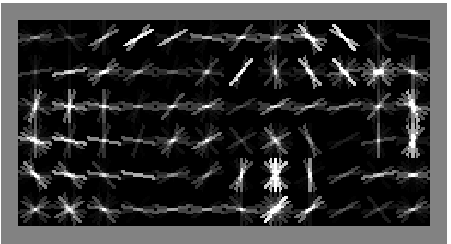}\\
\rotatebox{90}{{\em Target}}\quad\quad
  \includegraphics[height=1.4cm]{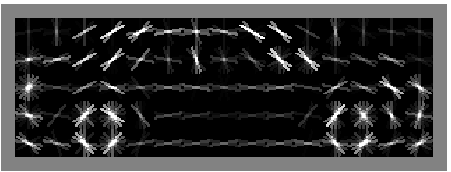}
  \includegraphics[height=1.4cm]{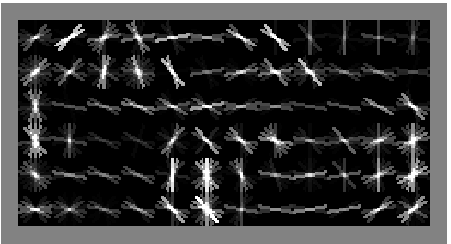}
  \includegraphics[height=1.4cm]{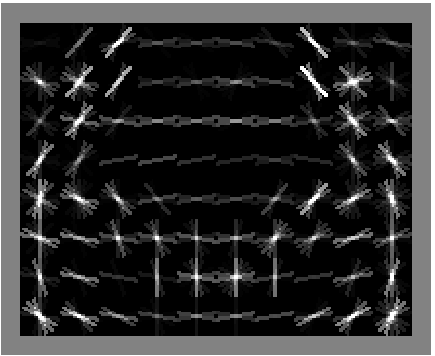}
  \includegraphics[height=1.4cm]{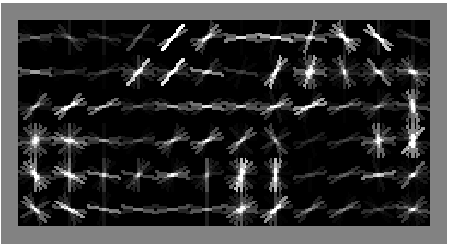}\\
\rotatebox{90}{\quad\quad\quad\quad\MVcovmat}\quad
  \includegraphics[width=0.4\textwidth]{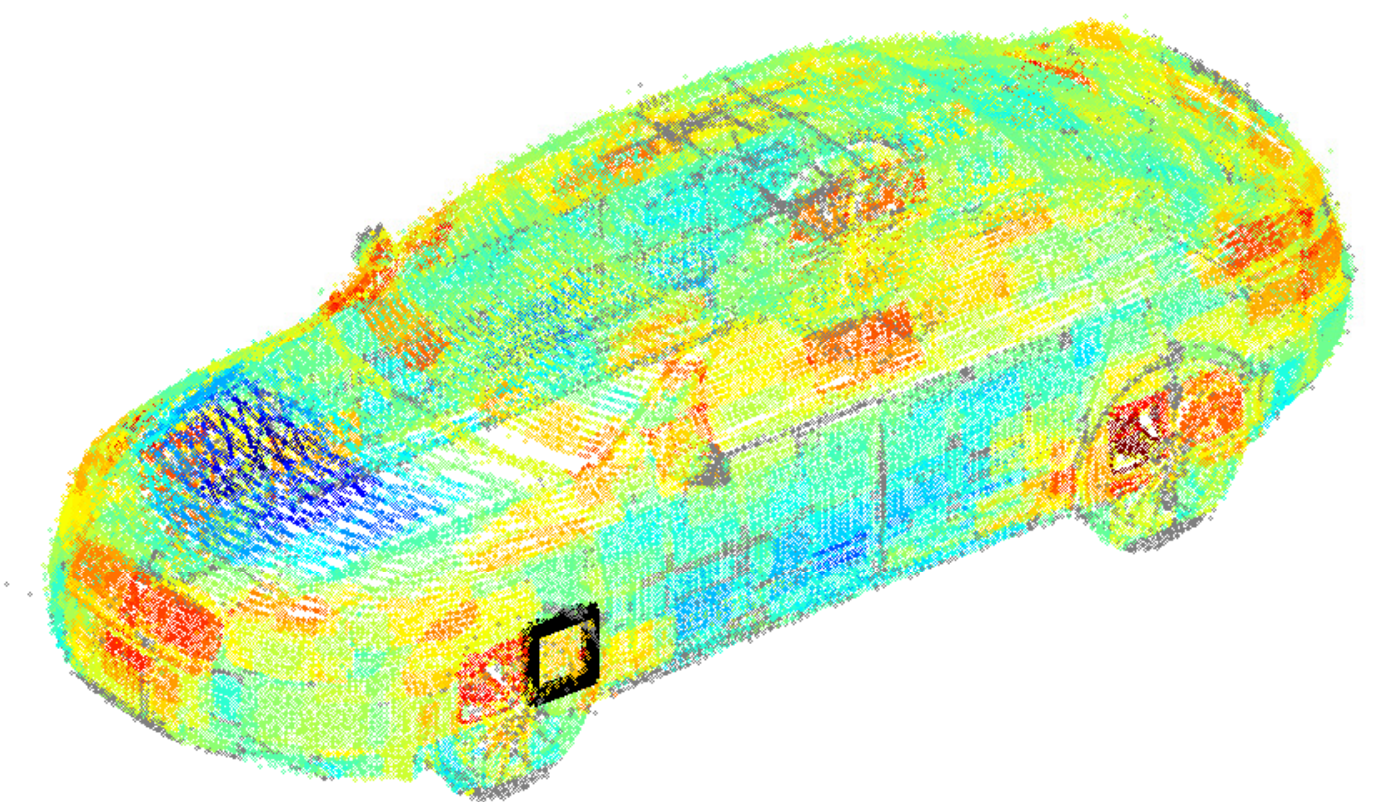}
  \includegraphics[width=0.4\textwidth]{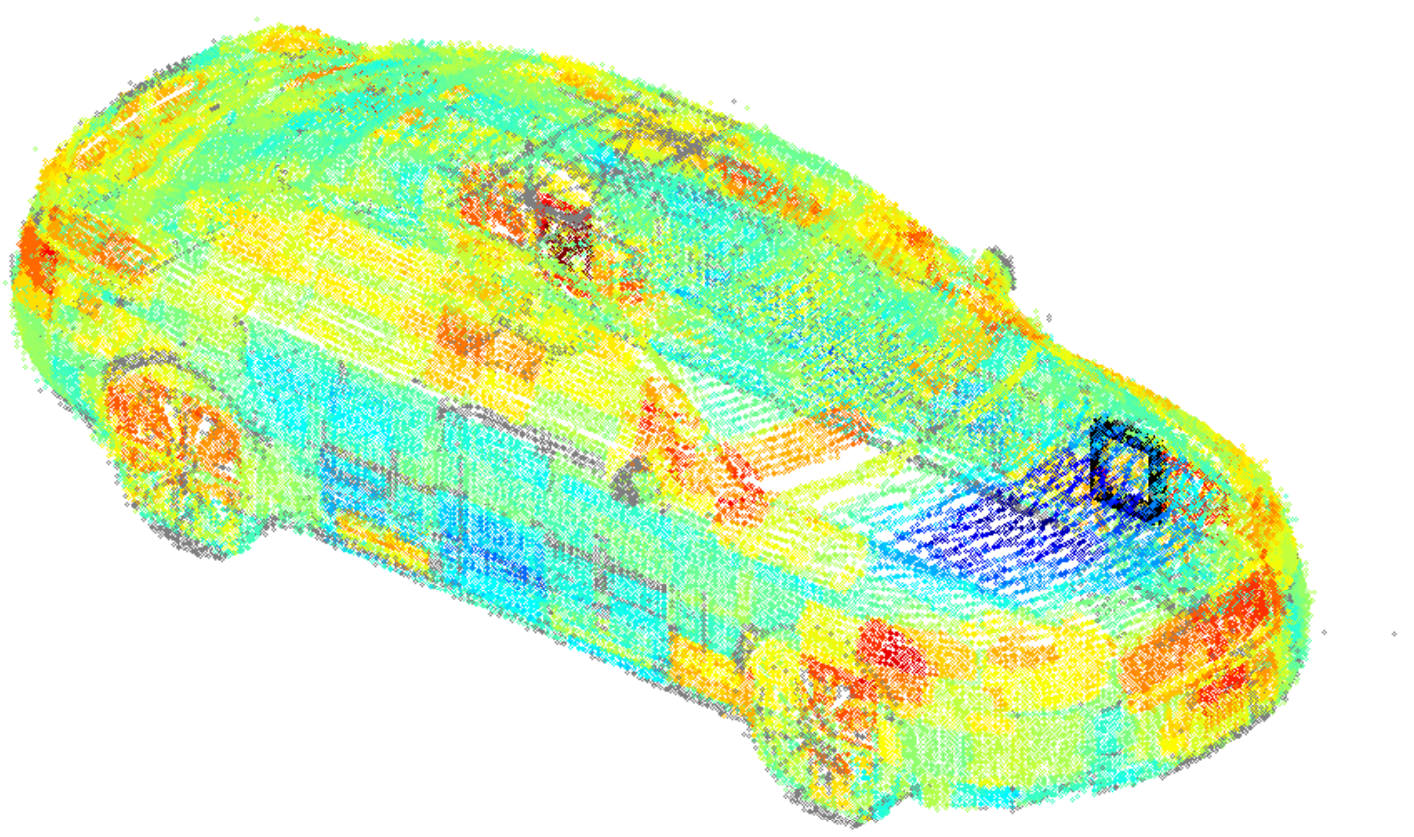}\\
\rotatebox{90}{\quad\quad\quad\MVcovmatNBtoALL}\quad
  \includegraphics[width=0.4\textwidth]{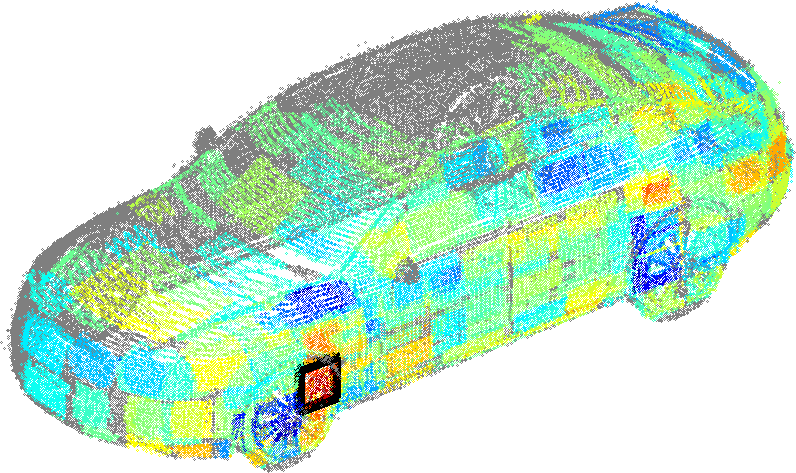}
  \includegraphics[width=0.4\textwidth]{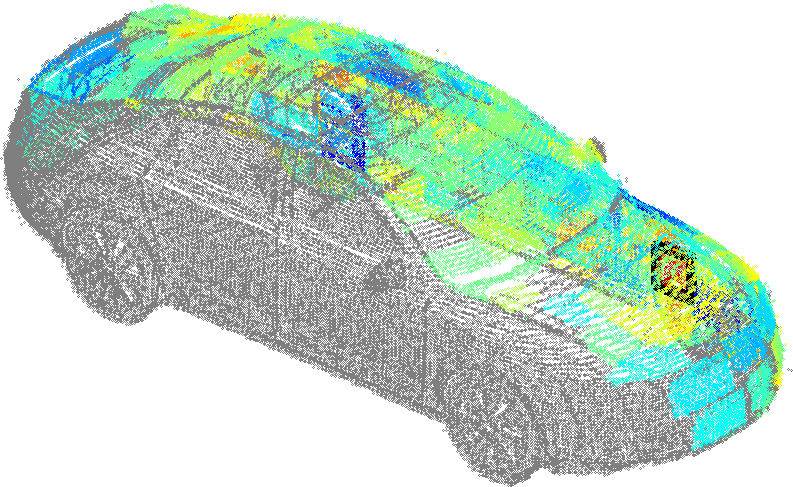}\\
\rotatebox{90}{\quad\quad\quad\quad\MVcorr}\quad
  \includegraphics[width=0.4\textwidth]{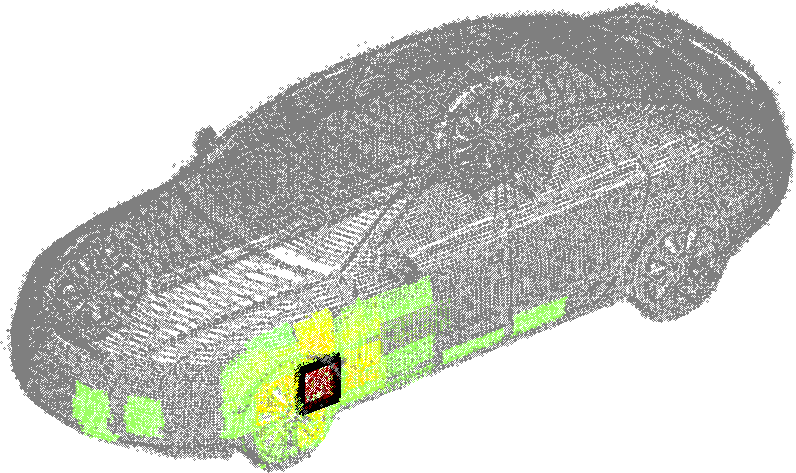}
  \includegraphics[width=0.4\textwidth]{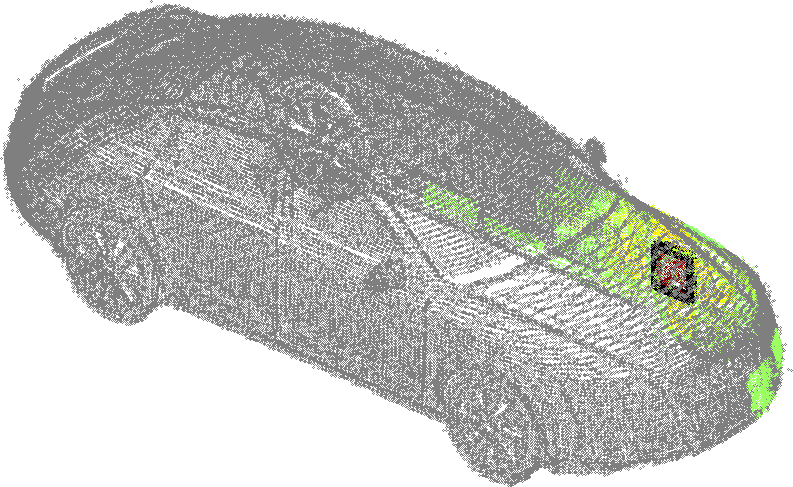}\\
  \caption{Prior visualization in 3D and \target model (row 1). Red
    indicates the {\em reference} cell. Prior visualizations in 3D for
    the red cell: \MVcovmat (row 2), \MVcovmatNBtoALL (row 3) and \MVcorr
    (row 4). The black cube indicates the {\em reference} cell
    back-projected into 3D.}
  \label{fig:priorVis}
\end{figure*}

Figure~\ref{fig:priorVis} visualizes the dependencies for an example
{\em reference} cell. On the top of the figure, an example \target
model of the ~{\em car} class, trained on the 3D object classes
dataset~\cite{savarese07iccv} is shown, along with the cell for which
the weights in the prior are visualized (denoted as red cell). As each
cell has $L=32$ dimensions, we average the dependencies across all of
them. Each cell is back-projected to 3D by aligning the model template
with the 3D CAD model. The alignment is two stage procedure involving
viewpoint alignment in the first stage and template to rendered object
alignment in the second stage. In Figure~\ref{fig:priorVis} rows 2, 3
and 4 visualize the cell dependencies for the \MVcovmat, \MVcovmatNBtoALL
and the \MVcorr priors, respectively (see Section~\ref{sec:approach} in the paper for
details on the different priors). The dependencies on the 3D CAD model
are visualized from two different views (left and right column).  Red
colors signify positive correlations, blue colors signify negative
correlations. If the cells are not correlated (all dependencies are
$0$), gray color is used.

\myparagraph{Observations.} The \MVcovmat model (row 2) reveals symmetric
structures in the object itself, which can be seen from the positive
dependencies (red color) on the 4 wheels of the {\em car} for the {\em
  reference} cell. \MVcovmatNBtoALL (row 3) and \MVcorr (row 4) establish
dependencies with only a few cells in the model (lots of gray points)
as they are both restricted to learning dependencies among neighboring
views (\MVcovmatNBtoALL) or neighboring cells (\MVcorr) only.

\end{document}